\documentclass[11pt]{article}
\usepackage[preprint]{acl}
\usepackage{times}
\usepackage{latexsym}
\usepackage[T1]{fontenc}
\usepackage[utf8]{inputenc}
\usepackage{microtype}
\usepackage{inconsolata}
\usepackage{graphicx}
\usepackage{booktabs}
\usepackage{tipa}
\usepackage{gb4e}
\noautomath
\usepackage{tabularx}
\usepackage{avm}
\usepackage{xcolor}
\usepackage{colortbl}
\usepackage[most]{tcolorbox}
\usepackage{comment}
\usepackage[caption=false,font=small]{subfig}

\newcommand{\accverylow}[1]{\cellcolor{green!8}#1}
\newcommand{\acclow}[1]{\cellcolor{green!14}#1}
\newcommand{\accmid}[1]{\cellcolor{green!22}#1}
\newcommand{\acchigh}[1]{\cellcolor{green!30}#1}
\newcommand{\accveryhigh}[1]{\cellcolor{green!38}#1}

\title{From Parallels to Divergence: Language Models’ Learning of Action Nominal Constructions}

\author{Iskar Deng, Shane Steinert-Threlkeld\\
  University of Washington \\
  \texttt{\{hd49,shanest\}@uw.edu}}

\begin{document}
\maketitle
\begin{abstract}
Action nominal constructions (ANCs) express events in nominal form, with their argument marking following patterns found in clauses or ordinary noun phrases. In this paper, we investigate how different grammatical configurations affect language models' learning of ANCs and related constructions, and how these learning preferences compare with typological distributions in natural languages. We use a grammar-engineered method for synthetic corpus construction that combines shared semantic input from natural corpora with computational grammars. We construct 96 target languages varying in clause word order, noun-phrase word order, alignment, complement system, and ANC strategy. For each language, we train GPT-2-small models from scratch and evaluate them using targeted minimal pairs covering 34 grammatical phenomena. The results show typological parallels in basic clause configurations but divergences in complement systems and ANC strategies, with models also learning combinations rare or unattested in natural languages. This partial alignment suggests that natural language distributions may reflect both general learning pressures and functional demands not fully captured by model learning experiments.\footnote{Code for our experiments is available at: \url{https://github.com/Iskar-Deng/anc-typology-bias}.}
\end{abstract}

\section{Introduction}

Language models (LMs) exhibit different learning preferences across grammatical systems. Prior work on word order found that models learn configurations that follow common word-order correlations more easily than those that violate them \citep{kuribayashi2024emergent,xu2025can}, and work on differential argument marking found that models more easily learn systems in which overt marking targets semantically atypical arguments \citep{deng2026differences}. Comparing these preferences with typological tendencies in natural languages shows that models can exhibit typologically aligned behavior even when these tendencies are not explicitly encoded, but this alignment does not extend to all typological patterns \citep{white2021examining,deng2026differences,popadich2026language}. Patterns of alignment and divergence can therefore help distinguish which typological tendencies may arise from learning pressures captured by LMs and which require additional factors to explain \citep{white2021examining,futrell2025how,deng2026differences}.

Prior work has mainly examined particular phenomena, such as word order and case marking \citep{kuribayashi2024emergent,xu2025can,deng2026differences}. However, a construction does not exist in isolation within a grammatical system, but may share some structural properties with other constructions while differing in others. We therefore use action nominal constructions (ANCs) as a case study, extending the study of LM learning preferences from individual phenomena to relationships across constructions. ANCs express events or processes in nominal form, allowing them to occur in positions typically occupied by noun phrases \citep{koptjevskaja1993nominalizations}. Across languages, ANC argument marking, predicate form, and internal word order may follow patterns found in finite clauses, clausal complements, or noun phrases \citep{koptjevskaja1993nominalizations,Ruditsky2025}. For example, English encodes the same event participants as subject and object in a clause, but as a possessive dependent and an \textit{of}-phrase in an ANC:
\begin{exe}
\ex
\begin{xlist}
\ex \textit{John destroyed the city.}
\ex \textit{John's destruction of the city}
\end{xlist}

\end{exe}

In this study, we compare models' learning preferences for ANCs and related constructions with natural language typology. We use a synthetic-language method based on grammar engineering. We derive shared semantic representations from natural-language data and realize the same underlying meanings through different target grammars. Using the Grammar Matrix \citep{bender-etal-2010-grammar,Ruditsky2025}, we combine four ANC strategies with three clause word orders, two noun-phrase word orders, two alignment systems, and two complement-system types, yielding 96 target languages. The semantic input comes from the BabyLM 2026 Strict corpus \citep{choshen2026babylmturns4}. For each target language, we train GPT-2-small models from scratch \citep{radford2019gpt2}. We then construct targeted minimal pairs covering 34 grammatical phenomena to evaluate the models' learning of the grammatical regularities encoded in each target grammar.

The results show that models reproduce typological tendencies in basic clause configurations relatively well. Their learning preferences for word-order and argument-marking combinations align with natural language distributions, suggesting that general learning pressures may help shape these tendencies. For more complex configurations involving complement systems and ANC strategies, the correspondence is only partial. Models learn not only typologically common configurations but also some combinations that are rare or unattested in the natural language sample. This divergence may relate to differences in constructional functions across languages and the division of labor among alternative forms of expression, reflecting communicative and usage pressures that the current controlled experiments may not fully capture.

After reviewing prior work using synthetic corpora and introducing ANCs in Section~\ref{sec:background}, we define the grammatical configurations in Section~\ref{chap:experimental-parameters}. Sections~\ref{chap:corpus-generation} and~\ref{chap:model-training} describe corpus generation and model training, respectively. We then present the evaluation framework and results in Section~\ref{chap:evaluation} and compare the findings with natural language typology in Section~\ref{chap:discussion}, before concluding in Section~\ref{chap:conclusion}.

\section{Background}
\label{sec:background}
\subsection{Synthetic Corpus Experiments}

Examining typological preferences in LMs requires comparing their learning across languages or grammatical configurations. However, direct comparisons across natural languages make it difficult to isolate the effects of grammatical properties, since languages also differ in corpus source, training data size, and lexical distributions \citep{cotterell2018languages,mielke2019kind}. Controlled synthetic-language experiments address this problem by controlling for these differences while systematically varying grammatical properties, allowing their effects on model learning to be examined more directly \citep{ravfogel-etal-2019-studying,white2021examining}.

Two main approaches have been used to construct such controlled corpora. Corpus perturbation constructs counterfactual languages by modifying syntactic structures in natural corpora. It preserves relatively natural input distributions, but is typically limited to modifying structures already present in the source data, making it difficult to introduce a new and globally coherent grammatical system \citep{wang2017galactic,ravfogel-etal-2019-studying,kallini2024mission,xu2025can,deng2026differences}. Artificial-grammar methods generate corpora from explicit rules, allowing direct control over parameter combinations and the generation of rare or unattested grammatical configurations. However, lexical distributions, generation probabilities, and construction frequencies are usually specified by the researcher, making it difficult to approximate the predicate--argument distributions and syntax--semantics interactions found in natural language \citep{white2021examining,kuribayashi2024emergent,el-naggar-etal-2025-word}.

\begin{table*}[t]
\centering
\small
\begin{tabular}{llll}
\toprule
Type & Transitive ANC & Intransitive ANC & Example language \\
\midrule
\textsc{SENT}
& John-\textsc{nom} destruction city-\textsc{acc}
& John-\textsc{nom} running
& Korean \\
\textsc{POSS-ACC}
& John-\textsc{gen} destruction city-\textsc{acc}
& John-\textsc{gen} running
& Selkup \\
\textsc{ERG-POSS}
& John-\textsc{instr} destruction city-\textsc{gen}
& John-\textsc{gen} running
& Russian \\
\textsc{NOMN}
& John-\textsc{gen} destruction city-\textsc{obl}
& John-\textsc{gen} running
& Icelandic \\
\bottomrule
\end{tabular}
\caption{Four major nominalization strategies illustrated with pseudo-examples. Case markers are simplified.}
\label{tab:anc-types}
\end{table*}

Given these limitations, we propose a grammar-engineered method for synthetic corpus construction. Different ANC encoding strategies involve multiple properties, including argument marking, predicate form, and internal word order, and are closely connected to the language's alignment and complement systems \citep{koptjevskaja1993nominalizations,Ruditsky2025}. Corpus construction therefore requires coordinated control over these properties while preserving relatively natural predicate--argument distributions. Our method derives shared semantic representations from natural corpora and realizes them through different target grammars. Implementation details are provided in Section~\ref{chap:corpus-generation}.

\subsection{Action Nominal Constructions}
\label{sec:action-nominal-constructions}

Action nominal constructions (ANCs) express events or processes in nominal form while retaining aspects of the corresponding predicate's argument structure \citep{koptjevskaja1993nominalizations}. Their head is typically derived from a verb, and event participants may be expressed as arguments within the construction. ANCs therefore exhibit both nominal and clausal properties: externally, they can occur in nominal positions such as subjects, objects, or prepositional complements; internally, they may retain argument selection, case marking, and other properties of verbal predicates, or use nominal encoding patterns such as genitive marking \citep{grimshaw1990argument,alexiadou2001functional}. ANCs, main clauses, and clausal complements may express the same event and its participants, but differ in their syntactic distribution and functions. Main clauses typically constitute independent assertions about events or states; clausal complements embed content such as propositions, facts, or events as arguments of matrix predicates; ANCs express events or processes in nominal form, allowing them to be referred to and further discussed \citep{koptjevskaja1993nominalizations}. Thus, ANCs and clausal complements may occupy similar argument positions, but their semantic functions are not identical, and their internal organization may exhibit different degrees of clausal and nominal properties.

The main differences among ANC strategies lie in how the arguments of the nominalized predicate are encoded. \citet{koptjevskaja1993nominalizations} distinguishes four major strategies along a continuum from more clause-like to more noun-phrase-like structures. Here, S denotes the sole argument of an intransitive predicate, A the agent-like argument of a transitive predicate, and P the patient-like argument of a transitive predicate. Sentential (SENT) ANCs retain the argument-marking patterns found in finite clauses and occupy the most clause-like end of the typology. In Possessive-Accusative (POSS-ACC) ANCs, S and A are recoded as genitive-like dependents, while P retains the object marking found in finite clauses. Ergative-Possessive (ERG-POSS) ANCs instead group S and P together through genitive-like marking, while A is marked separately, often by an oblique or instrumental expression. These two strategies combine nominal and clausal resources. Nominal (NOMN) ANCs are most closely aligned with ordinary nominal structure. S and A are encoded as genitive-like dependents, while P is assimilated to a dependent type available in ordinary noun phrases. A single language may employ more than one ANC strategy, with their distribution depending on lexical, semantic, or functional factors \citep{koptjevskaja1993nominalizations}. Table~\ref{tab:anc-types} illustrates these four strategies with simplified examples.

ANC arguments need not all be overtly expressed. When an argument is recoverable from context or irrelevant to the communicative goal of the sentence, it may remain unexpressed \citep{koptjevskaja1993nominalizations}. These strategies should therefore be understood as patterns of argument encoding when arguments are overtly realized, rather than as requiring every ANC to express all of its possible arguments. Their implementation in this study is described in Section~\ref{chap:experimental-parameters}, with more detailed descriptions and natural-language examples provided in Appendix~\ref{app:anc-strategies}.

\section{Grammatical Configurations}
\label{chap:experimental-parameters}

We construct target languages by combining five typological parameters: clause word order, noun-phrase word order, alignment, complement system, and ANC strategy. The possible values are listed in Table~\ref{tab:crossed-parameters}, yielding $3 \times 2 \times 2 \times 2 \times 4 = 96$ grammatical configurations. These parameters jointly determine word order, argument marking, and predicate form across constructions in the target languages.

\begin{table}[t]
\centering
\small
\begin{tabularx}{\columnwidth}{@{}lX@{}}
\toprule
Parameter & Values \\
\midrule
Clause word order & SOV / SVO / VOS \\
NP word order & GN / NG \\
Alignment & NOM-ACC / ERG-ABS \\
Complement system & Balancing / Deranking \\
ANC strategy & SENT / POSS-ACC / ERG-POSS / NOMN \\
\bottomrule
\end{tabularx}
\caption{Five typological parameters and their values.}
\label{tab:crossed-parameters}
\end{table}

\begin{table*}[t]
\centering
\small
\begin{tabularx}{\textwidth}{@{}XXX@{}}
\toprule
Original sentence
& Pseudo-English
& Generated form \\
\midrule
\textit{Mary is sleeping.}
& \textit{Mary sleeps}
& \textit{Mary sleeps} \\

\textit{I eat an apple.}
& \textit{I eats apple}
& \textit{I eats appleca} \\

\textit{He wants to eat an apple.}
& \textit{He wants that he eats apple}
& \textit{He wants he eats appleca} \\

\textit{I think Mary sleeps.}
& \textit{I thinks that mary sleeps}
& \textit{I thinks mary sleeps} \\

\textit{He is looking for a book.}
& \textit{He lookfors book}
& \textit{He lookfors bookca} \\

\textit{Why don't you get up a collection?}
& \textit{You getups collectnmz}
& \textit{You getups collectingca} \\

\textit{John's destruction of the city shocked everyone.}
& \textit{Johnge destroynmz cityob shocks everyone}
& \textit{Johnge destroying cityob shocks everyoneca} \\
\bottomrule
\end{tabularx}
\caption{Examples of pseudo-English and target-language realization. The target configuration uses SVO, GN, NOM-ACC, Balancing, and NOMN.}
\label{tab:pseudo-english-examples}
\end{table*}

Specifically, clause word order determines the order of predicates and core arguments in main clauses. Nominative--accusative alignment (NOM-ACC) marks P with \textit{-ca}, while ergative--absolutive alignment (ERG-ABS) marks A with the same suffix. Other core arguments receive zero marking. Main-clause predicates take the finite suffix \textit{-s}. Genitive dependents take \textit{-ge} and precede or follow the nominal head under GN or NG order, respectively. Clausal complements share main-clause word order and argument marking, with neither overt complementizers nor external case marking. 

The complement system determines predicate form in clausal complements. In Balancing constructions, the subordinate predicate has a form similar to one that can occur in an independent declarative clause. In Deranking constructions, it takes a reduced form that cannot normally occur as the predicate of an independent main clause \citep{stassen1985comparison,cristofaro2003subordination}. In this experiment, Balancing predicates take the finite suffix \textit{-s}, as in main clauses, while Deranking predicates take the nonfinite suffix \textit{-ing}.

All ANC predicates take \textit{-ing}. ANC strategy determines which internal arguments retain clausal marking and which receive genitive \textit{-ge} or oblique \textit{-ob} marking. ANC-internal word order depends jointly on clause word order, NP word order, and ANC strategy \citep{koptjevskaja1993nominalizations}. The ANC as a whole receives case or genitive marking according to its external role. Argument omission is permitted inside ANCs but not in main or complement clauses. Morphological suffixes attach directly to stems without hyphens or spaces.

Table~\ref{tab:pseudo-english-examples} provides example sentences for a configuration combining SVO, GN, NOM-ACC, Balancing, and NOMN. Detailed mapping rules are provided in Appendix~\ref{app:experimental-configurations}.

\section{Corpus Generation}
\label{chap:corpus-generation}

\begin{figure*}[t]
    \centering
    \includegraphics[width=\textwidth]{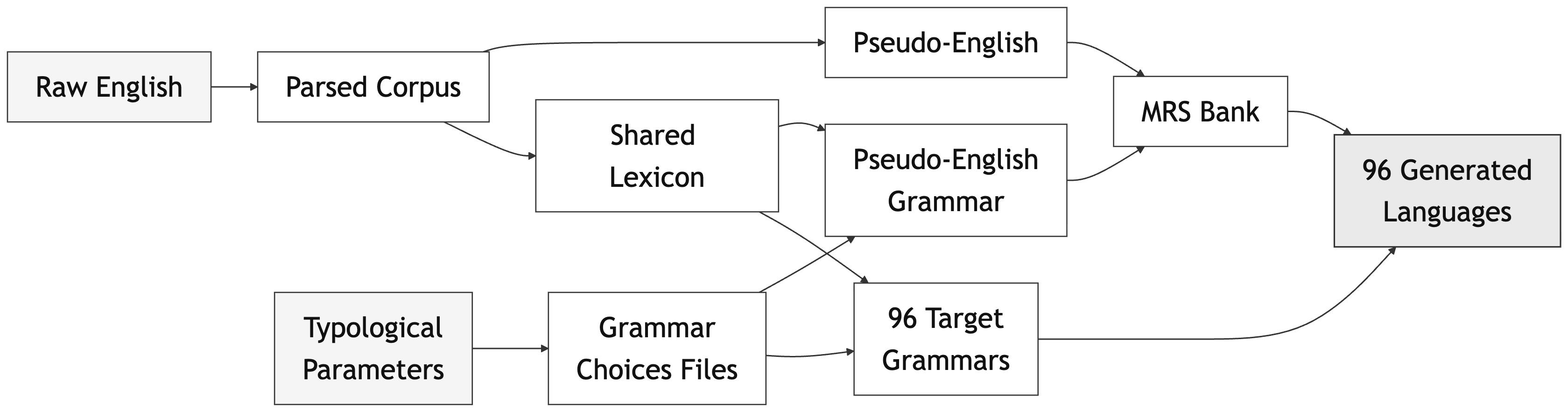}
    \caption{Overview of the corpus-generation pipeline.}
    \label{fig:pipeline}
\end{figure*}

This section describes how the grammatical configurations defined in Section~\ref{chap:experimental-parameters} are converted into parallel synthetic corpora. The pipeline takes natural English as a source of predicate--argument content, maps it into shared semantic representations, and then realizes the same underlying meanings through 96 target grammars. Figure~\ref{fig:pipeline} provides an overview of the pipeline.

We use the 100M-word BabyLM 2026 Strict corpus \citep{choshen2026babylmturns4} as our source corpus. This scale has supported successful language learning in prior work \citep{warstadt2023findings,kallini2024mission,xu2025can} and remains manageable for controlled experiments. We then use spaCy dependency parses \citep{spacy2020} to process this corpus, deriving pseudo-English for subsequent semantic parsing and a shared lexicon. Pseudo-English retains the core predicate--argument relations from the original sentences while removing additional surface phenomena such as adjectival modification and auxiliary structure, as shown in Table~\ref{tab:pseudo-english-examples}. Nominalizing suffixes and stem similarity between nouns and verbs are used to identify possible nominalizations and their source verbs. Only intransitive and transitive verbs are treated as nominalizable predicates, excluding verbs that take clausal complements.\footnote{We use \textit{-nmz} to mark nominalized predicates in pseudo-English, preventing the parser from treating them as ordinary nouns in the lexicon and preserving the intended ANC analysis. In the generated corpora, we use \textit{-ing}, allowing nominalized predicates to share surface forms with ordinary nouns so that models can learn associations between them.} The shared lexicon preserves forms extracted from the source corpus, retaining some of its lexical distribution and word-form regularities.

We construct computational grammars with different typological properties using the Grammar Matrix, a grammar customization system based on Head-Driven Phrase Structure Grammar (HPSG) \citep{pollard1994head,bender-etal-2010-grammar}. We encode the grammatical configurations defined in Section~\ref{chap:experimental-parameters} as choices files, each specifying the grammatical settings of one target language. A pseudo-English choices file is used to build the grammar for parsing pseudo-English. The Grammar Matrix generates HPSG grammars from these files, and the shared lexicon is then inserted into both the pseudo-English grammar and all target grammars. These grammars are compiled with ACE, a parsing and generation engine \citep{packard2018ace}.

Finally, the pseudo-English corpus is parsed into a shared bank of Minimal Recursion Semantics (MRS) representations, which encode predicates and their argument relations and serve as the semantic input to the target grammars \citep{copestake2005minimal}. Each of the 96 target grammars uses ACE to realize these representations according to its own morphosyntactic settings. Table~\ref{tab:pseudo-english-examples} shows original English sentences, their pseudo-English forms, and the realizations. When an input produces multiple candidates, we select a final output using rules such as the target ANC word order. Further details on corpus generation, the grammar-engineering framework, and tools are provided in Appendix~\ref{app:corpus-generation-details}.

\section{Model Training}
\label{chap:model-training}

For each of the 96 target languages, we train three GPT-2-small models from scratch on a generated corpus of approximately 4.1 million sentences. The three runs use different random seeds but share the same training configuration, differing only in parameter initialization and data shuffling. All models are trained for 70,000 steps, and we evaluate the final checkpoint. Because tokenized corpus sizes vary across languages, this schedule corresponds to approximately 2.59--2.96 corpus-equivalent epochs. Details of data preparation and the full training configuration are provided in Appendix~\ref{app:model-training}.

\section{Evaluation}
\label{chap:evaluation}

\begin{table*}[t]
\centering
\small
\setlength{\tabcolsep}{3pt}

\begin{minipage}[t]{0.49\textwidth}
\vspace{0pt}
\begin{tabularx}{\linewidth}{@{}lXr@{}}
\toprule
ID & Phenomenon & Acc. (\%) \\
\midrule
1.1 & Intransitive finite verb form
    & \accveryhigh{99.4} \\
1.2 & Intransitive S marker
    & \accmid{87.7} \\
1.3 & Intransitive S--V order
    & \accveryhigh{100.0} \\
1.4 & Transitive finite verb form
    & \accveryhigh{99.7} \\
1.5 & Transitive A marker
    & \accveryhigh{96.6} \\
1.6 & Transitive P marker
    & \accveryhigh{97.9} \\
1.7 & Transitive A--V order
    & \accveryhigh{100.0} \\
1.8 & Transitive V--P order
    & \accveryhigh{100.0} \\
1.9 & Intransitive V valency
    & \acclow{70.4} \\
1.10 & Transitive V valency
    & \acclow{76.2} \\
\midrule
2.1 & Intransitive genitive marker
    & \accmid{89.3} \\
2.2 & Transitive genitive marker
    & \acchigh{91.6} \\
2.3 & Genitive possessor--head order
    & \acchigh{94.9} \\
\midrule
3.1 & Clausal complement verb form
    & \accveryhigh{98.6} \\
3.2 & Clausal complement S marker
    & \acclow{75.2} \\
3.3 & Clausal complement A marker
    & \accmid{83.9} \\
3.4 & Clausal complement P marker
    & \accveryhigh{97.7} \\
3.5 & Clausal complement--V order
    & \accveryhigh{100.0} \\
3.6 & Clausal complement CV valency
    & \acclow{74.9} \\
\bottomrule
\end{tabularx}
\end{minipage}
\hfill
\begin{minipage}[t]{0.49\textwidth}
\vspace{0pt}
\begin{tabularx}{\linewidth}{@{}lXr@{}}
\toprule
ID & Phenomenon & Acc. (\%) \\
\midrule
4.1 & ANC verb form
    & \acchigh{91.9} \\
4.2 & ANC S marker
    & \accmid{82.1} \\
4.3 & ANC A marker
    & \accverylow{68.7} \\
4.4 & ANC P marker
    & \acchigh{90.1} \\
4.5 & ANC S--V order
    & \acchigh{90.3} \\
4.6 & ANC A--V order
    & \acchigh{93.1} \\
4.7 & ANC V--P order
    & \acchigh{92.9} \\
\midrule
5.1 & ANC external S marker
    & \accverylow{62.9} \\
5.2 & ANC external A marker
    & \accmid{82.7} \\
5.3 & ANC external P marker
    & \acchigh{91.4} \\
5.4 & ANC external genitive marker
    & \acclow{75.9} \\
\midrule
6.1 & ANC omitted S
    & \accveryhigh{98.5} \\
6.2 & ANC omitted A
    & \acclow{75.8} \\
6.3 & ANC omitted P
    & \accmid{83.9} \\
6.4 & ANC omitted A+P
    & \acchigh{90.5} \\
\bottomrule
\end{tabularx}
\end{minipage}

\caption{Targeted minimal-pair evaluation results.
Accuracies are averaged across the 96 configurations and three random seeds.
Darker shading indicates higher accuracy.}
\label{tab:e1-results-summary}
\end{table*}

We use targeted minimal-pair evaluations to test whether the models have learned the grammatical regularities encoded in their target grammars. The evaluation covers 34 phenomena in six groups, as shown in Table~\ref{tab:e1-results-summary}: basic finite clauses, noun phrases, clausal complements, ANC internal structure, ANC external marking, and ANC argument omission. The first three groups test the relevant basic grammar, while the remaining three focus on ANCs. External marking tests how ANCs are marked when functioning as matrix arguments or genitive possessors. Argument omission tests whether the models have learned that internal arguments may be omitted in ANCs but not in main or complement clauses in the target grammars.

For each phenomenon, we construct minimal pairs in each target language using templates and perturbation rules. The templates specify the structural environment, while the perturbation rules define how to modify a GOOD sentence to create its BAD counterpart. We first use GPT-4o to generate natural English source sentences matching the templates \citep{openai2024gpt4o}, then pass them through the pipeline described in Section~\ref{chap:corpus-generation} to generate target-language sentences, and finally apply the perturbation rules. We filter out invalid items, retaining 100 minimal pairs per phenomenon for each target language. A phenomenon may involve multiple templates or perturbation rules, and we balance the selected pairs across them. For example, in phenomenon 4.1 ANC Verb Form, one template requires a transitive ANC to function as the matrix object with only its internal P overtly expressed. The perturbation rule replaces the ANC predicate's nonfinite suffix \textit{-ing} with the finite suffix \textit{-s}. Under the same configuration as in Table~\ref{tab:pseudo-english-examples}, this yields the following pair:

\begin{exe}
\ex
\begin{xlist}
\ex GOOD: \textit{He likes readingca bookob}
\ex BAD: \textit{He likes readsca bookob}
\end{xlist}
\end{exe}

Each pair is scored with the model trained on the corpus of the corresponding target language. Accuracy is the proportion of pairs in which the model assigns a higher length-normalized log probability to the GOOD sentence. Some BAD sentences admit alternative grammatical analyses, and we report their parse rates in Appendix~\ref{app:evaluation-details}. We calculate mean accuracy across all 34 phenomena and across the 15 ANC-related phenomena to compare overall grammatical performance and ANC learning performance, respectively. Full details of minimal-pair construction are provided in Appendix~\ref{app:evaluation-details}.

\begin{figure*}[t]
    \centering
    \subfloat[Overall\label{fig:e1-accuracy-overall}]{%
        \includegraphics[width=0.49\textwidth]
        {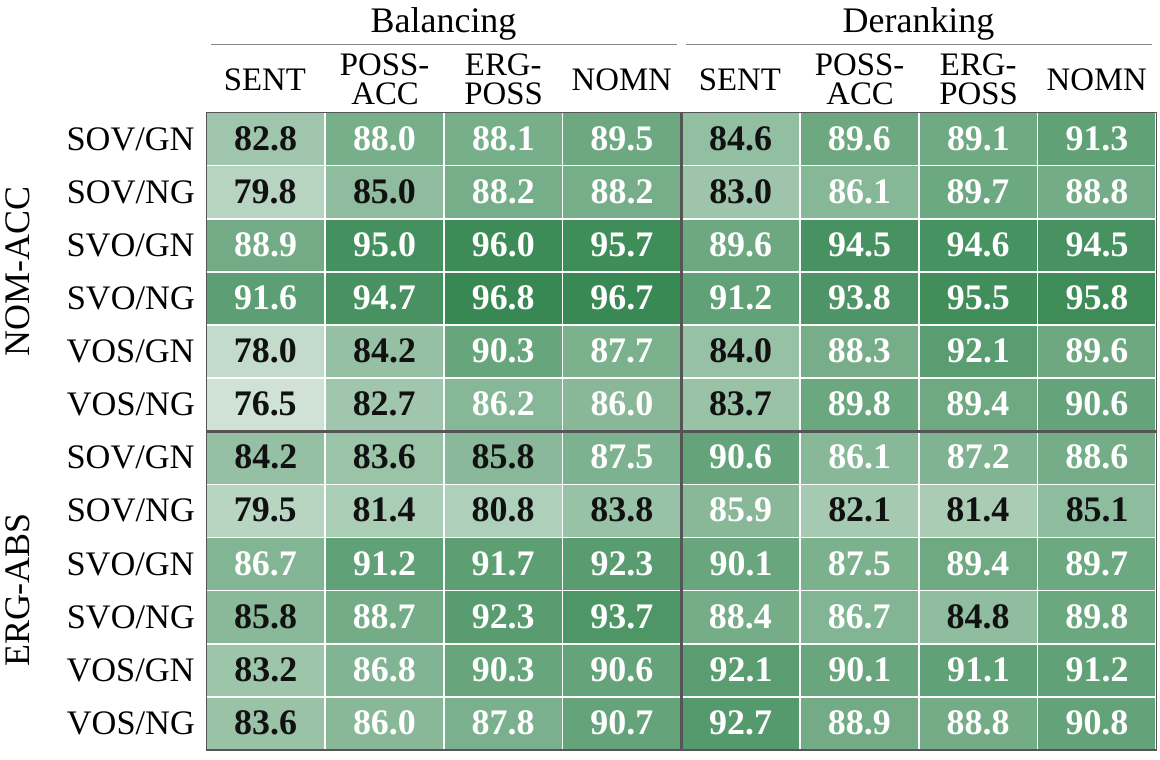}%
    }
    \hfill
    \subfloat[ANC-related\label{fig:e1-accuracy-anc}]{%
        \includegraphics[width=0.49\textwidth]
        {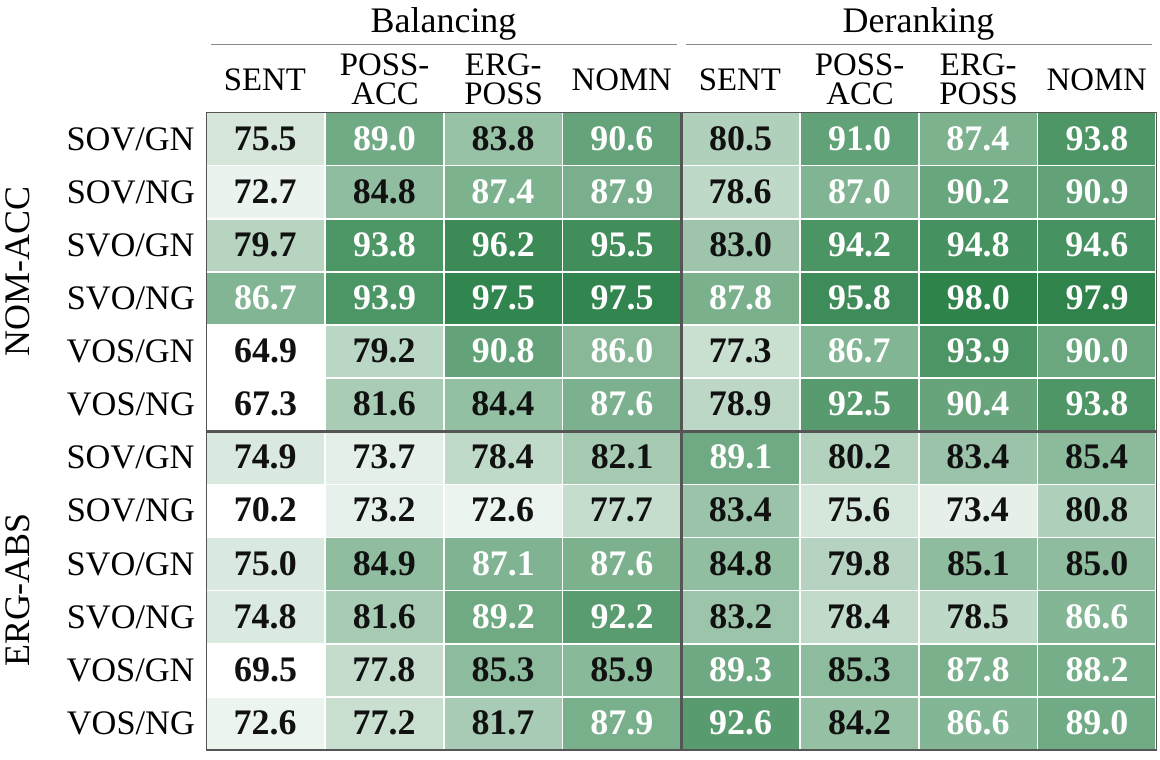}%
    }
    \caption{Mean accuracy across three seeds for each grammatical
    configuration, averaged over (a) all 34 phenomena and
    (b) the 15 ANC-related phenomena. Darker shading indicates higher accuracy.}
    \label{fig:e1-accuracy-overview}
\end{figure*}

Overall, the models learn many of the grammatical phenomena in the target grammars, with a mean accuracy of 88.4\% across all 34 phenomena, 96 configurations, and three random seeds. As Table~\ref{tab:e1-results-summary} shows, basic clause word order and finite verb form achieve near-perfect accuracy, while valency and some argument-marking phenomena are harder. Across grammatical configurations, Figure~\ref{fig:e1-accuracy-overview} shows that SVO + NOM-ACC configurations maintain relatively high accuracy in both the overall and ANC-related evaluations, while SENT performs relatively weakly under Balancing, especially in the ANC-related evaluation.

Further analysis reveals patterns consistent with prior model learning experiments. Overtly marked roles are generally easier to learn than zero-marked roles. For example, in phenomenon 1.5 Transitive A Marker, ERG-ABS performs better, while in phenomenon 1.6 Transitive P Marker, NOM-ACC performs better, consistent with the benefits of overt case marking reported by \citet{ravfogel-etal-2019-studying}. The models also appear sensitive to local surface patterns. In phenomenon 3.3 Clausal Complement A Marker, an incorrect genitive marker creates a local possessor--head sequence that models may prefer despite the sentence lacking a complete grammatical analysis, suggesting reliance on surface heuristics \citep{mccoy2019right}. Detailed results and error analyses are provided in Appendix~\ref{app:evaluation-details}.

\section{Discussion}
\label{chap:discussion}

In this section, we compare model learning preferences with typological distributions in natural languages, focusing on three parameter interactions. For each comparison, we report natural language counts and model accuracies averaged over the remaining grammatical parameters.

\subsection{Clause Word Order and Alignment}
\label{sec:clause-word-order-alignment}

The models' preference for alignment depends on clause word order. SVO and SOV favor NOM-ACC, whereas VOS favors ERG-ABS, with the largest difference between alignment systems occurring under SVO. These preferences align with the distributions in the natural language sample, where the configurations learned more easily by the models are also more frequent, although the VOS sample is small. Table~\ref{tab:discussion-wo-alignment-overlap} shows mean model accuracy and counts of natural languages with overt case marking for each combination \citep{dryer2013order,comrie2013alignment}.

\begin{table*}[t]
\centering
\small
\setlength{\tabcolsep}{3pt}
\begin{tabular}{@{}llllrrr@{}}
\toprule
Clause WO & Alignment & IV frame & TV frame & Overlap & Model acc. (\%) & Language count \\
\midrule
SVO & NOM-ACC & \textit{NP V-s} & \textit{NP V-s NP-ca} & 2 & 94.1 & 9 \\
SVO & ERG-ABS & \textit{NP V-s} & \textit{NP-ca V-s NP} & 0 & 89.3 & 1 \\
SOV & NOM-ACC & \textit{NP V-s} & \textit{NP NP-ca V-s} & 1 & 87.0 & 29 \\
SOV & ERG-ABS & \textit{NP V-s} & \textit{NP-ca NP V-s} & 0 & 84.6 & 16 \\
VOS & NOM-ACC & \textit{V-s NP} & \textit{V-s NP-ca NP} & 1 & 86.2 & 1 \\
VOS & ERG-ABS & \textit{V-s NP} & \textit{V-s NP NP-ca} & 2 & 89.0 & 2 \\
\bottomrule
\end{tabular}
\caption{Finite intransitive (IV) and transitive (TV) frames, mean accuracy across all 34 phenomena, and counts of natural languages by word order and alignment. Overlap counts consecutive units with identical morphological forms shared from the beginning of the two frames.}
\label{tab:discussion-wo-alignment-overlap}
\end{table*}

One possible explanation comes from the surface patterns shared by intransitive and transitive clauses. Under SVO + NOM-ACC, both clause types begin with a zero-marked argument followed by a finite predicate, sharing the sequence \textit{NP V-s}. Under ERG-ABS, overt marking on the transitive A breaks this shared beginning. VOS instead favors ERG-ABS, which allows the two clause types to share \textit{V-s NP}. SOV + NOM-ACC shares only the initial zero-marked argument, as the predicate occupies different positions in the two clause types. The overlap difference between alignment systems is therefore largest under SVO, where the models also show the strongest alignment preference. These results suggest that stable surface patterns shared across related clause types may facilitate learning of the corresponding configurations.

\subsection{Word Order and Complement System}
\label{sec:word-order-complement-system}

The effect of complement system also depends on word-order profile. Table~\ref{tab:discussion-wo-complement} shows that Balancing yields slightly higher overall accuracy under SVO, whereas Deranking performs better under SOV. \footnote{Among the verb-initial languages included in the complement-system data, most are VSO, whereas our experiments use VOS. We therefore focus on SVO and SOV.} This pattern partly aligns with natural language typology. In the sample of \citet{koptjevskaja1993nominalizations}, SVO languages mainly use Balancing, while Deranking is concentrated in SOV/GN.

A notable divergence is that the models also learn SVO + Deranking relatively well, although this combination is rare in the natural language sample. One possible explanation is that, under SVO, the matrix predicate separates the matrix subject from the arguments of the clausal complement, providing cues to argument roles and clause boundaries. Under SOV, by contrast, the matrix subject and embedded arguments are adjacent, followed by the embedded and matrix predicates, which may make the structures harder to distinguish. Deranking provides an additional cue to embedded predicates through nonfinite forms that differ from finite matrix predicates. In natural languages, if SVO order already distinguishes matrix clauses from their complements sufficiently well, the functional benefit of using Deranking may be smaller. In contrast, models under SVO may benefit from both word-order cues and the morphological cues supplied by Deranking, allowing them to learn this typologically uncommon combination relatively well.

\begin{table}[t]
\centering
\small
\setlength{\tabcolsep}{3pt}
\begin{tabular}{@{}lrrrr@{}}
\toprule
& \multicolumn{2}{c}{Model acc. (\%)}
& \multicolumn{2}{c}{Language count} \\
\cmidrule(lr){2-3}\cmidrule(lr){4-5}
Word-order profile & Bal. & Der. & Bal. & Der. \\
\midrule
SVO/GN & 92.2 & 91.2 & 5  & 1 \\
SVO/NG & 92.5 & 90.8 & 13 & 1 \\
SOV/GN & 86.2 & 88.4 & 7  & 24 \\
SOV/NG & 83.3 & 85.3 & 4  & 0 \\
\bottomrule
\end{tabular}
\caption{Mean accuracy across all 34 phenomena and natural language counts by word-order profile and complement system.}
\label{tab:discussion-wo-complement}
\label{tab:word-order-complement-system}
\end{table}

\subsection{Complement System and ANC Strategy}
\label{sec:complement-system-anc-strategy}

ANC strategy is closely associated with complement system in natural languages. As Table~\ref{tab:discussion-complement-anc-strategy} shows, SENT mainly occurs with Deranking, ERG-POSS and NOMN mainly occur with Balancing, and POSS-ACC is well represented under both systems \citep{koptjevskaja1993nominalizations}. The models reproduce these tendencies only partially. SENT performs better under Deranking on the ANC-related phenomena, consistent with the natural language distribution. However, NOMN performs well under both systems, including Deranking + NOMN, which is unattested in the sample. ERG-POSS also does not show the preference for Balancing found in natural languages.

This partial correspondence may reflect differences in how ANCs are used across languages. In Deranking languages, ANCs can serve as complements to utterance predicates and express content with relatively independent temporal reference and truth value. These uses may favor the retention of sentential argument coding and thus the SENT strategy. In Balancing languages, finite clausal complements can also serve these functions, while ANCs with process or manner readings may adopt encoding closer to that of ordinary noun phrases \citep{koptjevskaja1993nominalizations}. Our experiment uses shared semantic input across configurations and may therefore miss these functional pressures on ANC strategies.

\begin{table}[t]
\centering
\small
\setlength{\tabcolsep}{3pt}
\begin{tabular}{@{}lrrrr@{}}
\toprule
& \multicolumn{2}{c}{Model acc. (\%)}
& \multicolumn{2}{c}{Language count} \\
\cmidrule(lr){2-3}\cmidrule(lr){4-5}
ANC strategy & Bal. & Der. & Bal. & Der. \\
\midrule
SENT     & 73.7 & 84.0 & 3  & 14 \\
POSS-ACC & 82.6 & 85.9 & 13 & 12 \\
ERG-POSS & 86.2 & 87.5 & 22 & 3  \\
NOMN     & 88.2 & 89.7 & 14 & 0  \\
\bottomrule
\end{tabular}
\caption{Mean accuracy across the 15 ANC-related phenomena and natural language counts by ANC strategy and complement system. Languages may occur in more than one ANC strategy category.}
\label{tab:discussion-complement-anc-strategy}
\label{tab:complement-anc-strategy}
\end{table}

\section{Conclusion}
\label{chap:conclusion}

In this paper, we propose a grammar-engineered method for synthetic corpus construction that combines shared semantic input from natural corpora with computational grammars. This method retains some lexical and semantic properties of the original corpus while providing coordinated control over interrelated grammatical properties. We use 96 target languages to study LMs' learning of ANCs and related constructions and compare their learning preferences with natural language typology.

Our results reveal parallels between model preferences and natural language typology in basic clause configurations, but also divergences in configurations involving complement systems and ANC strategies. The models reproduce the observed word-order and alignment preferences relatively well, while also learning SVO + Deranking and Deranking + NOMN, combinations that are rare or unattested in the natural language sample.

One possible explanation is that existing grammatical cues influence subsequent encoding choices and their functional benefits during language change \citep{fedzechkina2020production,haspelmath2021coding}. Under SVO, word order already provides cues for distinguishing matrix clauses from their complements, so the additional benefit of Deranking may be limited, whereas the motivation to adopt Deranking may be stronger under SOV. ANCs develop on the basis of existing clausal and nominal structures, and their internal organization may reflect their uses in different complement systems \citep{koptjevskaja1993nominalizations}. In Deranking systems, a broader range of complement uses may favor the retention of SENT encoding. In Balancing systems, finite clausal complements serve some of these functions, while ANCs with process or manner readings may adopt more nominal encoding. These factors may help explain the distribution of representative configurations such as SVO + Balancing + NOMN and SOV/GN + Deranking + SENT in natural languages.

Overall, the parallels between model preferences and natural language typology in basic clause configurations suggest that general learning pressures may help shape these tendencies. More complex configurations involving complements and ANCs also depend on constructional functions and the division of labor among constructions, and their distribution may additionally reflect communicative and production demands \citep{kemp2018semantic,carr2020simplicity,fedzechkina2020production}. Thus, controlled learning experiments can capture some typological tendencies in basic clause configurations, while divergences in more complex configurations may reflect pressures from language use and change not fully captured by our experiments.

\section*{Limitations}
Our experiments use shared semantic input derived from an English corpus, retaining source-language properties such as the distinction between ANCs and clausal complements and their co-occurrence patterns with matrix predicates. These properties may differ across natural languages with different ANC strategies \citep{koptjevskaja1993nominalizations}. The resulting corpora may therefore make configurations closer to English in these respects, such as NOMN, easier for the models to learn. We have not tested this potential bias using other source languages or by controlling construction distributions.

Additionally, we construct target languages by crossing typological parameters, but changing one parameter may affect multiple cues across constructions. This design allows us to compare the effects of grammatical configurations and parameter interactions on learning, but does not isolate the contribution of each formal cue.

The experiments also do not examine how semantics and context influence the choice between ANCs and other complement forms, or how multiple ANC strategies coexist within a language. Future work could test how constructional functions and other pressures shape typological distributions, following the explanations proposed in Sections~\ref{chap:discussion} and~\ref{chap:conclusion}.

Finally, we evaluate a single model architecture under a fixed training setup, without comparisons across architectures, model scales, or training data sizes. All target languages use the original English-trained GPT-2 tokenizer, whose segmentation of synthetic word forms may affect the observed learning preferences. Future work could explore alternative models, training settings, and tokenization schemes to assess the robustness and generalizability of these learning preferences.

\bibliography{custom}

\appendix

\section{ANC Strategies}
\label{app:anc-strategies}

This appendix provides more detailed descriptions and natural-language examples of the four ANC strategies introduced in Section~\ref{sec:action-nominal-constructions}. A single language may have more than one ANC strategy, depending on the nominalizer, lexical class, construction type, or discourse context \citep{koptjevskaja1993nominalizations}.

Argument encoding refers to how an ANC signals the relation between the action nominal and its participants. This relation can be expressed through dependent marking, head marking, or word order \citep{koptjevskaja1993nominalizations,nichols1986head}. This study focuses primarily on how the major ANC strategies differ in dependent marking, where role marking appears on the participant NP. Table~\ref{tab:anc-types} illustrates the four strategies with simplified dependent-marking patterns. The examples below show how these patterns are realized in individual languages.

\subsection{Sentential ANCs}
\label{subsubsec:sentential-ancs}

Sentential (SENT) ANCs occupy the most clause-like end of the typology. Their defining property is that the arguments of the nominalized predicate retain the dependent-marking patterns found in finite clauses, rather than being recoded as genitive or possessive dependents inside an ordinary noun phrase \citep{koptjevskaja1993nominalizations}.

Korean is an example of the SENT strategy. In ordinary finite clauses, Korean marks subjects with nominative case, direct objects with accusative case, and indirect objects with dative case. In corresponding nominalized constructions, these arguments retain the same case marking. Because of this property, the boundary between ANCs and clausal complements can be relatively subtle in SENT languages. The examples below show that a Korean ANC can occur externally as a nominal constituent and bear accusative case as the object of the matrix predicate, while internally preserving clause-like dependent marking.

\begin{exe}
\ex Korean finite clause \citep[ex. 5.1]{koptjevskaja1993nominalizations}
\gll John i Mary eke c\textsuperscript{h}\ae k \u{u}l cu-\textschwa s'-ta \\
     John \textsc{nom} Mary \textsc{dat} book \textsc{acc} give-\textsc{past}-\textsc{fin} \\
\glt `John gave a book to Mary.'

\ex Korean SENT ANC \citep[ex. 5.6]{koptjevskaja1993nominalizations}
\gll [Ku ton \u{u}l Pak s\textschwa ns\ae\textipa{N} eke cu-k\textsci] l\u{u}l pal\ae-yo \\
     that money \textsc{acc} Pak mister \textsc{dat} give-\textsc{an} \textsc{acc} hope-\textsc{fin} \\
\glt `I hope you gave that money to Mr Pak.'
\end{exe}

Korean ANCs may also retain verbal properties such as tense, aspect, and adverbial modification. However, they do not preserve all properties of independent main clauses. For example, topic particles are more restricted in nominalized or subordinate contexts than in independent main clauses. Thus, although SENT ANCs are strongly clause-like in their internal argument encoding, they remain distinct from finite clauses in their external distribution and in some syntactic properties.

\subsection{Possessive-Accusative ANCs}
\label{subsubsec:possessive-accusative-ancs}

In Possessive-Accusative (POSS-ACC) ANCs, the S and A arguments of the nominalized predicate are recoded as genitive-like dependents, while the P argument retains the object marking found in finite clauses \citep{koptjevskaja1993nominalizations}.

Selkup is an example of the POSS-ACC strategy. In Selkup, objects in ordinary finite clauses may bear accusative marking. In corresponding ANCs, A or S takes genitive-like marking, while P retains accusative marking. In the example below, `my brother' is the A argument of the nominalized predicate and bears genitive marking, while `wife' is the P argument and retains accusative marking.

\begin{exe}
\ex Selkup POSS-ACC ANC
    \citep[ex.~6.10a]{koptjevskaja1993nominalizations}
\gll Mat a\v{s}\v{s}a t\textipa{E}nymy-s-ak
     [\textipa{tim\'{n}a}-n-y ima-p qo-pt\"a-$\emptyset$-ty] \\
     I:\textsc{nom} \textsc{neg}
     know-\textsc{past}-\textsc{1sg}
     brother-\textsc{gen}-\textsc{1sg.poss}
     wife-\textsc{acc}
     find-\textsc{an}-\textsc{nom}-\textsc{3sg.poss} \\
\glt `I did not know about my brother's finding a wife.'
\end{exe}

\subsection{Ergative-Possessive ANCs}
\label{subsubsec:ergative-possessive-ancs}

Ergative-Possessive (ERG-POSS) ANCs also stand between the sentential and nominal types. Unlike POSS-ACC ANCs, ERG-POSS groups S and P together by assigning them genitive-like marking, while A is marked separately, often by an oblique or instrumental expression \citep{koptjevskaja1993nominalizations}. ERG-POSS therefore creates an ergative-like alignment inside the ANC, contrasting with the accusative-like alignment in POSS-ACC ANCs.

Russian is an example of the ERG-POSS strategy. In Russian finite clauses, A is typically marked with nominative case and P with accusative case. In corresponding ANCs, however, P can be encoded as genitive, while A is expressed with instrumental case. The examples below contrast nominative A and accusative P in a finite clause with instrumental A and genitive P in the corresponding ANC.

\begin{exe}
\ex Russian finite clause \citep[ex. 1.8]{koptjevskaja1993nominalizations}
\gll Aleksandr zavoeva-l Egipet \\
     Alexander:\textsc{nom} conquer-\textsc{past.perf} Egypt:\textsc{acc} \\
\glt `Alexander conquered Egypt.'

\ex Russian ERG-POSS ANC \citep[ex. 1.9a]{koptjevskaja1993nominalizations}
\gll zavoeva-nij-a Egipt-a Aleksandr-om \\
     conquer-\textsc{an}-\textsc{gen} Egypt-\textsc{gen} Alexander-\textsc{instr} \\
\glt `the conquest of Egypt by Alexander'
\end{exe}

\subsection{Nominal ANCs}
\label{subsubsec:nominal-ancs}

Nominal (NOMN) ANCs are the type most closely aligned with ordinary nominal structure. In this type, S and A are encoded as genitive-like dependents, while P is assimilated to a dependent type available in ordinary noun phrases \citep{koptjevskaja1993nominalizations}.

NOMN can be divided into two subtypes: Double-Possessive and Possessive-Adnominal. In the Double-Possessive subtype, S, A, and P all receive genitive-like marking, and this pattern is especially natural in languages with multiple genitive positions. For example, English has two genitive positions, the prenominal \textit{'s}-genitive and the postnominal \textit{of}-genitive. When both A and P are overtly expressed, they can occupy these two positions, as in \textit{Peter's shooting of the prisoners}. In the Possessive-Adnominal subtype, S/A receive genitive-like marking, while P is expressed by an oblique or adpositional dependent available in ordinary noun phrases.

Since ANCs in actual use often omit arguments that are recoverable from context or irrelevant to the communicative goal, a single genitive slot is often sufficient to express the most relevant event participant. Thus, when only one argument is overtly expressed, the surface difference between Double-Possessive and Possessive-Adnominal ANCs is limited.

Finnish provides an example of the Double-Possessive subtype. Finnish can accommodate two formally identical but functionally distinct genitive dependents within the same noun phrase. When two genitive dependents occur together, the genitive farther from the head typically has a determiner-like function, specifying the reference of the head, while the genitive closer to the head has a descriptive or characterizing function. In ANCs, the former typically corresponds to A and the latter to P. In the example below, \textit{vanhempien} `parents' is the genitive dependent farther from the action nominal \textit{antaminen} `giving' and functions as A, while \textit{taloudellisen tuen} `economic support' is adjacent to the head and functions as P.

\begin{exe}
\ex Finnish NOMN ANC \citep[ex.~8.3]{koptjevskaja1993nominalizations}
\gll [Vanhempien taloudellisen tuen antaminen] on riippuvaista tuloista. \\
     parents:\textsc{gen} economic support:\textsc{gen} give:\textsc{an} is dependent incomes:\textsc{part} \\
\glt `Parents' giving of economic support is dependent on their incomes.'
\end{exe}

\section{Grammatical Specifications}
\label{app:experimental-configurations}

\begin{table*}[t]
\centering
\footnotesize
\begin{tabularx}{\textwidth}{@{}lXll@{}}
\toprule
Variable & Meaning & Possible values & Example configuration \\
\midrule
CLAUSE\_WO
& finite clause order of A/S, P, and V
& SOV, SVO, VOS
& SVO \\

FIN\_S\_MARK
& dependent marking of S in finite intransitive clauses
& 0
& 0 \\

FIN\_A\_MARK
& dependent marking of A in finite transitive clauses
& 0, \textit{-ca}
& 0 \\

FIN\_P\_MARK
& dependent marking of P in finite transitive clauses
& 0, \textit{-ca}
& \textit{-ca} \\

FIN\_V\_MARK
& verbal morphology on finite verbs
& \textit{-s}
& \textit{-s} \\

NP\_WO
& order of genitive dependent and nominal head
& GN, NG
& GN \\

GEN\_MARK
& marking of genitive dependents
& \textit{-ge}
& \textit{-ge} \\

COMP\_V\_MARK
& verbal morphology in clausal complements
& \textit{-s}, \textit{-ing}
& \textit{-s} \\

ANC\_IV\_ORDER
& internal order of S and the head in intransitive ANCs
& SV, VS
& SV \\

ANC\_TV\_ORDER
& internal order of A, P, and the head in transitive ANCs
& APV, AVP, PVA, VPA
& AVP \\

ANC\_S\_MARK
& dependent marking of S in ANCs
& 0, \textit{-ge}
& \textit{-ge} \\

ANC\_A\_MARK
& dependent marking of A in ANCs
& 0, \textit{-ca}, \textit{-ge}, \textit{-ob}
& \textit{-ge} \\

ANC\_P\_MARK
& dependent marking of P in ANCs
& 0, \textit{-ca}, \textit{-ge}, \textit{-ob}
& \textit{-ob} \\

ANC\_V\_MARK
& morphology on nominalized predicates
& \textit{-ing}
& \textit{-ing} \\
\bottomrule
\end{tabularx}
\caption{Possible morphosyntactic settings for the target languages and an example configuration with SVO, GN, NOM-ACC, Balancing, and NOMN. Here, 0 indicates zero marking.}
\label{tab:profile-variables}
\end{table*}

This appendix details how the five grammatical parameters introduced in Section~\ref{chap:experimental-parameters} determine the properties of each target language. Table~\ref{tab:profile-variables} summarizes the resulting grammatical settings and illustrates a configuration combining SVO, GN, NOM-ACC, Balancing, and NOMN. These settings include both properties determined by the parameter combinations and conventions shared across all target languages. The following subsections explain the mapping rules.

\subsection{Finite Clauses}
\label{subsec:finite-clauses}

Finite clauses are controlled by two main parameters: clause word order and alignment. Clause word order controls the relative order of the predicate and its core arguments in finite clauses. This study includes three values: SOV, SVO, and VOS. SOV and SVO are the two most frequent basic word orders cross-linguistically, while VOS serves as a representative verb-initial order \citep{dryer2013order}.

Alignment determines the dependent-marking pattern of A, S, and P in finite clauses. Nominative--accusative alignment (NOM-ACC) groups S and A together, while P is marked as the distinct argument. Ergative--absolutive alignment (ERG-ABS) groups S and P together, while A is marked as the distinct argument.

This study adopts a simplified marked-argument design: the two arguments that pattern together receive zero marking, while the distinct argument receives overt marking. This design reflects a common cross-linguistic case-marking pattern, in which the argument role distinguished from the other two core arguments is often the one that receives overt case marking \citep{comrie2013alignment}. For comparability, the same overt marker \textit{-ca} is used in both alignment systems. Table~\ref{tab:finite-alignment} shows how alignment determines finite clause argument marking.

\begin{table}[t]
\centering
\small
\begin{tabular}{@{}llll@{}}
\toprule
Alignment & S & A & P \\
\midrule
NOM-ACC & 0 & 0 & \textit{-ca} \\
ERG-ABS & 0 & \textit{-ca} & 0 \\
\bottomrule
\end{tabular}
\caption{Finite-clause argument marking under the two alignment systems.}
\label{tab:finite-alignment}
\end{table}

\subsection{Noun Phrases}
\label{subsec:noun-phrases}
Within noun phrases, this study models only one type of genitive
dependent relation. The only parameter manipulated here is NP word order, which determines the relative order of the possessor and the possessum. This parameter has two values: GN, in which the genitive dependent precedes the nominal head, and NG, in which the nominal head precedes the genitive dependent.

This parameter is needed because several ANC strategies recode some arguments of the nominalized predicate as genitive-like dependents. For simplicity, each nominal head is allowed only one direct genitive dependent. This means that two genitives cannot directly modify the same head, but a genitive dependent may itself contain another genitive relation.

\subsection{Complement System}
\label{subsec:complement-system}

The target grammars in this study contain three relevant verbal environments: main-clause verbs, verbs in clausal complements, and action nominal predicates. The form of these verbs is determined by the complement-system parameter. Typological studies of subordination often distinguish between balancing and deranking constructions. In balancing constructions, the subordinate predicate has a form similar to one that can occur in an independent declarative clause. In deranking constructions, the subordinate predicate appears in a reduced form that cannot normally occur as the predicate of an independent main clause \citep{stassen1985comparison,cristofaro2003subordination}.

Therefore, in this experiment, main-clause verbs are assigned the finite suffix \textit{-s}, while action nominal predicates are assigned the nonfinite nominalizing suffix \textit{-ing}. The complement-system parameter determines the form of verbs in clausal complements. In the balancing condition, complement verbs take the finite suffix \textit{-s}; in the deranking condition, complement verbs take the nonfinite suffix \textit{-ing}. Table~\ref{tab:complement-system} summarizes this mapping.

\begin{table}[t]
\centering
\small
\begin{tabular}{@{}llll@{}}
\toprule
System & Main clause & ANC & Complement \\
\midrule
Balancing & \textit{-s} & \textit{-ing} & \textit{-s} \\
Deranking & \textit{-s} & \textit{-ing} & \textit{-ing} \\
\bottomrule
\end{tabular}
\caption{Verb-form profiles under the two complement-system settings.}
\label{tab:complement-system}
\end{table}

\subsection{Clausal Complements}
\label{subsec:clausal-complements}

Apart from verb form, this study does not introduce additional distinctions for clausal complements. Argument marking in clausal complements follows the same alignment system as in main clauses, so A, S, and P in embedded clauses are realized according to the finite-clause marking pattern of the target language.

For simplicity, the target grammars do not include overt complementizers by default. Overt complementizers can mark a clause as the complement of a matrix predicate and provide an explicit cue to the boundary between the embedded clause and the main clause \citep{noonan2007complementation}. Since embedded clauses may occur without overt complementizers, this dimension is not manipulated in this study.

\subsection{ANC Case Marking}
\label{subsec:anc-case-marking}

The case marking of an ANC is controlled by the ANC strategy parameter. Natural languages may employ more than one ANC strategy, but each target language in this study is assigned exactly one ANC strategy for experimental control. The present experiment includes four ANC strategies: SENT, POSS-ACC, ERG-POSS, and NOMN.

Each strategy determines how the S, A, and P arguments of the nominalized predicate are marked. SENT preserves the argument-marking pattern of the corresponding finite clause. POSS-ACC realizes S and A as genitive-like dependents, while P retains the object marking of the finite clause. ERG-POSS realizes S and P as genitive-like dependents, whereas A receives oblique-like marking. NOMN realizes S and A as genitive-like dependents and P as an oblique-like dependent. Table~\ref{tab:anc-marking} summarizes these marking profiles.

\begin{table}[t]
\centering
\small
\begin{tabular}{@{}llll@{}}
\toprule
ANC strategy & S & A & P \\
\midrule
SENT     & $S_{\text{fin}}$ & $A_{\text{fin}}$ & $P_{\text{fin}}$ \\
POSS-ACC & \textit{-ge} & \textit{-ge} & $P_{\text{fin}}$ \\
ERG-POSS & \textit{-ge} & \textit{-ob} & \textit{-ge} \\
NOMN     & \textit{-ge} & \textit{-ge} & \textit{-ob} \\
\bottomrule
\end{tabular}
\caption{Dependent-marking profiles of the four ANC strategies. The subscript fin indicates the corresponding finite-clause marker.}
\label{tab:anc-marking}
\end{table}

\subsection{ANC Word Order}
\label{subsec:anc-word-order}

In this experiment, ANC-internal word order is derived from the interaction of clause word order, NP word order, and ANC strategy. This design is motivated by the observation that ANC-internal ordering can generally be related to ordering patterns already found in finite clauses and ordinary noun phrases \citep{koptjevskaja1993nominalizations}.

Concretely, SENT ANCs preserve the internal organization of the finite clause, so they directly retain finite-clause word order. The other three strategies combine structural properties of clauses and noun phrases, so their internal word order is determined according to the following three principles:

\begin{enumerate}
    \item The position of the possessive-like argument is determined by NP word order. Under GN order, this argument precedes the nominalized predicate; under NG order, it follows the nominalized predicate.
    \item The position of the non-possessive argument relative to the nominalized predicate follows the head--complement direction of the finite clause. In an OV language, this argument precedes the predicate; in a VO language, it follows the predicate.
    \item When the preceding two principles place A and P on the same side of the nominalized predicate, P is placed closer to the predicate than A. This ordering reflects the structural asymmetry between the two arguments, with P forming a constituent with the predicate before A is introduced outside that constituent.
\end{enumerate}

The ANC strategy determines which arguments occupy the possessive-like and non-possessive positions. In POSS-ACC and NOMN ANCs, S and A occupy the possessive-like position, and P occupies the non-possessive position. In ERG-POSS ANCs, S and P occupy the possessive-like position, while A occupies the non-possessive position.

Based on these principles, we obtain the ANC word-order patterns shown in Table~\ref{tab:anc-word-order}.

\begin{table}[t]
\centering
\footnotesize
\setlength{\tabcolsep}{2pt}
\begin{tabular}{@{}lllll@{}}
\toprule
Word Order & SENT & POSS-ACC & ERG-POSS & NOMN \\
\midrule
SOV + GN & SV/APV & SV/APV & SV/APV & SV/APV \\
SVO + GN & SV/AVP & SV/AVP & SV/PVA & SV/AVP \\
VOS + GN & VS/VPA & SV/AVP & SV/PVA & SV/AVP \\
SOV + NG & SV/APV & VS/PVA & VS/AVP & VS/PVA \\
SVO + NG & SV/AVP & VS/VPA & VS/VPA & VS/VPA \\
VOS + NG & VS/VPA & VS/VPA & VS/VPA & VS/VPA \\
\bottomrule
\end{tabular}
\caption{ANC-internal word orders determined by the interaction of clause word order, NP word order, and ANC strategy.}
\label{tab:anc-word-order}
\end{table}

\section{Corpus Construction}
\label{app:corpus-generation-details}

This appendix provides details of source-corpus processing, grammar construction, and candidate selection for the pipeline described in Section~\ref{chap:corpus-generation}. Background on the grammar-engineering framework and tools is provided at the end of this appendix in Section~\ref{app:grammar-engineering}.

\subsection{English Parsing and Pseudo-English}
\label{sec:english-to-pseudo}

In the English parsing stage, the pipeline uses spaCy dependency parses \citep{spacy2020} to extract the predicate--argument structures that can be expressed by the current grammar fragment. We do not directly parse the raw English corpus with an ACE-based HPSG grammar such as the English Resource Grammar \citep{flickinger2014erg}, because such grammars provide rich and fine-grained semantic analyses of English, which would increase the complexity of aligning the resulting MRS representations with the target grammars used in this experiment. Using dependency parses at this stage also allows the pipeline to filter out structures that are not modeled, including passives, adjectival predication, and clauses with multiple object candidates.

The experiment retains four constructions: intransitive clauses, transitive clauses, clausal-complement constructions, and nominal-predicate copular clauses. For each retained sentence, the system records the predicate and its core arguments. For clausal complements, the system records the matrix predicate and recursively analyzes the embedded clause. Both \textit{xcomp} and \textit{ccomp} dependencies are included. In the current pipeline, an \textit{xcomp} is treated as a subject-control structure, with the matrix subject supplying the subject of the embedded predicate.

The system also records three types of nominal modifiers for all nouns: possessives, \textit{of}-phrases, and \textit{by}-phrases. These records are used later in the treatment of action nominal constructions. They provide cues to the possible argument structure of an action nominal: possessives and \textit{by}-phrases can correspond to agent-like arguments, while \textit{of}-phrases can correspond to patient-like arguments.

Parsed English is then linearized as pseudo-English. Pseudo-English uses a fixed, English-like surface format to express the extracted predicate--argument structures. Intransitive clauses are realized as S--V, transitive clauses as A--V--P, and clausal-complement constructions are realized with an overt complementizer. Prepositions introducing PP objects and verb particles are incorporated into the predicate form. For example, \textit{look for} is represented as \textit{lookfor}. Further details on the pseudo-English grammar are given in Section~\ref{sec:choices-files-grammars}.

At the same time, the system constructs the shared lexicon. The lexicon contains nouns, intransitive verbs (\textit{iv}), transitive verbs (\textit{tv}), clausal-complement verbs (\textit{cv}), and action-nominal candidates. Predicates extracted from nominal-predicate copular clauses (\textit{cop\_n}) are incorporated into the transitive-verb class in the lexicon, since the Grammar Matrix configuration does not directly support nominal complements in copular constructions. Lexical items preserve forms extracted from the English corpus, retaining some of the lexical distribution and word-form regularities of the source corpus.

Action-nominal candidates are identified heuristically. The system first searches for nouns with common nominalizing suffixes, including \textit{-ation}, \textit{-tion}, \textit{-sion}, \textit{-ment}, \textit{-ance}, \textit{-ence}, \textit{-ure}, \textit{-al}, and \textit{-ing}. The suffix is then removed, and the remaining stem is compared with verbs in the verb lexicon. A noun is treated as a possible action nominal only when the noun stem and a verb share a sufficiently similar prefix, with the threshold set to 0.70. Nouns that are not identified as action nominals can still enter the lexicon as ordinary nouns, but their \textit{of}- and \textit{by}-modifiers are not interpreted as ANC arguments.

The current system uses only \textit{iv} and \textit{tv} verbs as nominalization sources. This is because \textit{cv} verbs must retain their clausal complement after nominalization. In the POSS-ACC strategy, this complement occupies an object-like position and requires the appropriate dependent marking. The current Grammar Matrix setup does not reliably assign the required case marking to such nominalized clausal complements. To preserve parallelism across the generated corpora, \textit{cv} verbs are therefore excluded from the action-nominal source set.

Table~\ref{tab:english-parsing-summary} summarizes the main outputs of English parsing and lexicon construction. Table~\ref{tab:pseudo-english-examples} gives examples of the pseudo-English realization.

\begin{table*}[t]
\centering
\footnotesize
\begin{tabular}{lrlr}
\toprule
\multicolumn{2}{c}{English Parsing} & \multicolumn{2}{c}{Shared Lexicon} \\
\cmidrule(lr){1-2}\cmidrule(lr){3-4}
Category & Count & Category & Count \\
\midrule
Parsed sentences & 12,441,836 & Nouns & 225,939 \\
Retained sentences & 4,095,236 & Intransitive verbs & 12,252 \\
\textit{iv} constructions & 883,158 & Transitive verbs & 35,312 \\
\textit{tv} constructions & 1,812,451 & Clausal-complement verbs & 5,220 \\
\textit{cv} constructions & 674,525 & Action-nominal candidates & 6,298 \\
\textit{cop\_n} constructions & 725,102 & & \\
\midrule
\multicolumn{2}{c}{ANC External Position} & \multicolumn{2}{c}{ANC Internal Arguments} \\
\cmidrule(lr){1-2}\cmidrule(lr){3-4}
Category & Count & Category & Count \\
\midrule
S & 16,084 & No argument & 220,501 \\
A & 43,155 & One S/A-like argument & 24,712 \\
P & 156,671 & One P-like argument & 18,264 \\
Copular complement & 45,675 & Two arguments & 719 \\
Genitive possessor & 643 & & \\
ANC-internal argument & 1,968 & & \\
\bottomrule
\end{tabular}
\caption{Summary of English parsing, shared-lexicon construction, and action-nominal distribution for the training split.}
\label{tab:english-parsing-summary}
\end{table*}

\subsection{Grammar Construction and Patching}
\label{app:grammar-construction}

\subsubsection{Choices Files and Grammars}
\label{sec:choices-files-grammars}

On the grammar-construction side, the pipeline converts the typological parameters defined in Section~\ref{chap:experimental-parameters} into HPSG grammars that can be used for parsing and generation with ACE \citep{packard2018ace}. Each grammar is first specified by a Grammar Matrix choices file. A choices file is the input format used by the Grammar Matrix to record the main grammatical settings of a language, such as basic word order, case system, genitive order, clausal-complement system, nominalization strategy, and the relevant morphological rules.

For the 96 target grammars, the choices files are generated automatically from the parameter grid. Each choices file corresponds to one parameter combination, and its language ID records the parameter values for that combination. In addition to these explicit parameters, the script derives more specific grammar settings according to the rules detailed in Appendix~\ref{app:experimental-configurations}, including ANC-internal word order, verb form, and case marking. Grammatical dimensions that are not part of the experiment are kept inactive.

The pipeline then injects the shared lexicon extracted from the source corpus in Section~\ref{sec:english-to-pseudo} into the grammars. After the Grammar Matrix generates an initial Type Description Language (TDL) grammar from a choices file, the pipeline replaces the lexical entries with the shared lexicon. Nouns are written as common-noun lexical entries, while intransitive verbs, transitive verbs, and clausal-complement verbs are written into the corresponding verbal lexical types. Predicates extracted from \textit{cop\_n} constructions are added to the transitive-verb class.

In addition to the 96 target grammars, the pipeline builds a pseudo-English grammar for parsing the pseudo-English corpus from Section~\ref{sec:english-to-pseudo} into MRS. This grammar uses a separate pseudo-English choices file, which is different from the target-language choices files generated from the typological parameter grid.

The pseudo-English grammar differs from the target grammars in several implementation details. It uses an overt complementizer \textit{that}, allowing clausal-complement structures to be parsed more reliably. It also uses the dedicated nominalization marker \textit{-nmz}. Since \textit{-nmz} does not occur in ordinary English word forms, it prevents nominalized predicates in pseudo-English from being confused with common nouns. The target grammars instead use \textit{-ing} as the nominalization suffix, so that nominalized forms in the generated corpora retain word-form cues similar to English nominalization morphology.

Finally, the TDL grammars generated by the Grammar Matrix are compiled into ACE grammar images. The compiled pseudo-English grammar is used to parse pseudo-English into MRS, while the target grammars are used to generate the 96 target-language corpora from MRS. The full set of choices files is included in the project repository.

\subsubsection{Grammar Patching}
\label{sec:grammar-patching}

The ANC library in the Grammar Matrix has a limitation in word-order control. In this system, ANC-internal word order must be selected as a single word-order profile, such as SOV or SVO \citep{Ruditsky2025}. However, as discussed in Section~\ref{subsec:anc-word-order}, under some parameter combinations, the S and A arguments within ANCs may appear on opposite sides of the nominalized predicate. For example, SVO/GN + ERG-POSS yields SV/PVA, and SOV/NG + ERG-POSS yields VS/AVP. These split patterns cannot be naturally represented by a single Grammar Matrix ANC-WO setting.

To address this problem, the pipeline first sets a single ANC word-order value in the choices file and uses it to generate the initial TDL grammar. In split patterns, this setting is used only to determine the position of the possessive-like argument relative to the nominalized predicate. For example, in split configurations with GN order, it is set to SOV.

The pipeline then changes the \texttt{HEAD.ANC-WO} value of the three non-sentential transitive ANC lexical rules from \texttt{+} to \texttt{-}, allowing the non-possessive argument to use the ordinary complement-head path. As a result, the possessive-like argument is still ordered through the specifier path according to NP order, while the non-possessive argument is ordered through the ordinary complement-head path according to the finite-clause direction. This makes it possible for the generated grammars to realize the split orders defined in the experiment.

However, this treatment introduces overgeneration. Since the non-possessive argument of a transitive ANC is allowed to use the ordinary complement-head path, both A-first and P-first derivations may be available when both ANC-internal arguments occur on the same side of the nominalized predicate. As a result, a single MRS can generate multiple surface candidates that differ only in the relative order of A and P. For example, under an SOV, GN, NOM-ACC, Balancing, ERG-POSS configuration, one MRS generates the following two outputs:

\begin{quote}
\begin{verbatim}
1. I herob sermonge tellingca hears
2. I sermonge herob tellingca hears
\end{verbatim}
\end{quote}

For this reason, the pipeline applies post-generation selection. If the candidates differ only in the relative order of the A and P arguments, it selects the candidate that matches the language's effective ANC transitive order. In the example above, the parameter configuration gives the target ANC order APV. The markers identify \textit{herob} as A and \textit{sermonge} as P, so the system keeps the first output.

\subsection{MRS Parsing and Surface Selection}
\label{sec:mrs-bank-selection}

After pseudo-English has been generated, it is parsed with the pseudo-English grammar to produce the MRS bank. The MRS bank then serves as the shared semantic input to the 96 target grammars. Each target grammar realizes the same MRSs according to its own morphosyntactic settings, producing one target-language corpus. Table~\ref{tab:pseudo-english-examples} shows the realizations produced by the example grammar profile in Table~\ref{tab:profile-variables}.

During MRS construction, the pipeline applies a small amount of MRS normalization. For example, the Grammar Matrix uses \texttt{COG-ST} to represent the discourse status of nominal referents: \texttt{uniq-id} means that the referent is uniquely identifiable, while \texttt{in-foc} means that the referent is in focus. The pipeline replaces these more specific subtypes with the more general value \texttt{cog-st} to keep the MRS compatible with the different ANC strategies during generation.

After generation, the pipeline gathers all outputs for the same input item and selects one final sentence. Overgeneration can arise at two stages in the pipeline: pseudo-English parsing can produce multiple MRSs for the same pseudo-English item, and target grammar generation can produce multiple surface candidates from the same MRS.

The first main source is ANC-internal order overgeneration, discussed in Section~\ref{sec:grammar-patching}. The second comes from pseudo-English parsing. The pseudo-English grammar does not always produce a unique MRS for each input item. A common case is that the same stem occurs in both the \textit{iv} and \textit{tv} lexicons, so an action nominal with only one overt argument can receive two analyses. The argument may be analyzed as the S of an intransitive nominalization, or as the remaining A or P of a transitive nominalization, with the other argument omitted. When these two MRSs are generated by a target grammar, they may surface with different argument markers. For example, under the ERG-POSS strategy:

\begin{quote}
\begin{verbatim}
Theirob suffering seems
Theirge suffering seems
\end{verbatim}
\end{quote}

Here, \textit{theirob} corresponds to the A-like analysis, while \textit{theirge} corresponds to the S-like analysis. In such cases, the pipeline keeps the S-like candidate by default, that is, the output with \texttt{ANC\_S\_MARK}.

In addition, a small number of overgeneration cases cannot be resolved by the rules above. These usually come from accidental attachment ambiguity. For example, the generator may produce both \textit{the mention seems that the mention of the rejection sinks into the mind} and \textit{the mention of the rejection seems that the mention sinks into the mind}, which differ in whether \textit{of the rejection} attaches to the matrix nominal or the embedded nominal. For these cases, the pipeline chooses one output with a fixed random seed, so that the result remains reproducible.

\subsection{Grammar Engineering Background and Tools}
\label{app:grammar-engineering}

This subsection introduces the grammatical framework, semantic representation, and tools used in the pipeline described above.

\subsubsection{Head-Driven Phrase Structure Grammar}
\label{subsec:head-driven-phrase-structure-grammar}
Head-Driven Phrase Structure Grammar (HPSG) is a lexicalist, constraint-based grammatical framework \citep{pollard1994head,sag2003syntactic}. In HPSG, words and phrases are represented as typed feature structures. These structures encode not only syntactic category, but also multiple layers of linguistic information such as valence requirements, morphosyntactic features, and semantic contribution. Compared with many more abstract formal grammars, HPSG representations include syntactic constraints, lexical constraints, and a semantic interface, making the framework well suited for describing interactions among multiple grammatical modules in natural language.

\subsubsection{Minimal Recursion Semantics}
\label{subsec:minimal-recursion-semantics}

\begin{figure*}[t]
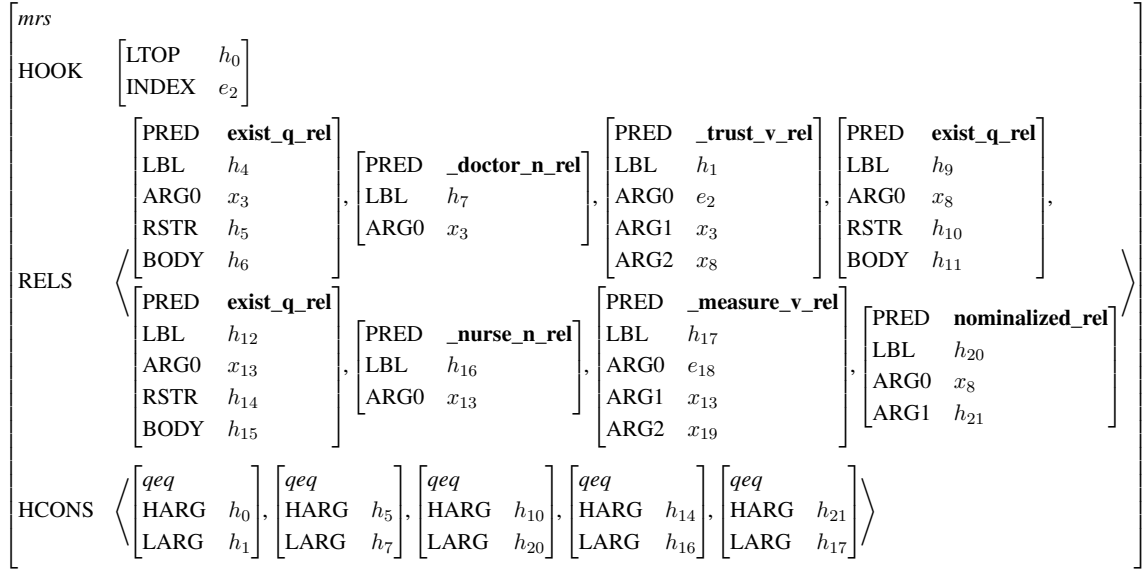

\centering
\resizebox{0.95\textwidth}{!}{
\begin{avm}
\[
\textit{mrs} \\
HOOK &
\[
LTOP  & $h_0$ \\
INDEX & $e_2$
\]
\\
RELS &
\<
\[
PRED & \textbf{exist\_q\_rel} \\
LBL  & $h_4$ \\
ARG0 & $x_3$ \\
RSTR & $h_5$ \\
BODY & $h_6$
\],
\[
PRED & \textbf{\_doctor\_n\_rel} \\
LBL  & $h_7$ \\
ARG0 & $x_3$
\],
\[
PRED & \textbf{\_trust\_v\_rel} \\
LBL  & $h_1$ \\
ARG0 & $e_2$ \\
ARG1 & $x_3$ \\
ARG2 & $x_8$
\],
\[
PRED & \textbf{exist\_q\_rel} \\
LBL  & $h_9$ \\
ARG0 & $x_8$ \\
RSTR & $h_{10}$ \\
BODY & $h_{11}$
\], \\
\[
PRED & \textbf{exist\_q\_rel} \\
LBL  & $h_{12}$ \\
ARG0 & $x_{13}$ \\
RSTR & $h_{14}$ \\
BODY & $h_{15}$
\],
\[
PRED & \textbf{\_nurse\_n\_rel} \\
LBL  & $h_{16}$ \\
ARG0 & $x_{13}$
\],
\[
PRED & \textbf{\_measure\_v\_rel} \\
LBL  & $h_{17}$ \\
ARG0 & $e_{18}$ \\
ARG1 & $x_{13}$ \\
ARG2 & $x_{19}$
\],
\[
PRED & \textbf{nominalized\_rel} \\
LBL  & $h_{20}$ \\
ARG0 & $x_8$ \\
ARG1 & $h_{21}$
\]
\>
\\
HCONS &
\<
\[
\textit{qeq} \\
HARG & $h_0$ \\
LARG & $h_1$
\],
\[
\textit{qeq} \\
HARG & $h_5$ \\
LARG & $h_7$
\],
\[
\textit{qeq} \\
HARG & $h_{10}$ \\
LARG & $h_{20}$
\],
\[
\textit{qeq} \\
HARG & $h_{14}$ \\
LARG & $h_{16}$
\],
\[
\textit{qeq} \\
HARG & $h_{21}$ \\
LARG & $h_{17}$
\]
\>
\]
\end{avm}
}
\caption{MRS representation of \textit{The doctor trusted the nurse's measurement}.}
\label{fig:mrs-example}
\end{figure*}

Minimal Recursion Semantics (MRS) is a semantic representation commonly used in HPSG-based grammar engineering \citep{copestake2005minimal}. MRS represents sentence meaning using elementary predications, argument relations, and scope constraints, thereby encoding predicate-argument structure while allowing certain scope relations to remain underspecified. In this work, MRS serves as a shared semantic interface across grammars. The same MRS can represent the same event structure, participant relations, and basic semantic content, while different target grammars can realize this semantic input in different surface forms. Figure~\ref{fig:mrs-example} provides an example MRS for a sentence containing an ANC.

\subsubsection{The Grammar Matrix}
\label{subsec:grammar-matrix}
The Grammar Matrix is an HPSG-based grammar customization system designed to support the rapid development of typologically diverse computational grammars \citep{bender-etal-2002-grammar,bender-etal-2010-grammar}. Researchers specify typological properties of the target language in a choices file, such as basic word order, case marking, tense/aspect/mood, and inflectional morphology. The system then generates a set of HPSG grammar source files, which can subsequently be compiled into a grammar for parsing and generation.

\subsubsection{The ANC Library}
\label{subsec:anc-library-grammar-matrix}

\citet{Ruditsky2025} builds on the nominalized-clause library in the Grammar Matrix \citep{howell-etal-2018-nominalized} and extends it with an implementation specifically designed for ANCs. Users can define one or more nominalization strategies for a language, and for each strategy, specify the argument-marking pattern, internal word order, determiner requirements, and permitted types of modification. Each strategy is then associated with the relevant nominalizing morphology, which derives an action nominal from a verbal lexical entry.

During this derivation, the syntactic category of the lexical item changes from verb to noun, while the event semantics and argument relations of the source verb are preserved. The system can therefore derive corresponding action nominal forms from verbal lexical entries through nominalization rules, rather than requiring a separate lexical entry for every nominalized form. The nominalization rules also adjust the valence of the derived item so that its participants can be realized according to the requirements of the selected ANC strategy.

The library allows action nominals to exhibit both nominal and verbal properties. A nominalized form may take nominal morphology, including case, number, and possessive marking, while retaining verbal morphology where the language permits it. The internal word order of an ANC can be specified independently of the basic word order of finite clauses, and its modification pattern can be configured to allow adjectival modification, adverbial modification, or both.

With respect to argument realization, the core arguments of an ANC are optional by default. If a language requires a particular argument to be overtly expressed, that argument can instead be specified as obligatory. The system also treats transitive and intransitive nominalizations separately, allowing them either to share a single strategy or to be subject to different constraints.

\subsubsection{DELPH-IN Tools}
\label{subsec:delph-in-tools}

ACE and PyDelphin are two DELPH-IN tools used in this study. ACE is a parsing and generation engine for compiled HPSG grammars: given a compiled grammar, ACE can parse surface strings into MRS representations and generate surface strings from MRS inputs \citep{packard2018ace}. PyDelphin provides an interface for working with DELPH-IN representations and for calling ACE within a Python experimental pipeline \citep{goodman2019python}. The same set of MRS inputs is passed to different target grammars, and ACE generates the corresponding morphosyntactic realizations.

\section{Training Setup}
\label{app:model-training}

\subsection{Training Corpus}

The source corpus is split into train, dev, and test sets with a 90/5/5 ratio. The train split is used to build the shared lexicon and is then passed through the corpus-generation pipeline described in Section~\ref{chap:corpus-generation}, producing 96 target-language training corpora of approximately 4.1 million sentences each, with sentence counts differing by fewer than 40 across languages. The dev and test splits are processed using the lexicon and grammars built from the train split, and do not add new lexical items to the shared lexicon. Each model is trained only on the corpus for its corresponding target language.

\subsection{Tokenization}

This experiment uses the GPT-2 byte-level BPE tokenizer \citep{radford2019gpt2}, shared across all target languages. In the target grammars, morphological markers are designed to attach directly to lexical stems. Because these stems are themselves derived from English, this design keeps the input closer to the surface conditions under which models learn raw English. The fixed tokenizer represents these novel surface forms as sequences of subword units. Inspection of the tokenizer output indicates that morphological markers such as \textit{-ca}, \textit{-ge}, and \textit{-ing} can remain identifiable as subword-level cues, although token boundaries need not coincide with morpheme boundaries.

After tokenization, the 96 training corpora contain approximately 24.2--27.6 million tokens each, with a mean of approximately 25.4 million. Some of this variation appears to reflect how the GPT-2 tokenizer segments different surface forms. Although sentence counts are closely matched across languages, differences in tokenized corpus size mean that a fixed training schedule covers different numbers of corpus-equivalent epochs.

\subsection{Training Configuration}

The experiment uses the GPT-2-small configuration with approximately 124 million parameters \citep{radford2019gpt2}. Model parameters are randomly initialized before training, so all models are trained from scratch. For each target language, we train three models with different random seeds. The three runs use the same target-language corpus, tokenizer, architecture, and training hyperparameters, differing only in random initialization and data shuffling. All models are trained for 70,000 steps. Because the tokenized corpus sizes vary slightly across languages, the fixed training schedule corresponds to approximately 2.59--2.96 corpus-equivalent epochs. Each model required approximately 1.5 hours of training on a single NVIDIA GeForce RTX 4060 GPU. The training configuration is shown in Table~\ref{tab:training-config}.

\begin{table}[htbp]
\centering
\small
\begin{tabular}{ll}
\toprule
Parameter & Value \\
\midrule
Model configuration & GPT-2-small \\
Tokenizer & GPT-2 tokenizer \\
Initialization & Random \\
Seeds & 42, 43, 44 \\
Block size & 32 \\
Training steps & 70,000 \\
Train batch size & 16 \\
Eval batch size & 16 \\
Gradient accumulation & 2 \\
Learning rate & $5\times10^{-4}$ \\
Warmup steps & 590 \\
Weight decay & 0.01 \\
Logging steps & 100 \\
Evaluation steps & 5,000 \\
Save steps & 5,000 \\
Save total limit & 1 \\
Saved model & Final-step model \\
\bottomrule
\end{tabular}
\caption{Training configuration for each target-language model.}
\label{tab:training-config}
\end{table}

\newcounter{ancevalphenomenon}
\newcommand{\ancevalphenomenon}[3]{%
    \paragraph{#1 #2}%
    \renewcommand{\theancevalphenomenon}{\thesubsection\space(#1)}%
    \refstepcounter{ancevalphenomenon}\label{#3}%
}
\newcounter{ancevalprompt}

\section{Evaluation Materials and Results}
\label{app:evaluation-details}

This appendix provides construction details and result summaries for the targeted minimal-pair evaluation described in Section~\ref{chap:evaluation}. For each phenomenon, we report the target phenomenon, source template, perturbation rule, representative minimal-pair example, results summary, and heatmap. The minimal-pair examples use the parameter configuration closest to English, combining SVO, GN, NOM-ACC, Balancing, and NOMN, as detailed in Table~\ref{tab:profile-variables}.

\subsection{Evaluation Construction Details}
\label{app:evaluation-construction}
\label{sec:minimal-pair-construction}

\subsubsection{Source Sentence Generation}

For each phenomenon to be tested, we define two components: source templates and a perturbation rule. The source templates specify the structural environment in which the phenomenon must occur. For example, when testing the finite form of an intransitive verb, the source template is an intransitive clause containing only S and V. The perturbation rule specifies how a GOOD sentence should be modified to construct a corresponding BAD sentence for the target grammatical phenomenon. When a phenomenon uses multiple source templates or multiple perturbation variants, these variants are balanced whenever possible.

After defining the source templates, we use GPT-4o with a controlled prompt to generate natural English source sentences \citep{openai2024gpt4o}. These source sentences are then passed through the same semantic generation pipeline described in Section~\ref{chap:corpus-generation}. They are first simplified into pseudo-English, then parsed into MRS, and finally generated by each of the 96 target grammars. Items are filtered out if they cannot be extracted, parsed, or generated, if they contain words outside the shared lexicon, or if the generated output cannot be reliably matched to the target template.

BAD sentences are constructed from the GOOD sentences using the perturbation rule. Different types of phenomena use different perturbation strategies. For word-order phenomena, the perturbation swaps the linear order of the relevant constituents. For morphological phenomena, the perturbation replaces the suffix with one that is ungrammatical in that position. Special phenomena such as valency and ANC omission use dedicated perturbation rules.

For each phenomenon, we select 100 GOOD--BAD minimal pairs for each target language from sentences generated from a shared pool of English source sentences. Box~\ref{box:evaluation-source-prompt} shows the prompt template for generating English source sentences. All phenomena use the same general instructions, with a specific construction block. The full implementation of the templates and perturbation rules is available in the project repository.

\refstepcounter{ancevalprompt}
\label{box:evaluation-source-prompt}
\begin{tcolorbox}[
  breakable,
  colback=gray!5,
  colframe=gray!60,
  boxrule=0.4pt,
  arc=1mm,
  left=5pt,
  right=5pt,
  top=5pt,
  bottom=5pt,
  title={Box~\theancevalprompt. Source sentence generation prompt}
]
\footnotesize
\ttfamily
\setlength{\parindent}{0pt}

You are helping generate English source sentences for a controlled grammar evaluation. Generate natural English sentences that instantiate the target construction described below. The sentences should be grammatical, semantically plausible, and close to ordinary corpus English. Avoid producing a mechanical list with repeated lexical items or repeated sentence frames.
\par\medskip

General requirements:\par
- Each sentence must clearly instantiate the target construction.\par
- Avoid structures that are irrelevant to the target construction and may interfere with extraction or evaluation.\par
- Use ordinary, high-frequency vocabulary whenever possible.\par
- Vary lexical items and noun phrase types where appropriate.\par
- Return one sentence per line, with no numbering, explanation, or section headings.
\par\medskip

Target construction (example: 1.1 Intransitive finite verb form):\par
- Generate 160 ordinary finite intransitive clauses.\par
- Each sentence must contain one overt subject and one finite intransitive verb.\par
- The target verb should be a lexical intransitive verb; do not make an auxiliary, modal, or aspectual marker the target.\par
- Generate approximately half of the sentences with simple subjects and half with possessive/genitive subjects.\par
- Possessive/genitive subjects should contain one possessor and one head noun.\par
- Avoid transitive clauses, clausal complements, passives, coordination, relative clauses, nominalizations, idioms, and unstable argument structures.\par
- Use ordinary, high-frequency intransitive verbs.
\par\medskip

Output format:\par
- Return one sentence per line.\par
- Do not include numbering, explanations, or section headings.\par
\end{tcolorbox}

\subsubsection{Valency Phenomena}
\label{subsec:valency-phenomena}

For the valency phenomena, the perturbation replaces the target verb with another verb whose valency does not match the environment. We first construct verb pairs for these phenomena. For 1.9 and 1.10, we extract intransitive/transitive verb pairs from the \texttt{intransitive.\allowbreak{}jsonl} and \texttt{transitive.\allowbreak{}jsonl} paradigms in BLiMP \citep{warstadt-etal-2020-blimp-benchmark}. For 3.6, we use the MegaAcceptability dataset \citep{white_rawlins_2020} to construct pairs of clausal-complement verbs and ordinary transitive verbs. This dataset provides acceptability ratings for English verbs in clausal complement frames and transitive frames.

Next, we use prompts to generate new source sentences for the target verbs, and then use source templates to filter out items that do not match the target structure or contain words outside the shared lexicon. We then replace the target verb in the GOOD sentence with the mismatched verb to form the BAD sentence.

Since many English verbs can occur in multiple valency frames, a grammar may still parse the BAD sentence if the relevant verb--valency combination is attested in the training set and therefore recorded in the lexicon.

\subsubsection{Case Marking between Clausal Complements and ANCs}
\label{subsec:case-marking-clausal-complements-ancs}

In the clausal-complement and ANC-internal case marking tests, the perturbation is mainly constructed by replacing markers across constructions. The embedded argument marker in a clausal complement is replaced with the corresponding ANC marker, while the ANC-internal argument marker is replaced with the corresponding marker in a non-ANC clause. When the two markers are identical on the surface in a particular configuration, we instead use another marker that is ungrammatical in that position.

It is important to note that identical surface marking does not mean that the two constructions are identical. Under the Deranking + SENT strategy, the verb form and argument marking inside an ANC are the same as those in a clausal complement. However, the ANC as a whole can occur in argument position and can receive external case or genitive marking, whereas a clausal complement can only occur as the complement of a clausal-complement verb.

\subsubsection{BAD Sentence Parsability}
\label{subsec:bad-sentence-parsability}

In minimal-pair construction, the BAD sentence is created by an external perturbation rule and therefore should not be accepted by the grammar under the target structure. However, some BAD sentences can still receive an alternative analysis in the grammar. We record such cases as BAD parses. The BAD parse rate is the proportion of BAD sentences that can be parsed by the grammar for a given phenomenon and grammar.

In this study, BAD parse rate is used as a diagnostic measure alongside minimal-pair accuracy. It identifies phenomena in which alternative grammatical analyses are available and provides additional context for interpreting the evaluation results. Section~\ref{app:per-phenomenon-evaluation} discusses BAD parse patterns for the relevant phenomena and reports BAD parse rate alongside the accuracy heatmaps.

In the heatmaps in this appendix, the numbers report three-seed mean accuracy, and darker green indicates higher accuracy on a 50--100\% scale. Accuracies below 50\% use the lightest shading. Orange bars indicate the grammar-based BAD parse rate on a 0--100\% scale, with the full width of the track corresponding to 100\%.

\subsection{Per-Phenomenon Evaluation}
\label{app:per-phenomenon-evaluation}

The following sections provide the construction details and results for each of the 34 phenomena, organized into the six evaluation groups.

\subsubsection{Basic Finite Clauses}
\label{app:basic-finite-clauses}

\ancevalphenomenon{1.1}{Intransitive Finite Verb Form}{app:e1-1-intran-v-form}

This phenomenon tests whether the models learned the finite form of intransitive verbs in finite intransitive clauses. The source templates include two types of subjects: simple NPs and NPs with a genitive possessor.

In all target grammars, the finite verb form is realized with the suffix \textit{-s}. The BAD sentence therefore replaces the target intransitive verb with its nonfinite form, changing \textit{V-s} to \textit{V-ing}. An example is shown below:

\begin{exe}
\ex\label{ex:e1-1-intran-v-form}
\begin{xlist}
\ex GOOD: \textit{Girlge teacher laughs}
\ex BAD:  \textit{Girlge teacher laughing}
\end{xlist}
\end{exe}

\begin{figure}[htbp]
\centering
\includegraphics[width=\columnwidth]{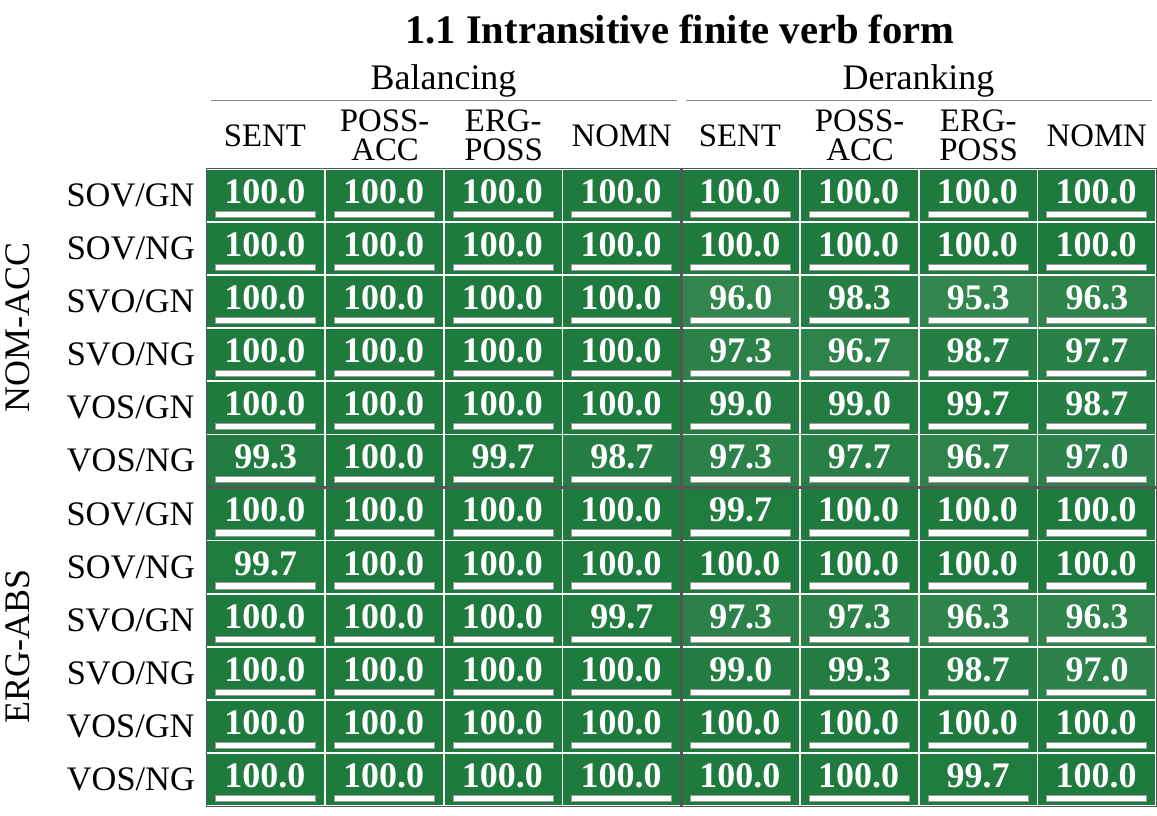}
\caption{Accuracy and BAD parse rate for 1.1 Intransitive finite verb form.}
\label{fig:e1-1-intran-v-form}
\end{figure}

Figure~\ref{fig:e1-1-intran-v-form} shows that the models perform near ceiling on this phenomenon. The mean accuracy across the 96 configurations is 99.4\%, with a minimum of 95.3\% and a maximum of 100\%. The BAD parse rate is 0 for all grammars. Overall, this result shows that the models robustly learned that finite intransitive clauses require the finite verb form.

\ancevalphenomenon{1.2}{Intransitive S Marker}{app:e1-2-intran-s-marker}

This phenomenon tests whether the models learned that the S argument in a finite intransitive clause should not receive an overt marker. The source templates include two types of subjects: simple NPs and NPs with a genitive possessor.

In all target grammars, the finite intransitive S marker is zero. The BAD sentence therefore adds an overt marker, either \textit{-ca} or \textit{-ge}, to the head of S. An example is shown below:

\begin{exe}
\ex\label{ex:e1-2-intran-s-marker}
\begin{xlist}
\ex GOOD: \textit{Doctorge nurse waits}
\ex BAD:  \textit{Doctorge nurseca waits}
\end{xlist}
\end{exe}

\begin{figure}[htbp]
\centering
\includegraphics[width=\columnwidth]{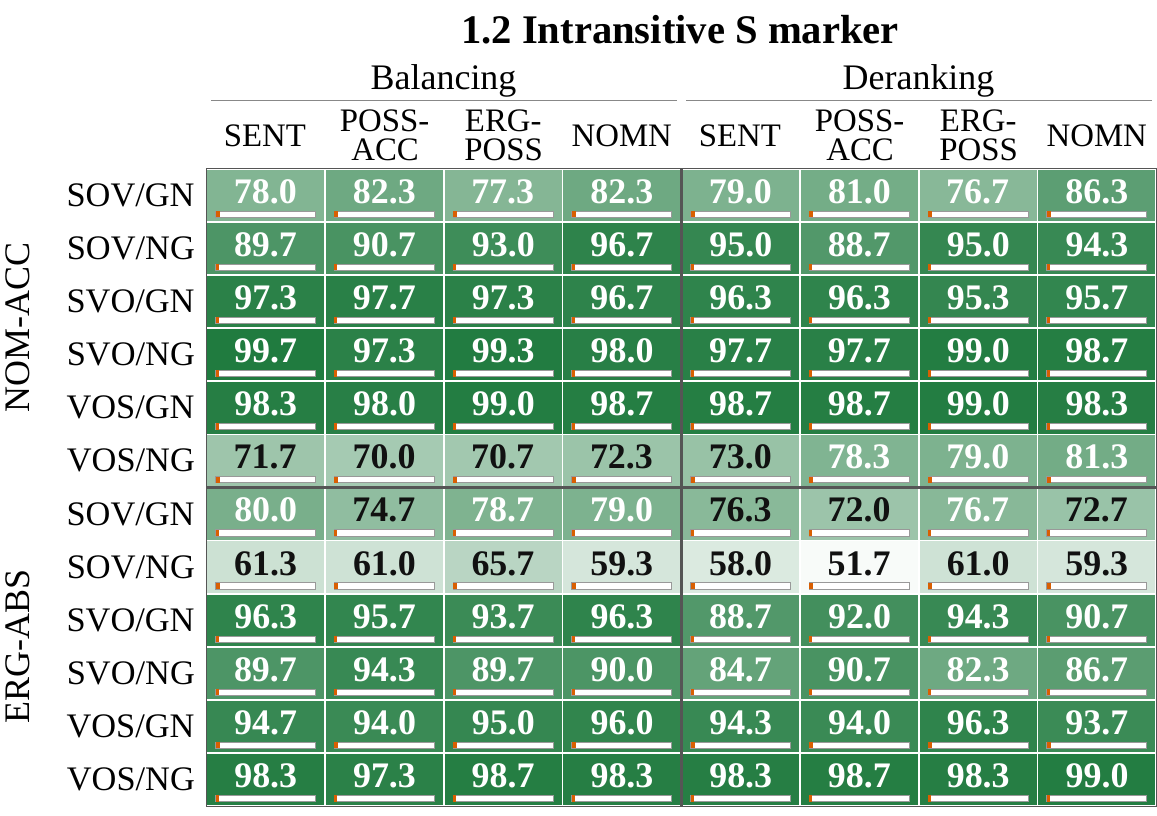}
\caption{Accuracy and BAD parse rate for 1.2 Intransitive S marker.}
\label{fig:e1-2-intran-s-marker}
\end{figure}

Figure~\ref{fig:e1-2-intran-s-marker} shows that models generally learned this phenomenon well. The mean accuracy across the 96 configurations is 87.7\%, with a minimum of 51.7\% and a maximum of 99.7\%. The BAD parse rate is low, ranging from 3.0\% to 4.0\% across grammars.

The weakest configuration is SOV/NG + ERG-ABS, with a mean accuracy of 59.7\%. Another weaker configuration is VOS/NG + NOM-ACC, with a mean accuracy of 74.5\%. These errors mainly reflect an interaction between alignment and word order. In these configurations, adding a marker to S can make the BAD sentence resemble a transitive argument-marking pattern. For example, under SOV/NG + ERG-ABS, the GOOD sentence \textit{Nurse doctorge waits} is perturbed into \textit{Nurseca doctorge waits}. Since \textit{-ca} is the legal A marker in this configuration, the model may treat \textit{Nurseca} as a transitive A and \textit{doctorge} as a transitive P. This suggests that the model does not robustly distinguish genitive possessor marking from transitive argument marking in this configuration.

\ancevalphenomenon{1.3}{Intransitive S--V Order}{app:e1-3-intran-sv-order}

This phenomenon tests whether the models learned the linear order between S and the finite verb in finite intransitive clauses. Intransitive clauses distinguish two surface orders: SOV and SVO yield S--V order, while VOS yields V--S order.

The BAD sentence is constructed by swapping the entire subject span and the finite verb. An example is shown below:

\begin{exe}
\ex\label{ex:e1-3-intran-sv-order}
\begin{xlist}
\ex GOOD: \textit{Bakerge bread rises}
\ex BAD:  \textit{Rises bakerge bread}
\end{xlist}
\end{exe}

\begin{figure}[htbp]
\centering
\includegraphics[width=\columnwidth]{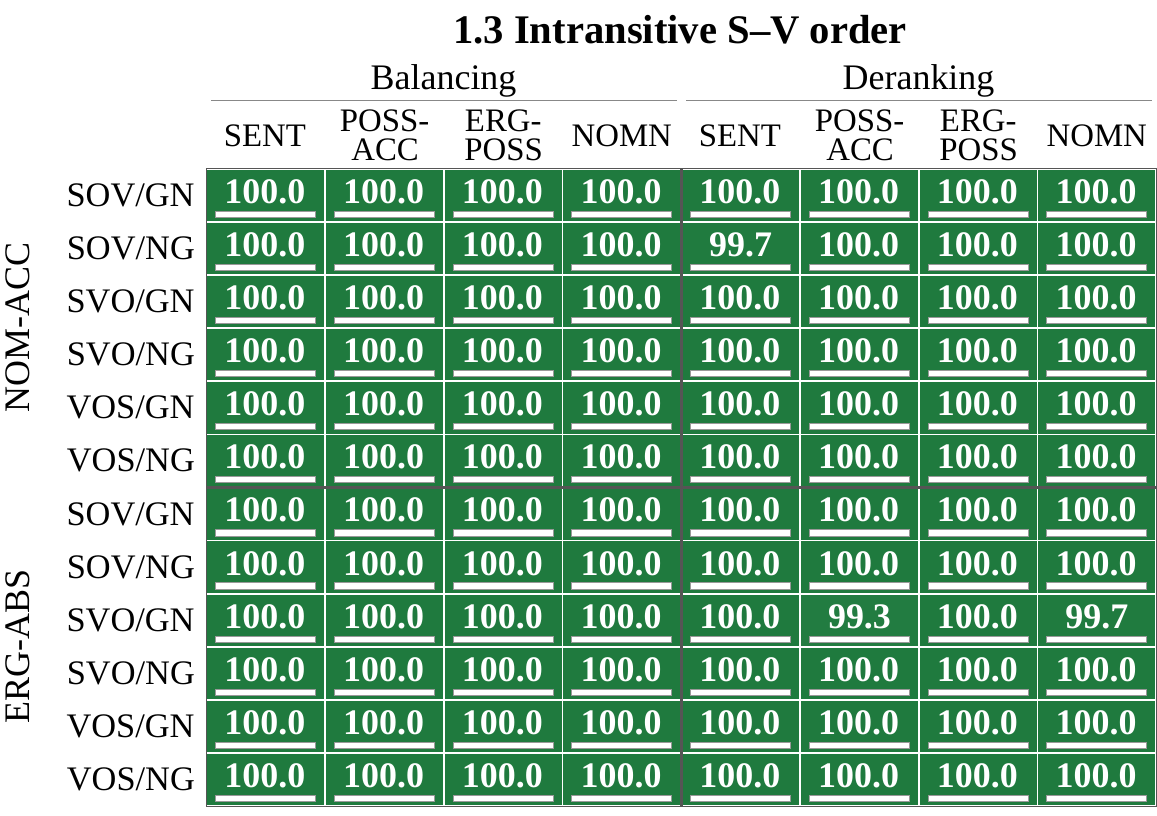}
\caption{Accuracy and BAD parse rate for 1.3 Intransitive S--V order.}
\label{fig:e1-3-intran-sv-order}
\end{figure}

Figure~\ref{fig:e1-3-intran-sv-order} shows that the models perform at ceiling on this phenomenon. The mean accuracy across the 96 configurations is 99.99\%, with a minimum of 99.3\% and a maximum of 100\%. The BAD parse rate is 0 for all grammars. Overall, this result shows that the models robustly learned the basic S--V order of finite intransitive clauses.

\ancevalphenomenon{1.4}{Transitive Finite Verb Form}{app:e1-4-tran-v-form}

This phenomenon tests whether the models learned the finite form of transitive verbs in finite transitive clauses. The source templates include transitive clauses in which A and P may be simple NPs or NPs with a genitive possessor.

In all target grammars, the finite verb form is realized with the suffix \textit{-s}. The BAD sentence therefore replaces the target transitive verb with its nonfinite form, changing \textit{V-s} to \textit{V-ing}. An example is shown below:

\begin{exe}
\ex\label{ex:e1-4-tran-v-form}
\begin{xlist}
\ex GOOD: \textit{Nurse holds cupca}
\ex BAD:  \textit{Nurse holding cupca}
\end{xlist}
\end{exe}

\begin{figure}[htbp]
\centering
\includegraphics[width=\columnwidth]{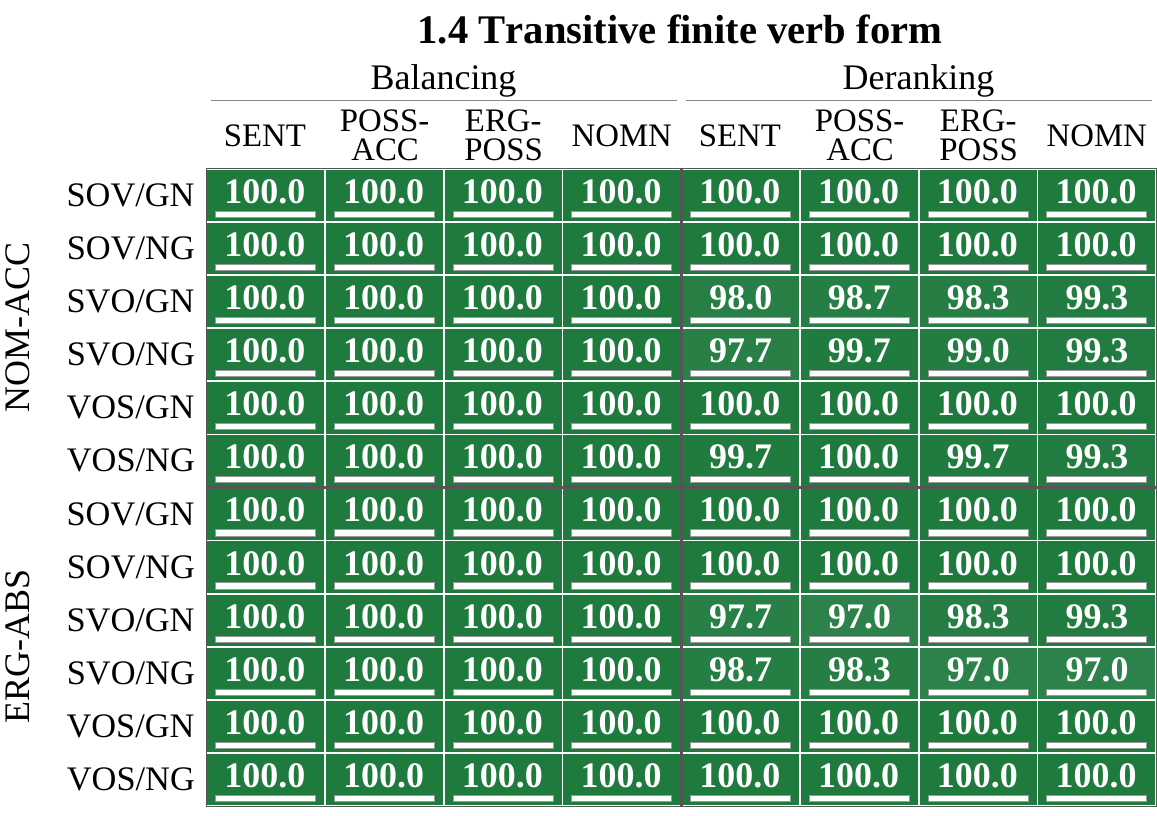}
\caption{Accuracy and BAD parse rate for 1.4 Transitive finite verb form.}
\label{fig:e1-4-tran-v-form}
\end{figure}

Figure~\ref{fig:e1-4-tran-v-form} shows that the models perform near ceiling on this phenomenon. The mean accuracy across the 96 configurations is 99.7\%, with a minimum of 97.0\% and a maximum of 100\%. The BAD parse rate is 0 for all grammars. The few remaining errors are mostly in SVO grammars with Deranking, especially SVO/NG + ERG-ABS + Deranking. This reflects the fact that \textit{V-ing} is more common in deranking systems than in balancing systems.

\ancevalphenomenon{1.5}{Transitive A Marker}{app:e1-5-tran-a-marker}

This phenomenon tests whether the models learned the marking of the A argument in basic finite transitive clauses. The source templates include four types of transitive clauses, where A and P can each be either a simple NP or an NP with a genitive possessor.

In the target grammars, A marking is determined by alignment. In NOM-ACC grammars, A has zero marking, while in ERG-ABS grammars, A has \textit{-ca}. So in NOM-ACC grammars, the BAD sentence adds \textit{-ca} or \textit{-ge} to A. In ERG-ABS grammars, the BAD sentence either removes \textit{-ca} from A or replaces it with \textit{-ge}. An example is shown below:

\begin{exe}
\ex\label{ex:e1-5-tran-a-marker}
\begin{xlist}
\ex GOOD: \textit{Doctorge assistant finds fileca}
\ex BAD:  \textit{Doctorge assistantca finds fileca}
\end{xlist}
\end{exe}

\begin{figure}[htbp]
\centering
\includegraphics[width=\columnwidth]{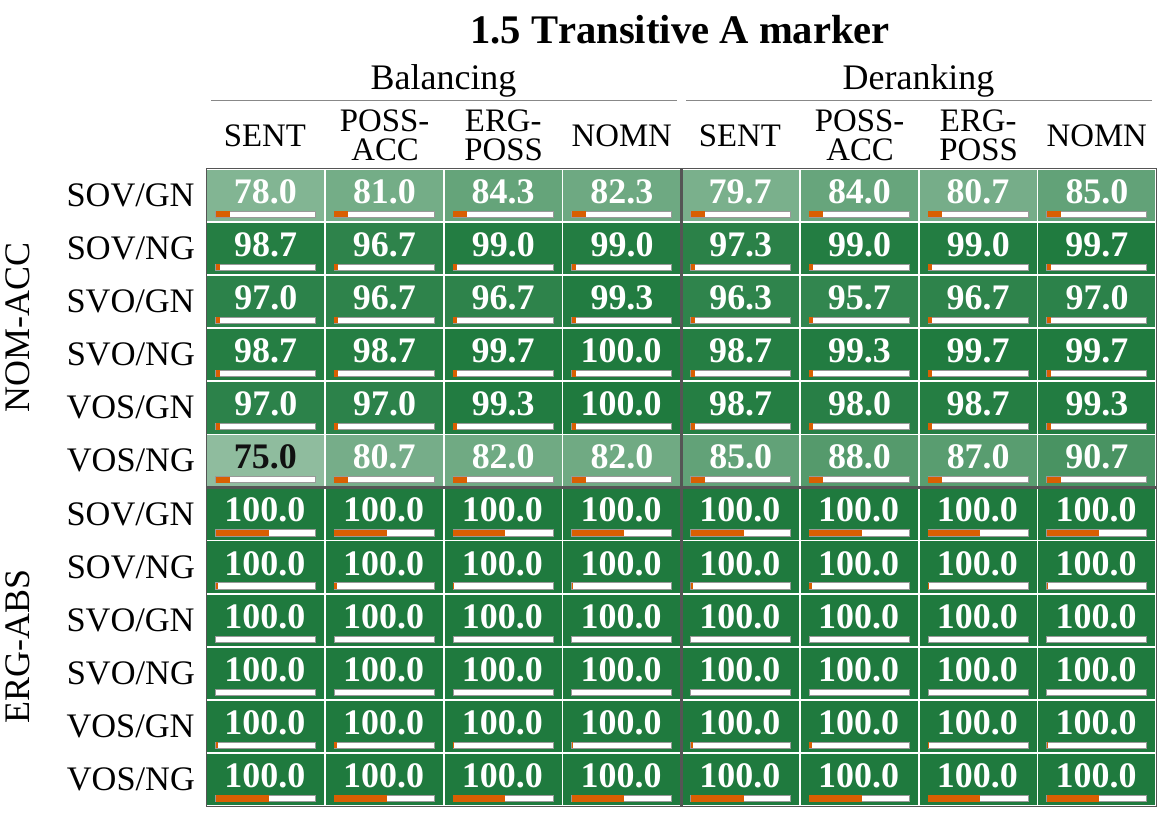}
\caption{Accuracy and BAD parse rate for 1.5 Transitive A marker.}
\label{fig:e1-5-tran-a-marker}
\end{figure}

Figure~\ref{fig:e1-5-tran-a-marker} shows that the models generally learned transitive A marking. The mean accuracy across the 96 configurations is 96.6\%, with a minimum of 75.0\% and a maximum of 100\%. The BAD parse rate ranges from 0.0\% to 53.0\%. However, BAD parse rate and model accuracy are asymmetric with respect to alignment: NOM-ACC grammars have a lower mean BAD parse rate of 7.3\% but a lower mean accuracy of 93.2\%; ERG-ABS grammars have a higher mean BAD parse rate of 18.1\% but reach 100\% accuracy.

Within NOM-ACC, the weakest configuration is SOV/GN + NOM-ACC, with a mean accuracy of 81.9\% and a BAD parse rate of 14.0\%. For example, \textit{Nurse childge kneeca cleans} is perturbed into \textit{Nursege childge kneeca cleans}. This BAD sentence cannot be parsed by the grammar, but the model still assigns it a higher score. A likely reason is that the model does not robustly identify \textit{-ge} as an illegal marker in the A position.

In contrast, although ERG-ABS grammars have a higher BAD parse rate, these parses often require a larger semantic reanalysis. Under SOV/GN + ERG-ABS, the mean accuracy is 100\% even though the BAD parse rate is 52.5\%. For example, \textit{Nurseca bandage touchs} is perturbed into \textit{Nursege bandage touchs}. This BAD sentence can be parsed as ``the nurse's bandage touches'', where the original A is reanalyzed as a genitive possessor, while the original P becomes the intransitive S. This analysis is syntactically available, but semantically unnatural, since many inanimate NPs that occur as P are not suitable as intransitive S.

\ancevalphenomenon{1.6}{Transitive P Marker}{app:e1-6-tran-p-marker}

This phenomenon tests whether the models learned the marking of the P argument in basic finite transitive clauses. The source templates include four types of transitive clauses, where A and P can each be either a simple NP or an NP with a genitive possessor.

In the target grammars, P marking is determined by alignment. In NOM-ACC grammars, P bears \textit{-ca}, while in ERG-ABS grammars, P has zero marking. In NOM-ACC grammars, the BAD sentence either removes \textit{-ca} from P or replaces it with \textit{-ge}. In ERG-ABS grammars, the BAD sentence incorrectly adds \textit{-ca} or \textit{-ge} to P. The perturbation applies only to the P head. An example is shown below:

\begin{exe}
\ex\label{ex:e1-6-tran-p-marker}
\begin{xlist}
\ex GOOD: \textit{Shop sells breadca}
\ex BAD:  \textit{Shop sells bread}
\end{xlist}
\end{exe}

\begin{figure}[htbp]
\centering
\includegraphics[width=\columnwidth]{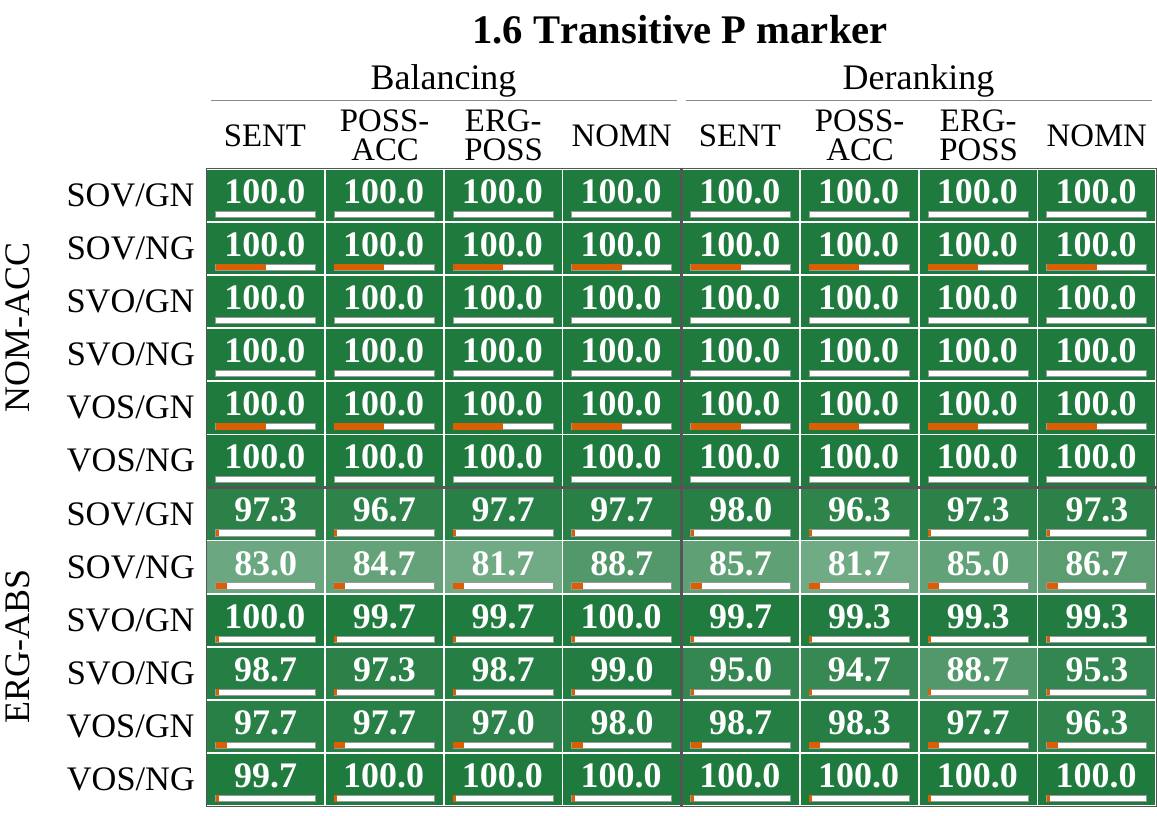}
\caption{Accuracy and BAD parse rate for 1.6 Transitive P marker.}
\label{fig:e1-6-tran-p-marker}
\end{figure}

Figure~\ref{fig:e1-6-tran-p-marker} shows that the models generally learned transitive P marking. The mean accuracy across the 96 configurations is 97.9\%, with a minimum of 81.7\% and a maximum of 100\%. The BAD parse rate ranges from 0.0\% to 50.0\%. Specifically, NOM-ACC grammars all reach 100\% accuracy, but have a higher mean BAD parse rate of 16.7\%; ERG-ABS grammars have a lower mean accuracy of 95.8\%, but a lower mean BAD parse rate of 5.7\%.

This phenomenon mirrors the pattern in Section~\ref{app:e1-5-tran-a-marker}. For example, SOV/NG + NOM-ACC reaches 100\% accuracy, but has a BAD parse rate of 50.0\%. The GOOD sentence \textit{Student answerca copys} is perturbed into \textit{Student answerge copys}. This BAD sentence can be parsed as roughly ``the answer's student copies'', where the original P is reanalyzed as a genitive possessor. Since this parse also involves semantic reanalysis, the model still consistently prefers the GOOD sentence. In contrast, in ERG-ABS grammars, P has zero marking and therefore lacks an overt P cue, making it more vulnerable to interference from incorrect markers. The weakest configuration is SOV/NG + ERG-ABS, with a mean accuracy of 84.6\% and a BAD parse rate of 11.0\%.

\ancevalphenomenon{1.7}{Transitive A--V Order}{app:e1-7-tran-av-order}

This phenomenon tests whether the models learned the relative order between the A argument and the finite transitive verb in basic finite transitive clauses. The source templates include four types of transitive clauses, where A and P can each be either a simple NP or an NP with a genitive possessor.

The BAD sentence is constructed by swapping the A span and the V+P chunk. An example is shown below:

\begin{exe}
\ex\label{ex:e1-7-tran-av-order}
\begin{xlist}
\ex GOOD: \textit{Bakerge daughter slices breadca}
\ex BAD:  \textit{Slices breadca bakerge daughter}
\end{xlist}
\end{exe}

\begin{figure}[htbp]
\centering
\includegraphics[width=\columnwidth]{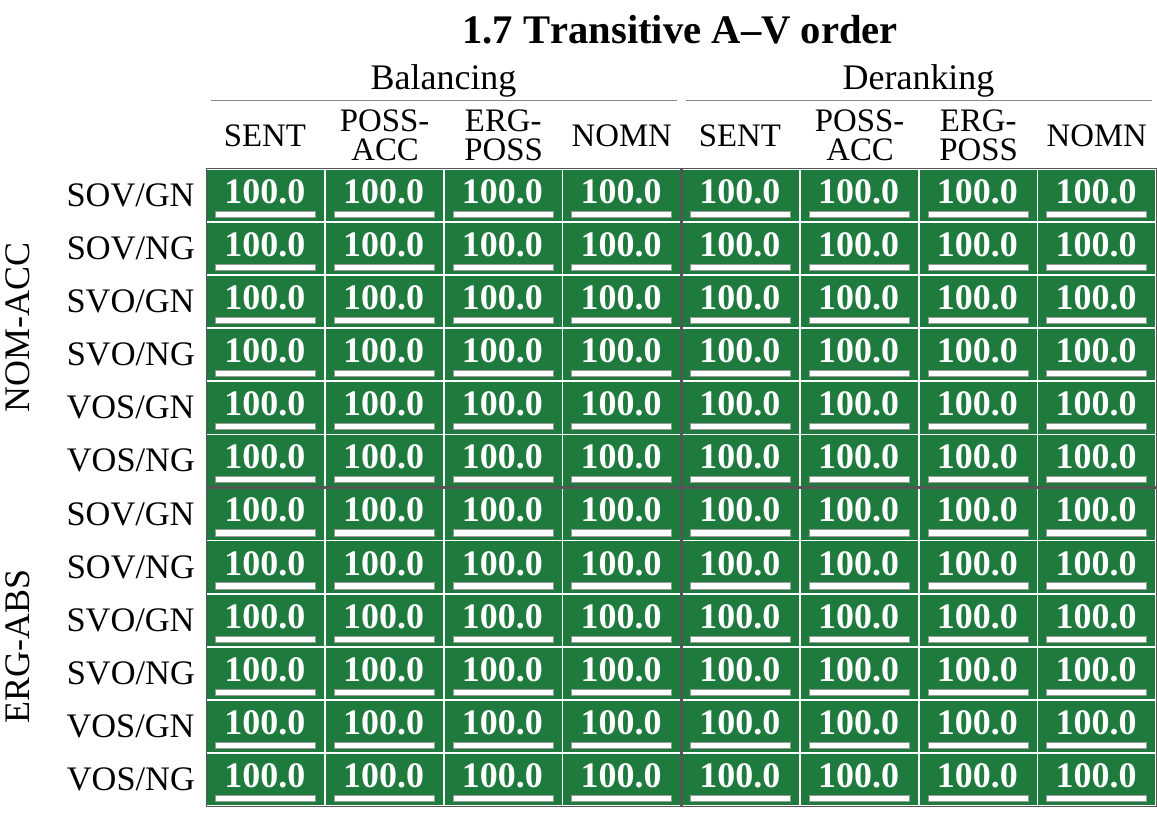}
\caption{Accuracy and BAD parse rate for 1.7 Transitive A--V order.}
\label{fig:e1-7-tran-av-order}
\end{figure}

Figure~\ref{fig:e1-7-tran-av-order} shows that the models reach ceiling on this phenomenon. All 96 grammars have 100\% accuracy, and the BAD parse rate is 0 for all grammars. This shows that the models learned the order relation between A and V very robustly.

\ancevalphenomenon{1.8}{Transitive V--P Order}{app:e1-8-tran-vp-order}

This phenomenon tests whether the models learned the relative order between the finite transitive verb and the P argument in basic finite transitive clauses. The source templates include four types of transitive clauses, where A and P can each be either a simple NP or an NP with a genitive possessor.

The BAD sentence is constructed by swapping V and the P span, while the A span remains in place. An example is shown below:

\begin{exe}
\ex\label{ex:e1-8-tran-vp-order}
\begin{xlist}
\ex GOOD: \textit{Shopkeeper closes windowca}
\ex BAD:  \textit{Shopkeeper windowca closes}
\end{xlist}
\end{exe}

\begin{figure}[htbp]
\centering
\includegraphics[width=\columnwidth]{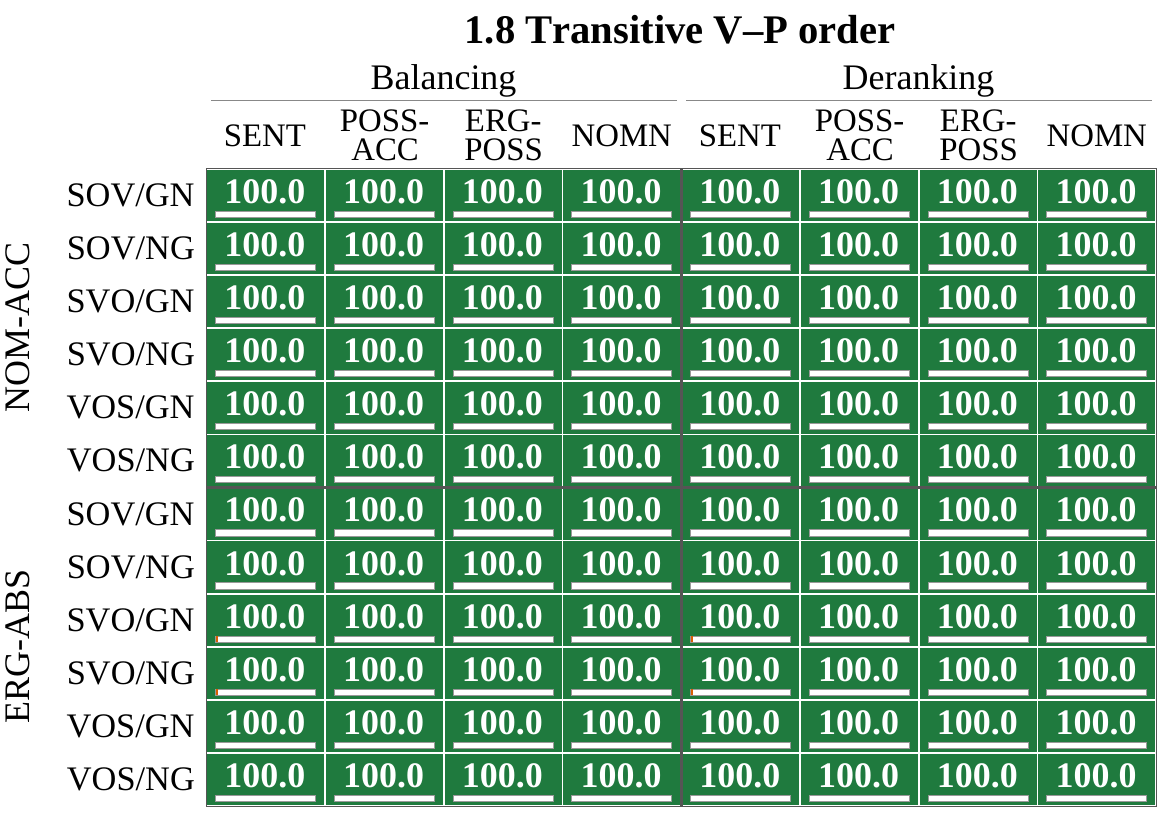}
\caption{Accuracy and BAD parse rate for 1.8 Transitive V--P order.}
\label{fig:e1-8-tran-vp-order}
\end{figure}

Figure~\ref{fig:e1-8-tran-vp-order} shows that the models reach ceiling on this phenomenon. All 96 grammars have 100\% accuracy. The BAD parse rate is also nearly zero, with a mean of 0.08\%. This shows that the models learned the order relation between V and P very robustly.

\ancevalphenomenon{1.9}{Intransitive V Valency}{app:e1-9-intran-v-valency}

This phenomenon tests whether the models learned verb valency in basic finite intransitive clauses. The GOOD sentence is a finite clause containing S and an intransitive verb. The source templates include simple-NP subjects and subjects with a genitive possessor.

This phenomenon uses a valency pair table for perturbation. The verb pairs come from the BLiMP intransitive paradigm \citep{warstadt-etal-2020-blimp-benchmark}. The GOOD verb is an intransitive verb, while the BAD verb is a transitive verb that does not match the environment. The source sentences are regenerated with prompts around the GOOD verbs. The BAD sentence is constructed by replacing only the target verb stem while keeping the finite form unchanged. An example is shown below:

\begin{exe}
\ex\label{ex:e1-9-intran-v-valency}
\begin{xlist}
\ex GOOD: \textit{Clerk smiles}
\ex BAD:  \textit{Clerk admires}
\end{xlist}
\end{exe}

\begin{figure}[htbp]
\centering
\includegraphics[width=\columnwidth]{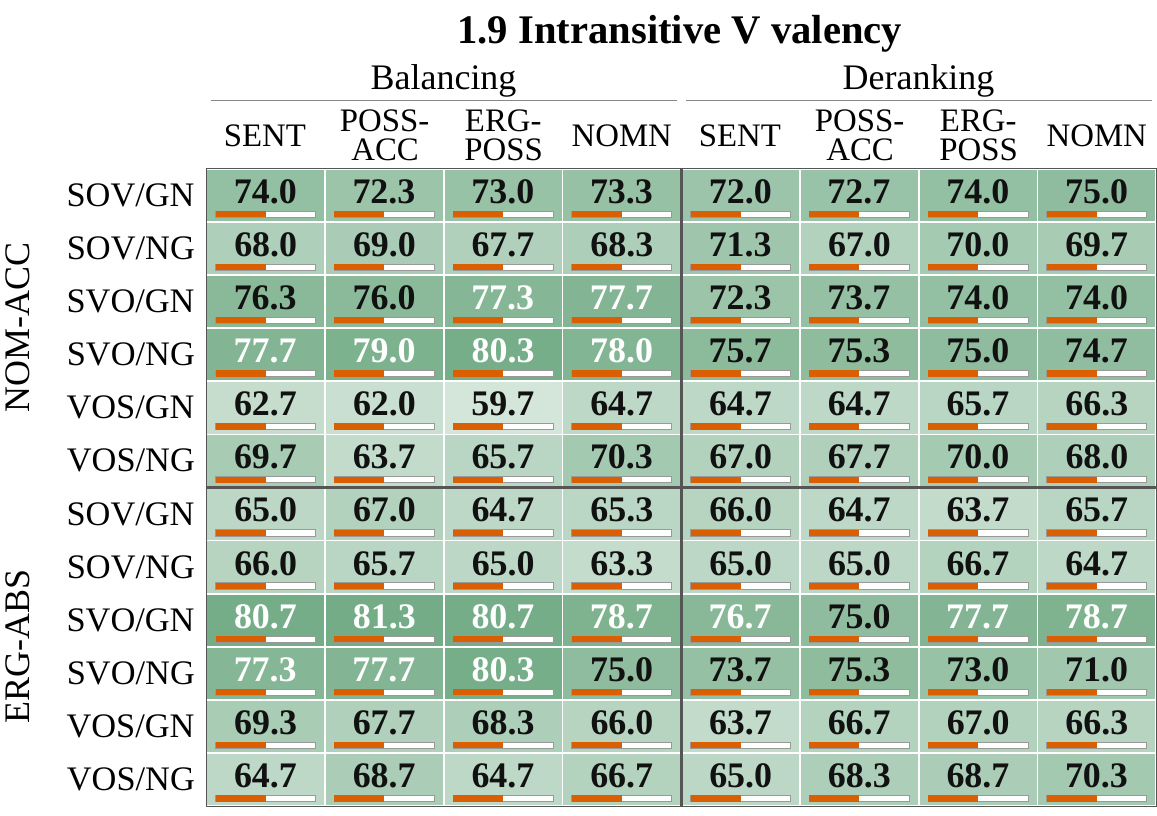}
\caption{Accuracy and BAD parse rate for 1.9 Intransitive V valency.}
\label{fig:e1-9-intran-v-valency}
\end{figure}

Figure~\ref{fig:e1-9-intran-v-valency} shows that this phenomenon is much harder than the preceding form, marker, and word-order phenomena. The mean accuracy across the 96 configurations is 70.4\%, with a minimum of 59.7\% and a maximum of 81.3\%. The BAD parse rate is 50.0\% for all grammars, showing that about half of the BAD verbs occurred without a complement in the original training corpus.

Across clause word orders, SVO grammars have the highest mean accuracy at 76.6\%. SOV and VOS grammars are lower, at 68.1\% and 66.4\%, respectively. The other parameters have only small effects. This suggests that SVO word order helps the models distinguish the environments of intransitive and transitive verbs. The overall accuracy is still not high, showing that lexical argument structure remains difficult for small models trained on small corpora.

\ancevalphenomenon{1.10}{Transitive V Valency}{app:e1-10-tran-v-valency}

This phenomenon tests whether the models learned verb valency in basic finite transitive clauses. The GOOD sentence is a finite clause containing A, a transitive verb, and P. The source templates include four types of transitive clauses, where A and P can each be either a simple NP or an NP with a genitive possessor.

This phenomenon also uses a valency pair table for perturbation. The verb pairs come from the BLiMP transitive paradigm \citep{warstadt-etal-2020-blimp-benchmark}. The GOOD verb is a transitive verb, while the BAD verb is an intransitive verb that does not match the environment. The source sentences are regenerated with prompts around the GOOD verbs. The BAD sentence is constructed by replacing only the target verb stem while keeping the finite form unchanged. An example is shown below:

\begin{exe}
\ex\label{ex:e1-10-tran-v-valency}
\begin{xlist}
\ex GOOD: \textit{Guest insults hostca}
\ex BAD:  \textit{Guest sighs hostca}
\end{xlist}
\end{exe}

\begin{figure}[htbp]
\centering
\includegraphics[width=\columnwidth]{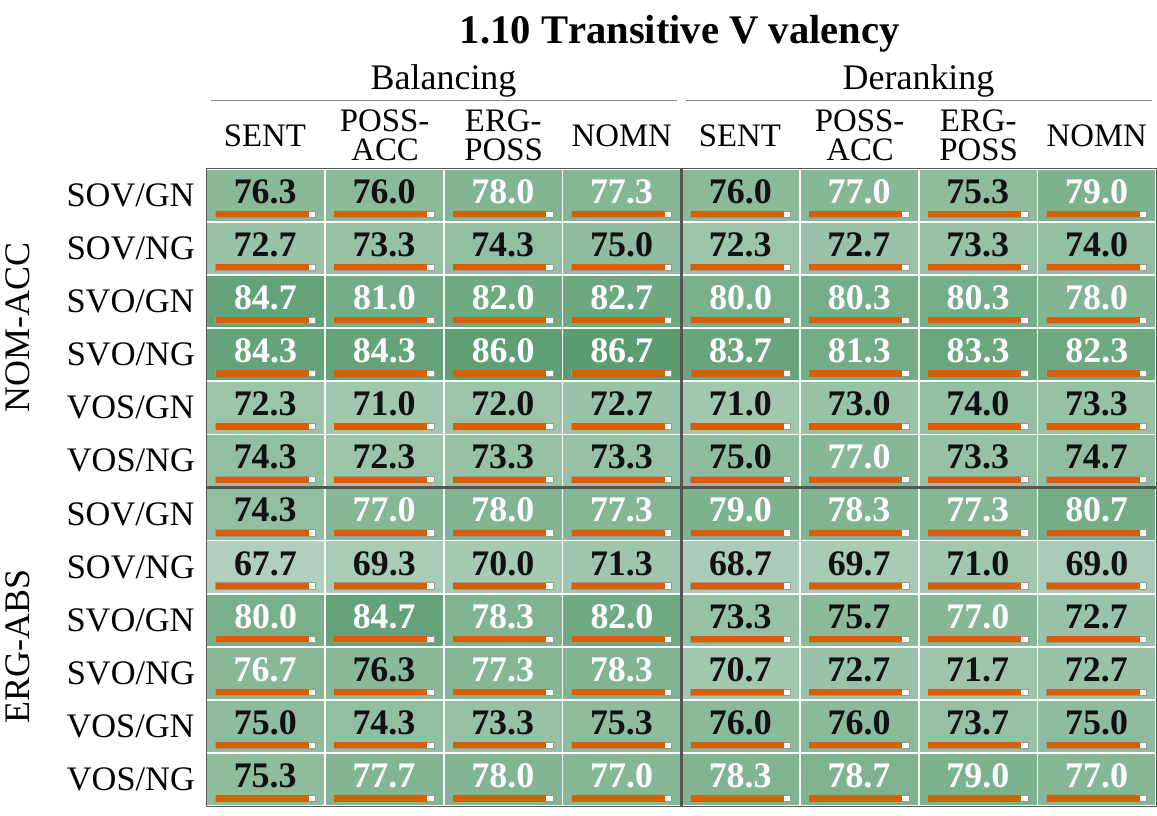}
\caption{Accuracy and BAD parse rate for 1.10 Transitive V valency.}
\label{fig:e1-10-tran-v-valency}
\end{figure}

Figure~\ref{fig:e1-10-tran-v-valency} shows that this phenomenon is also difficult, but the models perform slightly better here than on Intransitive V Valency. The mean accuracy across the 96 configurations is 76.2\%, with a minimum of 67.7\% and a maximum of 86.7\%. The BAD parse rate is 93.0\% for all grammars, indicating that most BAD verbs also occurred in transitive frames in the training corpus, so accuracy reflects both valency sensitivity and preferences between grammatical alternatives.

Across clause word orders, SVO grammars have the highest mean accuracy at 79.4\%. SOV and VOS grammars are lower, at 74.4\% and 74.8\%, respectively. The other parameters have only small effects. This pattern is consistent with the intransitive valency result.

\subsubsection{Noun Phrases}
\label{app:noun-phrases}

\ancevalphenomenon{2.1}{Intransitive Genitive Marker}{app:e1-11-intran-gen-marker}

This phenomenon tests whether the models learned the genitive marker on the possessor inside the subject NP of an intransitive clause. The GOOD sentence is a finite intransitive clause in which S is an NP with one genitive possessor.

In all target grammars, the genitive marker is \textit{-ge}. The BAD sentence is constructed by modifying only the \textit{-ge} on the possessor, either deleting it or replacing it with \textit{-ca}. An example is shown below:

\begin{exe}
\ex\label{ex:e1-11-intran-gen-marker}
\begin{xlist}
\ex GOOD: \textit{Studentge father laughs}
\ex BAD:  \textit{Student father laughs}
\end{xlist}
\end{exe}

\begin{figure}[htbp]
\centering
\includegraphics[width=\columnwidth]{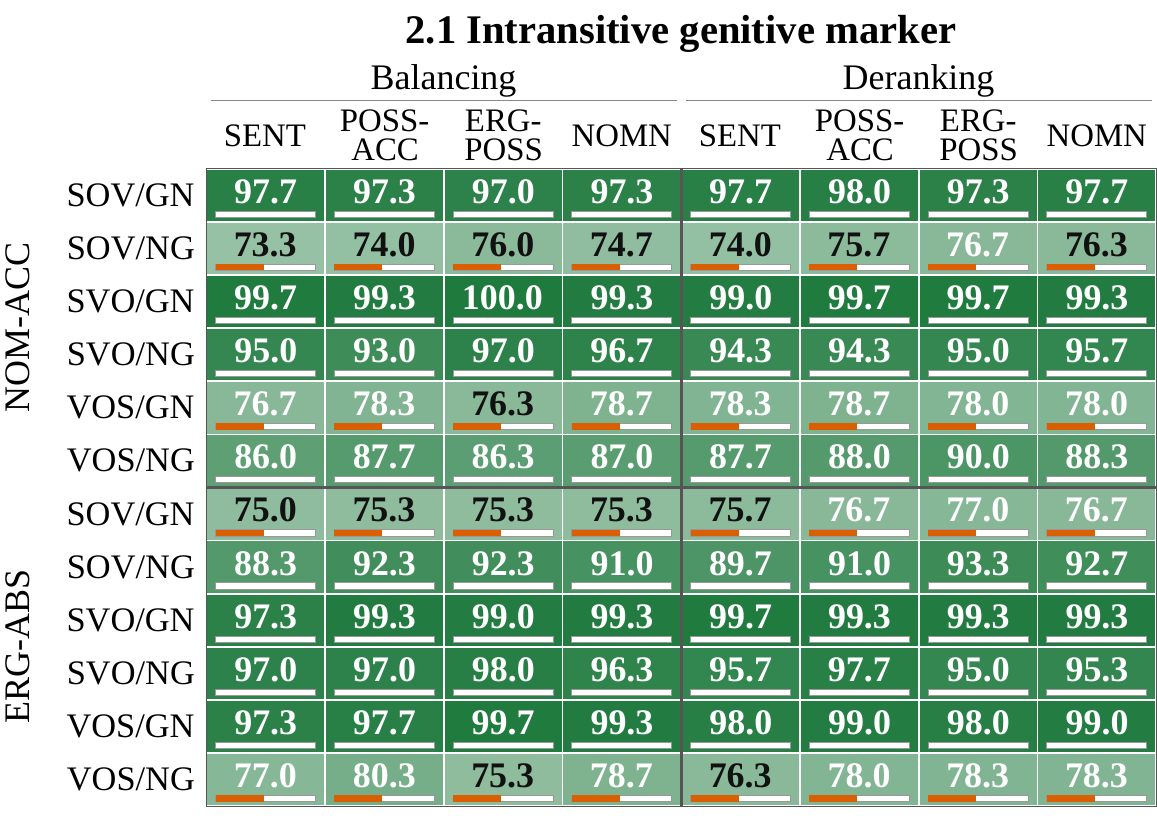}
\caption{Accuracy and BAD parse rate for 2.1 Intransitive genitive marker.}
\label{fig:e1-11-intran-gen-marker}
\end{figure}

Figure~\ref{fig:e1-11-intran-gen-marker} shows that the models generally learned genitive marking inside intransitive subjects. The mean accuracy across the 96 configurations is 89.3\%, with a minimum of 73.3\% and a maximum of 100\%. The BAD parse rate ranges from 0\% to 48\%, with a mean of 16.0\%.

SVO grammars perform best, with a mean accuracy of 97.6\%. The weaker configurations are mainly found under SOV and VOS grammars. The four weakest groups are SOV/NG + NOM-ACC, VOS/GN + NOM-ACC, SOV/GN + ERG-ABS, and VOS/NG + ERG-ABS, with mean accuracies of 75.1\%, 77.9\%, 75.9\%, and 77.8\%, respectively.

The main errors occur when the model treats BAD sentences with \textit{-ca} as legal transitive patterns. For example, under SOV/GN + ERG-ABS, the GOOD sentence \textit{Studentge father laughs} can be perturbed into \textit{Studentca father laughs}. Since \textit{-ca} is the A marker in this configuration, the BAD sentence resembles an SOV transitive clause, roughly interpretable as ``the student laughs the father''.

\ancevalphenomenon{2.2}{Transitive Genitive Marker}{app:e1-12-tran-gen-marker}

This phenomenon tests whether the models learned the genitive marker on possessors inside A/P argument NPs in transitive clauses. The GOOD sentence is a finite transitive clause in which at least one argument NP has a genitive possessor.

The source templates and perturbation rules cover four cases: targeting the possessor inside A when A has a possessor, targeting the possessor inside P when P has a possessor, and targeting the possessor inside A or P separately when both A and P have possessors.

In all target grammars, the genitive marker is \textit{-ge}. The BAD sentence is constructed by modifying only the \textit{-ge} on the target possessor, either deleting it or replacing it with \textit{-ca}. An example is shown below:

\begin{exe}
\ex\label{ex:e1-12-tran-gen-marker}
\begin{xlist}
\ex GOOD: \textit{Olivia calls studentge roommateca}
\ex BAD:  \textit{Olivia calls studentca roommateca}
\end{xlist}
\end{exe}

\begin{figure}[htbp]
\centering
\includegraphics[width=\columnwidth]{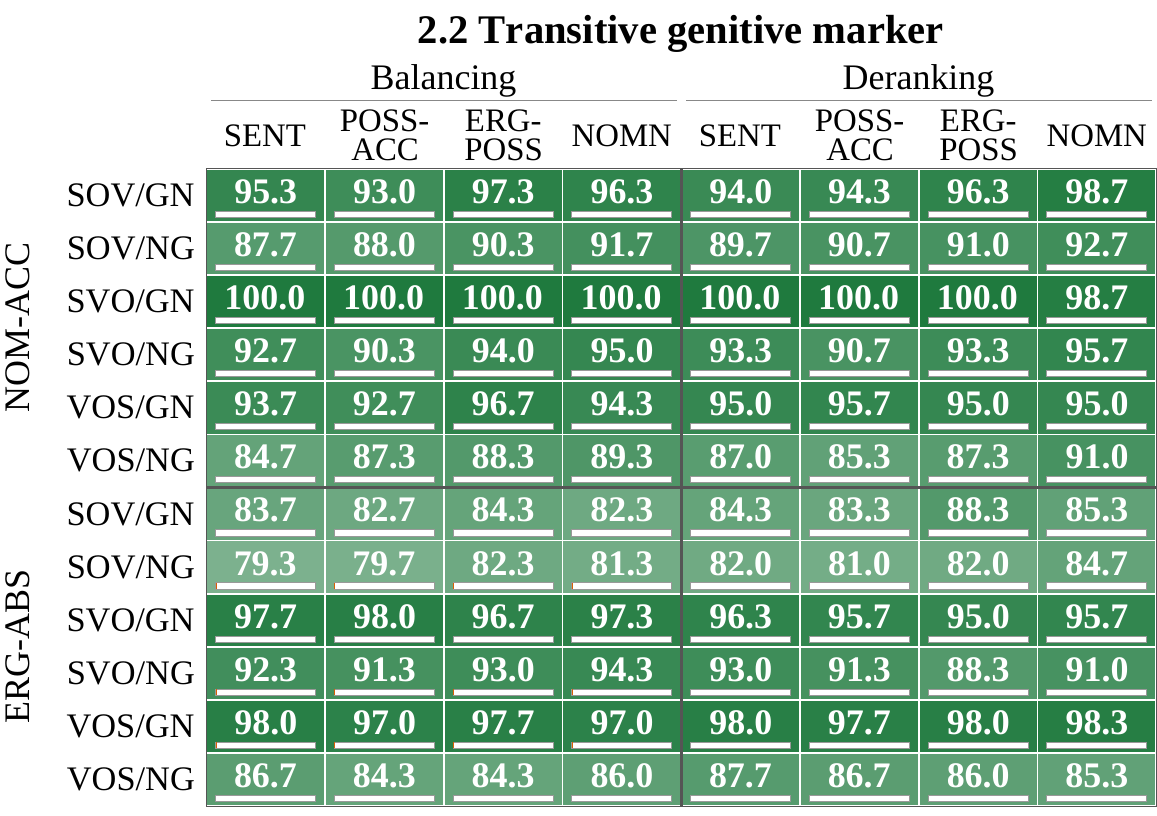}
\caption{Accuracy and BAD parse rate for 2.2 Transitive genitive marker.}
\label{fig:e1-12-tran-gen-marker}
\end{figure}

Figure~\ref{fig:e1-12-tran-gen-marker} shows that the models generally learned genitive marking inside transitive argument NPs. The mean accuracy across the 96 configurations is 91.6\%, with a minimum of 79.3\% and a maximum of 100\%. The BAD parse rate is very low, with a mean of only 0.1\%.

The result is broadly consistent with Section~\ref{app:e1-11-intran-gen-marker}. SVO grammars perform best, with a mean accuracy of 95.3\%; SOV and VOS grammars are lower, at 87.9\% and 91.5\%, respectively. The weakest group is SOV/NG + ERG-ABS, with a mean accuracy of 81.5\%. Overall accuracy is slightly higher than in Section~\ref{app:e1-11-intran-gen-marker}, possibly because after the marker perturbation, the model can more easily identify that the verb cannot accommodate three arguments.

\ancevalphenomenon{2.3}{Genitive Possessor--Head Order}{app:e1-13-gen-order}

This phenomenon tests whether the models learned the relative order of the possessor and the head noun inside genitive NPs. The source templates place the target genitive NP in one of three positions: intransitive S, transitive A, or transitive P.

In the target grammars, NP word order is either GN or NG. In GN grammars, the possessor precedes the head noun; in NG grammars, the head noun precedes the possessor. The BAD sentence is constructed by swapping only the possessor and the head noun inside the target genitive NP. An example is shown below:

\begin{exe}
\ex\label{ex:e1-13-gen-order}
\begin{xlist}
\ex GOOD: \textit{Reporter interviews clerkge bossca}
\ex BAD:  \textit{Reporter interviews bossca clerkge}
\end{xlist}
\end{exe}

\begin{figure}[htbp]
\centering
\includegraphics[width=\columnwidth]{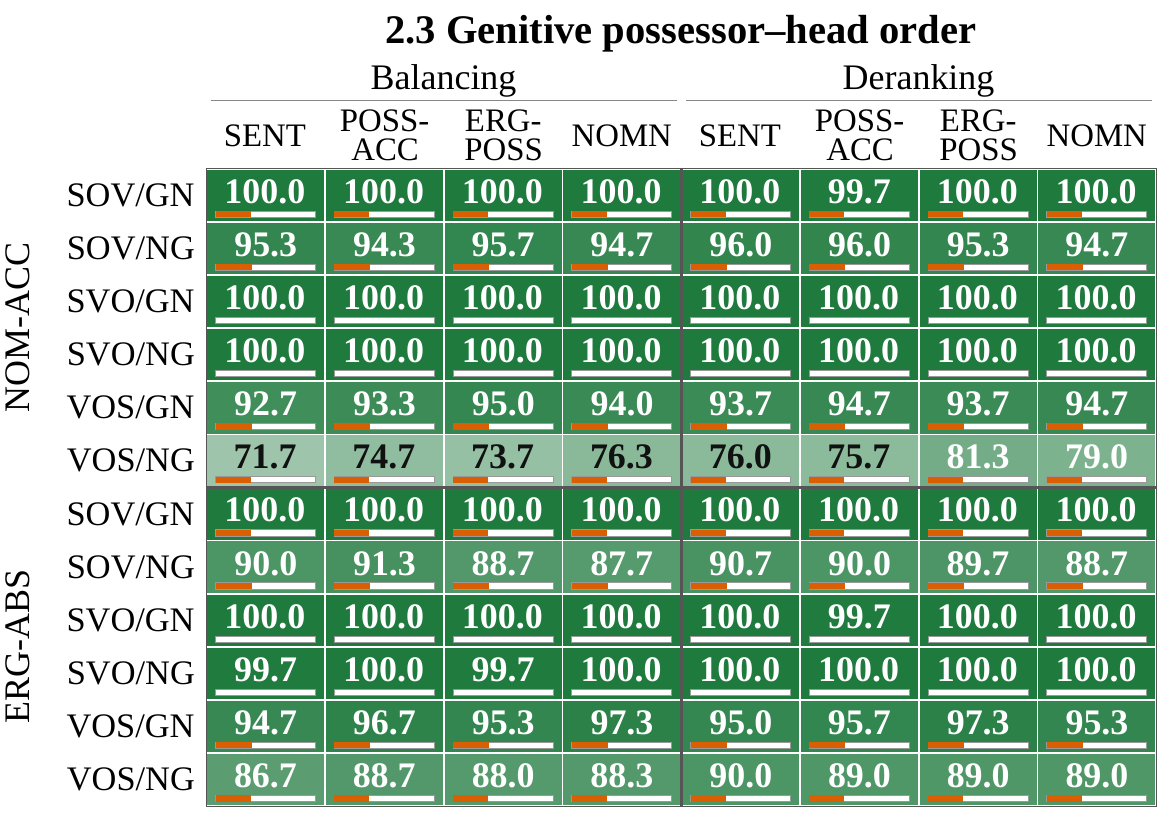}
\caption{Accuracy and BAD parse rate for 2.3 Genitive possessor--head order.}
\label{fig:e1-13-gen-order}
\end{figure}

Figure~\ref{fig:e1-13-gen-order} shows that the models generally learned genitive possessor--head order. The mean accuracy across the 96 configurations is 94.9\%, with a minimum of 71.7\% and a maximum of 100\%. The BAD parse rate ranges from 0\% to 36\%, with a mean of 23.7\%.

Across typological parameters, SVO grammars perform best, with a mean accuracy of 100\%. SOV grammars are also high, with a mean accuracy of 96.2\%, while VOS grammars are weaker, at 88.6\%. NP order also has an effect. GN grammars have a mean accuracy of 98.3\%, compared with 91.6\% for NG grammars.

The weaker performance mainly comes from cases where the perturbed genitive sequence can be reanalyzed as part of another argument. For example, the VOS/NG + NOM-ACC configuration has a mean accuracy of 76.0\%. Under this configuration, the GOOD sentence \textit{Chases rabbitca dog noahge} is perturbed into \textit{Chases rabbitca noahge dog}. The GOOD sentence has the structure \textit{Chases [rabbitca] [dog noahge]}, roughly corresponding to ``Noah's dog chases the rabbit.'' However, the BAD sentence can be reanalyzed as \textit{Chases [rabbitca noahge] [dog]}, roughly corresponding to ``the dog chases Noah's rabbit.'' Both analyses are semantically reasonable, so this case should not be treated as a simple model mistake. Nevertheless, it suggests that genitive order is more prone to confusion in this configuration.

\subsubsection{Clausal Complements}
\label{app:clausal-complements}

\ancevalphenomenon{3.1}{Clausal Complement Verb Form}{app:e1-14-clausal-v-form}

This phenomenon tests whether the models learned the form of the embedded verb in clausal complements. The GOOD sentence is a finite matrix clause whose matrix verb is a clausal-complement verb, and whose complement is an embedded intransitive or transitive clause. The source templates cover both embedded IV and embedded TV clauses.

In the target grammars, embedded verb form is determined by the complement system. In Balancing grammars, embedded V uses the finite form \textit{V-s}; in Deranking grammars, embedded V uses the nonfinite form \textit{V-ing}. The BAD sentence changes the embedded verb form to the other form. An example is shown below:

\begin{exe}
\ex\label{ex:e1-14-clausal-v-form}
\begin{xlist}
\ex GOOD: \textit{Captain notices boat drifts}
\ex BAD:  \textit{Captain notices boat drifting}
\end{xlist}
\end{exe}

\begin{figure}[htbp]
\centering
\includegraphics[width=\columnwidth]{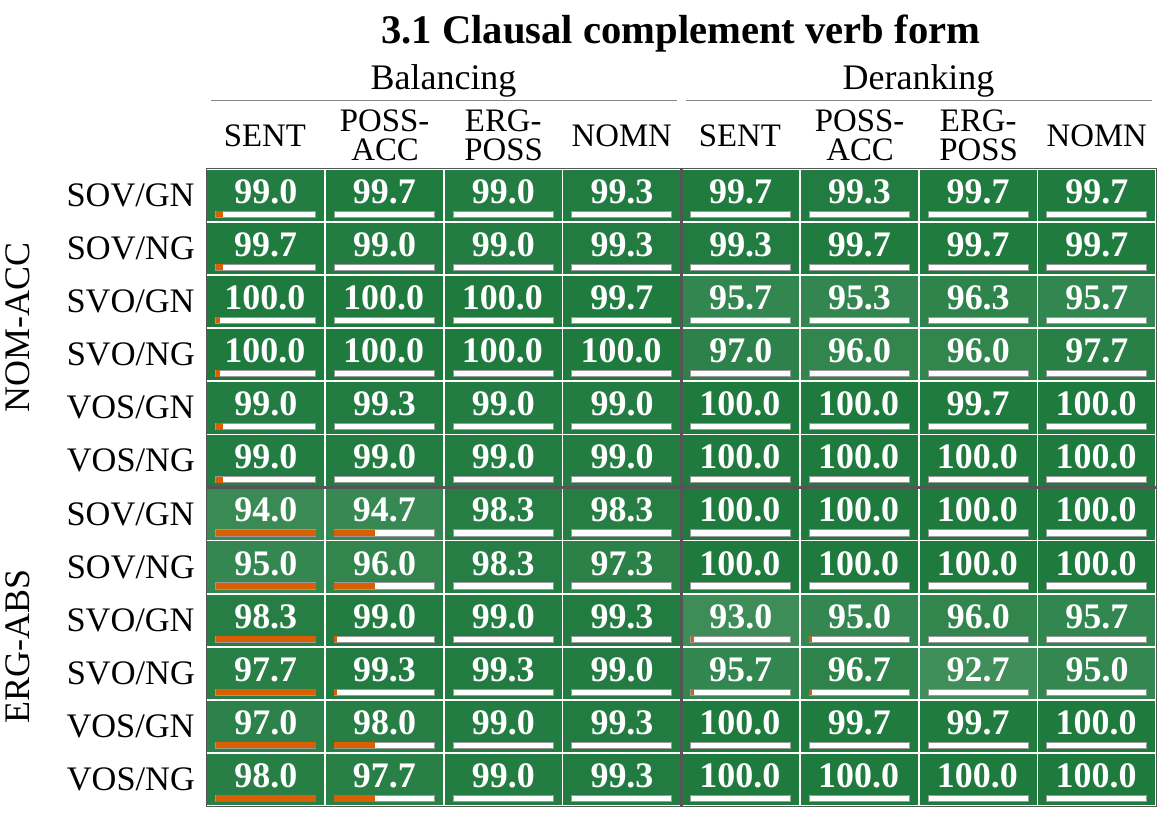}
\caption{Accuracy and BAD parse rate for 3.1 Clausal complement verb form.}
\label{fig:e1-14-clausal-v-form}
\end{figure}

Figure~\ref{fig:e1-14-clausal-v-form} shows that the models generally learned clausal-complement verb form well. The mean accuracy across the 96 configurations is 98.6\%, with a minimum of 92.7\% and a maximum of 100\%. The BAD parse rate ranges from 0\% to 100\%, with a mean of 8.5\%. Balancing and Deranking differ only slightly, with mean accuracies of 98.7\% and 98.4\%, respectively.

Among these configurations, Balancing + ERG-ABS + SENT has a BAD parse rate of 100\%, but its mean accuracy remains high at 96.7\%. In this configuration, once the embedded verb is perturbed to \textit{V-ing}, the whole embedded sequence can be reanalyzed by the grammar as an ANC functioning as matrix P. For example, the GOOD sentence \textit{Farmerca cow calves reports} is perturbed into \textit{Farmerca cow calveing reports}. The sequence \textit{cow calveing} can be analyzed as an ANC. At the same time, under ERG-ABS alignment, matrix P does not require an overt marker, and verbs such as \textit{reports} may also occur in ordinary transitive frames in the training corpus. Together, these factors make the perturbed sentence parsable.

Nevertheless, the models still maintain high accuracy in these configurations. This shows that the models distinguish well between the distribution of clausal-complement verbs and ordinary transitive verbs, and also learn the difference between the complement types they select.

\ancevalphenomenon{3.2}{Clausal Complement S Marker}{app:e1-15-clausal-s-marker}

This phenomenon tests whether the models learned the marker on embedded intransitive S in clausal complements. The GOOD sentence is a finite matrix clause whose complement is an embedded intransitive clause. In the source templates, the embedded S can be either a simple noun or an NP with a genitive possessor.

In all target grammars, the embedded S marker is zero, the same as the finite intransitive S marker. The BAD sentence is constructed by adding \textit{-ge} only to the head of the embedded S. An example is shown below:

\begin{exe}
\ex\label{ex:e1-15-clausal-s-marker}
\begin{xlist}
\ex GOOD: \textit{She knows bus stops}
\ex BAD:  \textit{She knows busge stops}
\end{xlist}
\end{exe}

\begin{figure}[htbp]
\centering
\includegraphics[width=\columnwidth]{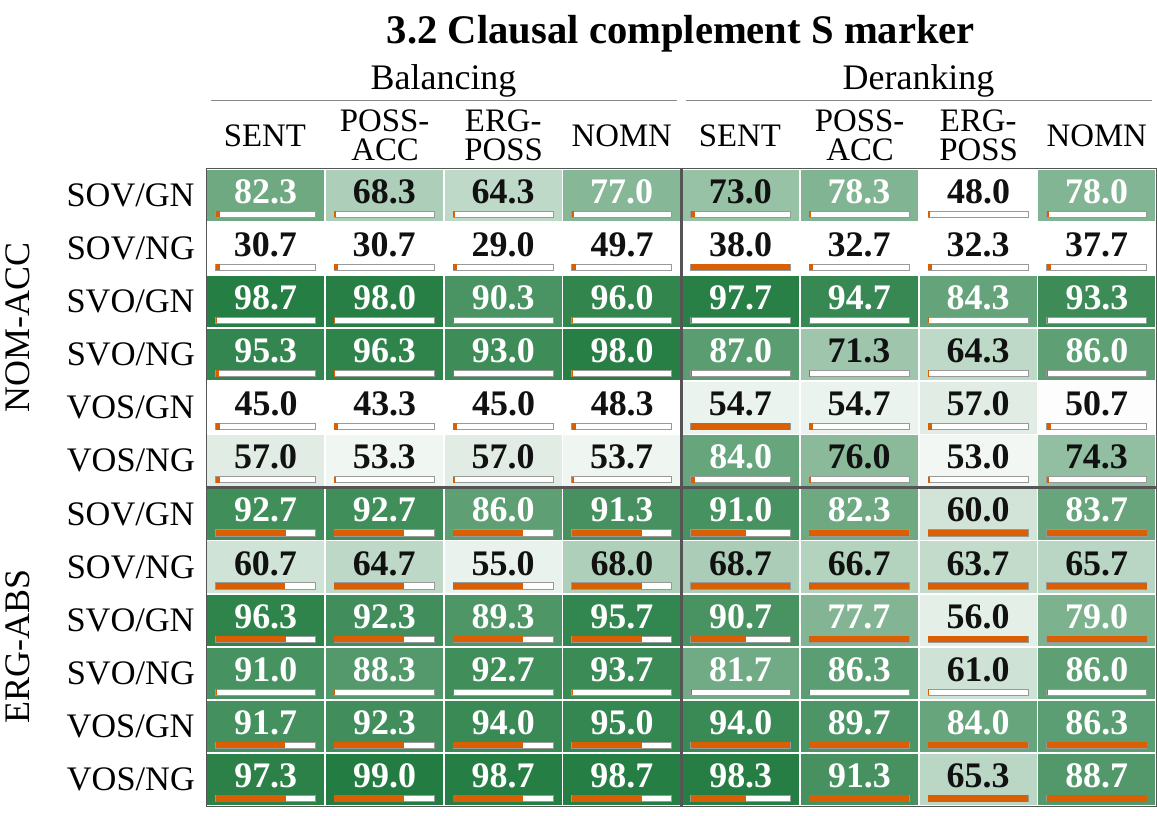}
\caption{Accuracy and BAD parse rate for 3.2 Clausal complement S marker.}
\label{fig:e1-15-clausal-s-marker}
\end{figure}

Figure~\ref{fig:e1-15-clausal-s-marker} shows that this phenomenon is harder than the corresponding finite-clause S marker phenomenon. The mean accuracy across the 96 configurations is 75.2\%, with a minimum of 29.0\% and a maximum of 99.0\%. The BAD parse rate ranges from 1.0\% to 100\%, with a mean of 37.3\%. Across typological parameters, SVO grammars perform best, with a mean accuracy of 87.6\%; VOS grammars are intermediate, at 74.1\%; and SOV grammars are weakest, at 63.8\%. The difference by alignment is also clear: ERG-ABS grammars reach 83.6\% accuracy, compared with 66.7\% for NOM-ACC grammars.

This phenomenon shows a strong asymmetry between BAD parse rate and model accuracy. The weakest configuration is SOV/NG + NOM-ACC, with a mean accuracy of only 35.1\%, even though its mean BAD parse rate is also low, at 15.9\%. For example, under this configuration, the GOOD sentence \textit{Coach player recovers believes} is perturbed into \textit{Coach playerge recovers believes}. The GOOD sentence roughly corresponds to ``the coach believes that the player recovers.'' In the BAD sentence, however, \textit{Coach playerge} forms an NG genitive NP on the surface, so the whole sequence may be interpreted as something like ``the player's coach recovers believes.'' Although this BAD sentence cannot be parsed by the grammar because \textit{believes} lacks a subject under this reanalysis, this local genitive-like pattern may still increase the model's score for the BAD sentence.

By contrast, SOV/GN + ERG-ABS + Deranking + POSS-ACC has a BAD parse rate of 100.0\%, but still reaches 82.3\% accuracy. Under this configuration, the GOOD sentence \textit{Heca guest leaveing thinks} is perturbed into \textit{Heca guestge leaveing thinks}. The GOOD sentence roughly corresponds to ``He thinks that the guest leaves.'' In the BAD sentence, \textit{guestge leaveing} can be analyzed as an ANC, with \textit{guestge} as the ANC-internal S. The whole sentence is therefore similar to ``He thinks [the guest's leaving].'' This analysis can be parsed by the grammar, but it reanalyzes the original clausal complement as an ANC in the matrix P position. The model still prefers the GOOD sentence, suggesting that it can distinguish the complement preferences of clausal-complement verbs from those of ordinary transitive verbs.

\ancevalphenomenon{3.3}{Clausal Complement A Marker}{app:e1-16-clausal-a-marker}

This phenomenon tests whether the models learned the marker on embedded transitive A in clausal complements. The GOOD sentence is a finite matrix clause whose complement is an embedded transitive clause. In the source templates, matrix A, embedded A, and embedded P can each be either a simple NP or an NP with a genitive possessor.

In clausal complements, embedded argument marking follows the finite-clause system. Therefore, the embedded A marker is determined by alignment. In NOM-ACC, embedded A is zero-marked; in ERG-ABS, embedded A bears \textit{-ca}. The BAD sentence is constructed by perturbing only the head of embedded A, replacing it with the A marker used under the corresponding ANC strategy. If this replacement would not create a surface difference, we instead use another marker. Specifically, we use \textit{-ge} for SENT, POSS-ACC, and NOMN, and \textit{-ob} for ERG-POSS. An example is shown below:

\begin{exe}
\ex\label{ex:e1-16-clausal-a-marker}
\begin{xlist}
\ex GOOD: \textit{Lawyer claims witness sees attackca}
\ex BAD:  \textit{Lawyer claims witnessge sees attackca}
\end{xlist}
\end{exe}

\begin{figure}[htbp]
\centering
\includegraphics[width=\columnwidth]{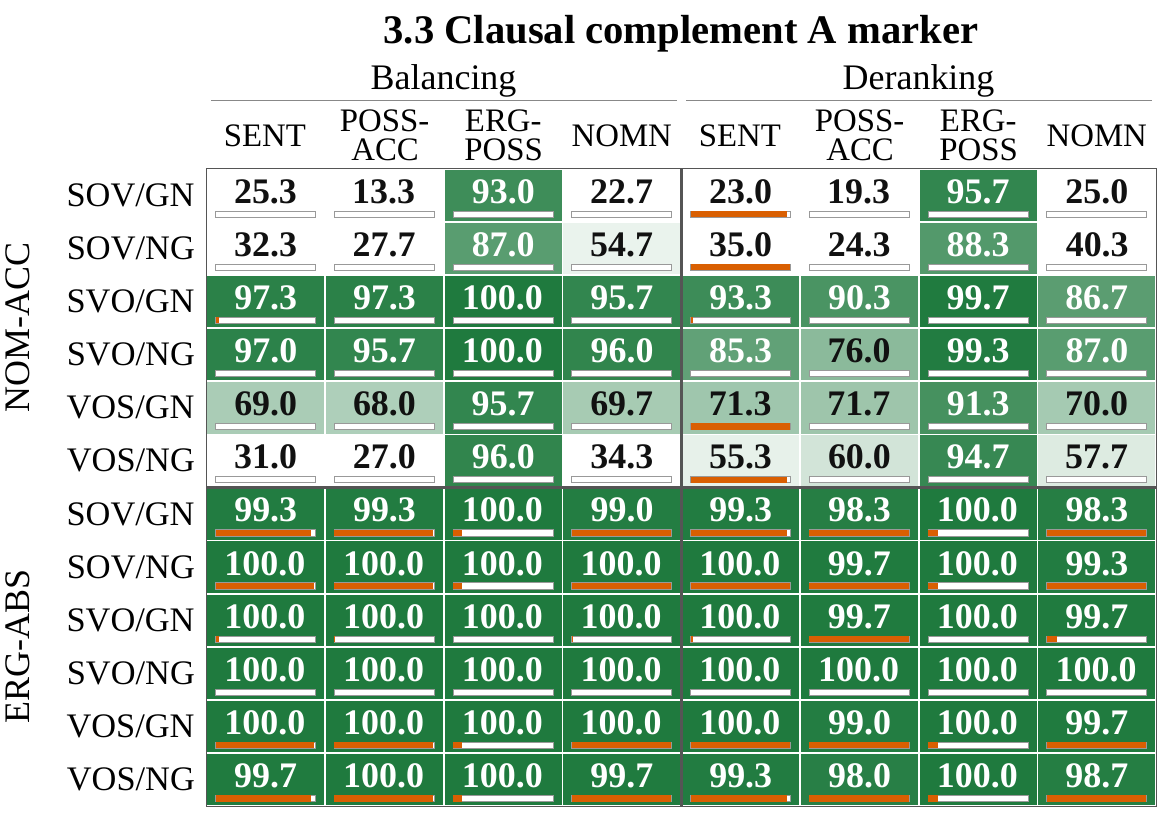}
\caption{Accuracy and BAD parse rate for 3.3 Clausal complement A marker.}
\label{fig:e1-16-clausal-a-marker}
\end{figure}

The results are shown in Figure~\ref{fig:e1-16-clausal-a-marker}. The mean accuracy across the 96 configurations is 83.9\%, with a minimum of 13.3\% and a maximum of 100\%. The BAD parse rate ranges from 0\% to 100\%, with a mean of 30.8\%.

Across typological parameters, the models learn this phenomenon well under ERG-ABS, with a mean accuracy of 99.7\%, even though the mean BAD parse rate is 53.3\%. By contrast, NOM-ACC is much harder, with a mean accuracy of only 68.0\%, even though the mean BAD parse rate is only 8.2\%. This mismatch has the same source as in Section~\ref{app:e1-15-clausal-s-marker}.

Within NOM-ACC, two cases perform relatively well. The first is the ERG-POSS strategy, with a mean accuracy of 95.1\%. The second is SVO word order, with a mean accuracy of 93.5\%. Other NOM-ACC configurations are much weaker. In particular, SOV/GN + NOM-ACC has a mean accuracy of only 39.7\%, while its BAD parse rate is only 12.0\%.

The main difficulty in NOM-ACC comes from the zero-marking of embedded A. Under the perturbation rule, in SENT, POSS-ACC, and NOMN, the BAD sentence adds \textit{-ge} to embedded A. Therefore, when embedded A and the following embedded P are adjacent, they can form a local genitive-like sequence on the surface. For example, under SOV/GN + NOM-ACC + Balancing + POSS-ACC, the GOOD sentence \textit{Lawyer witness attackca sees claims} is perturbed into \textit{Lawyer witnessge attackca sees claims}. The GOOD sentence roughly corresponds to ``the lawyer claims that the witness sees the attack.'' In the BAD sentence, \textit{witnessge attackca} is superficially close to an NP with a genitive possessor, roughly ``the witness's attack,'' but this analysis cannot form a complete grammatical sentence because the head noun \textit{attackca}, reanalyzed as the embedded S, should not bear the accusative marker \textit{-ca}. Thus, although this BAD sentence cannot be parsed by the grammar, the model can still be misled by the local surface pattern.

The better performance of ERG-POSS and SVO under NOM-ACC supports this interpretation. Under ERG-POSS, the BAD marker is \textit{-ob}, which cannot be confused with ordinary genitive marking. In SVO, the embedded order is A--V--P, so embedded A and embedded P are not adjacent.

\ancevalphenomenon{3.4}{Clausal Complement P Marker}{app:e1-17-clausal-p-marker}

This phenomenon tests whether the models learned the marker on embedded transitive P in clausal complements. The GOOD sentence is a finite matrix clause whose complement is an embedded transitive clause. In the source templates, matrix A, embedded A, and embedded P can each be either a simple NP or an NP with a genitive possessor.

In clausal complements, embedded argument marking follows the finite-clause system. Therefore, the embedded P marker is determined by alignment. In NOM-ACC, embedded P bears \textit{-ca}; in ERG-ABS, embedded P is zero-marked. The BAD sentence is constructed by perturbing only the head of embedded P, replacing it with the P marker used under the corresponding ANC strategy. If this replacement would not create a surface difference, we instead use another marker that is ungrammatical in that position. Specifically, we use \textit{-ge} for SENT, POSS-ACC, and ERG-POSS, and \textit{-ob} for NOMN. An example is shown below:

\begin{exe}
\ex\label{ex:e1-17-clausal-p-marker}
\begin{xlist}
\ex GOOD: \textit{Guard believes visitor opens safeca}
\ex BAD:  \textit{Guard believes visitor opens safeob}
\end{xlist}
\end{exe}

\begin{figure}[htbp]
\centering
\includegraphics[width=\columnwidth]{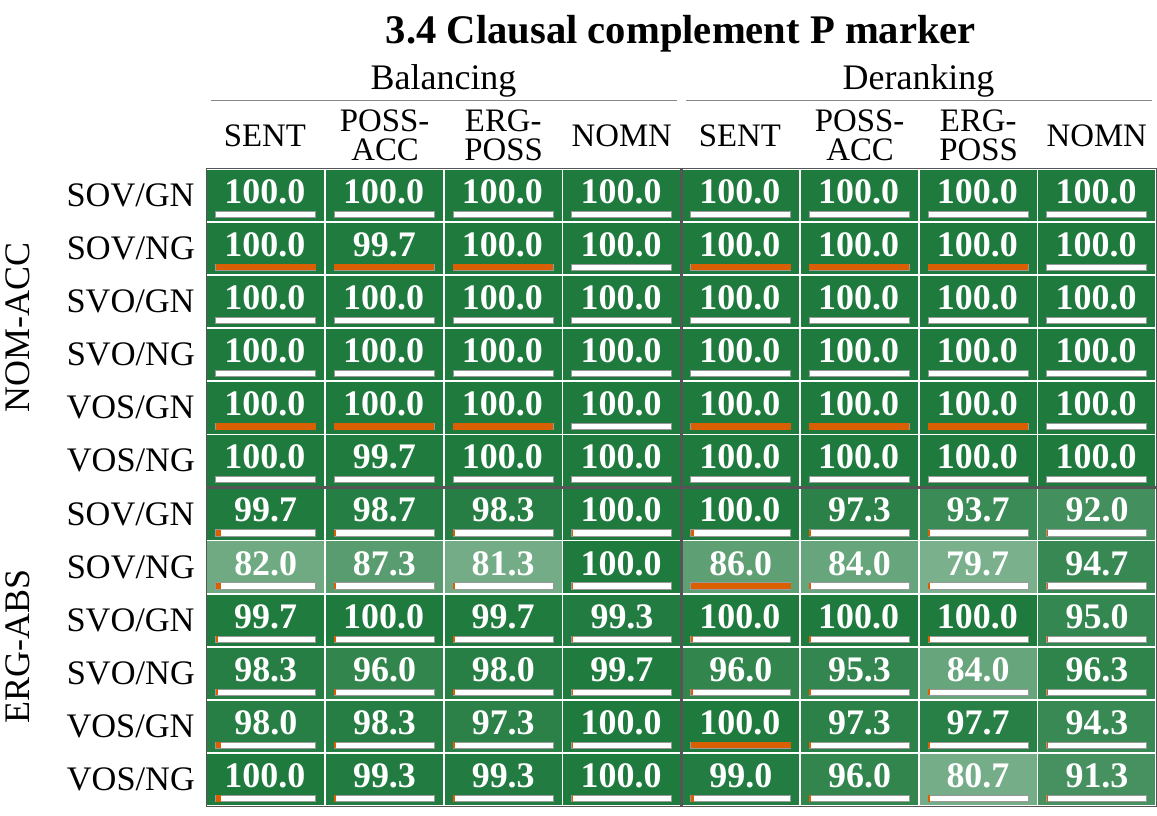}
\caption{Accuracy and BAD parse rate for 3.4 Clausal complement P marker.}
\label{fig:e1-17-clausal-p-marker}
\end{figure}

Figure~\ref{fig:e1-17-clausal-p-marker} shows that the models learned clausal complement P marking well overall. The mean accuracy across the 96 configurations is 97.7\%, with a minimum of 79.7\% and a maximum of 100\%, substantially higher than in Section~\ref{app:e1-16-clausal-a-marker}. The BAD parse rate ranges from 0\% to 100\%, with a mean of 15.6\%.

The error pattern mirrors Clausal Complement A Marker in Section~\ref{app:e1-16-clausal-a-marker}. NOM-ACC configurations all reach near-perfect accuracy, while ERG-ABS is relatively weaker, with a mean accuracy of 95.4\%. The weakest case is SOV/NG + ERG-ABS + Deranking + ERG-POSS, with a mean accuracy of 79.7\%. For example, under SOV/NG + ERG-ABS + Deranking + ERG-POSS, the GOOD sentence \textit{Guardca visitorca safe opening believes} is perturbed into \textit{Guardca visitorca safege opening believes}. The reason for the errors here is the same as in the previous section. After adding \textit{-ge}, the local sequence inside the embedded transitive clause becomes superficially similar to a possessor--head relation.

However, the models learn this phenomenon better than the previous one. A potential reason is that in Clausal Complement A Marker, the argument changed into a genitive-like form is embedded A, as in \textit{witnessge attackca}. An animate A is semantically natural as a genitive possessor. By contrast, in the present phenomenon, embedded P becomes a genitive-like dependent, as in the NG sequence \textit{visitorca safege}. Since embedded P is often an inanimate object, reinterpreting it as a genitive possessor is semantically more restricted, so the overall interference is weaker.

\ancevalphenomenon{3.5}{Clausal Complement--Matrix Verb Order}{app:e1-18-clausal-comp-v-order}

This phenomenon tests whether the models learned the relative order between the clausal complement as a whole and the matrix verb. The GOOD sentence is a finite matrix clause whose complement is an embedded intransitive or transitive clause. The source templates include both embedded IV and embedded TV complements. The BAD sentence is constructed by swapping the whole embedded clausal complement span with the matrix verb, while leaving the internal structure of the embedded clause unchanged. An example is shown below:

\begin{exe}
\ex\label{ex:e1-18-clausal-comp-v-order}
\begin{xlist}
\ex GOOD: \textit{Driver claims car stalls}
\ex BAD:  \textit{Driver car stalls claims}
\end{xlist}
\end{exe}

\begin{figure}[htbp]
\centering
\includegraphics[width=\columnwidth]{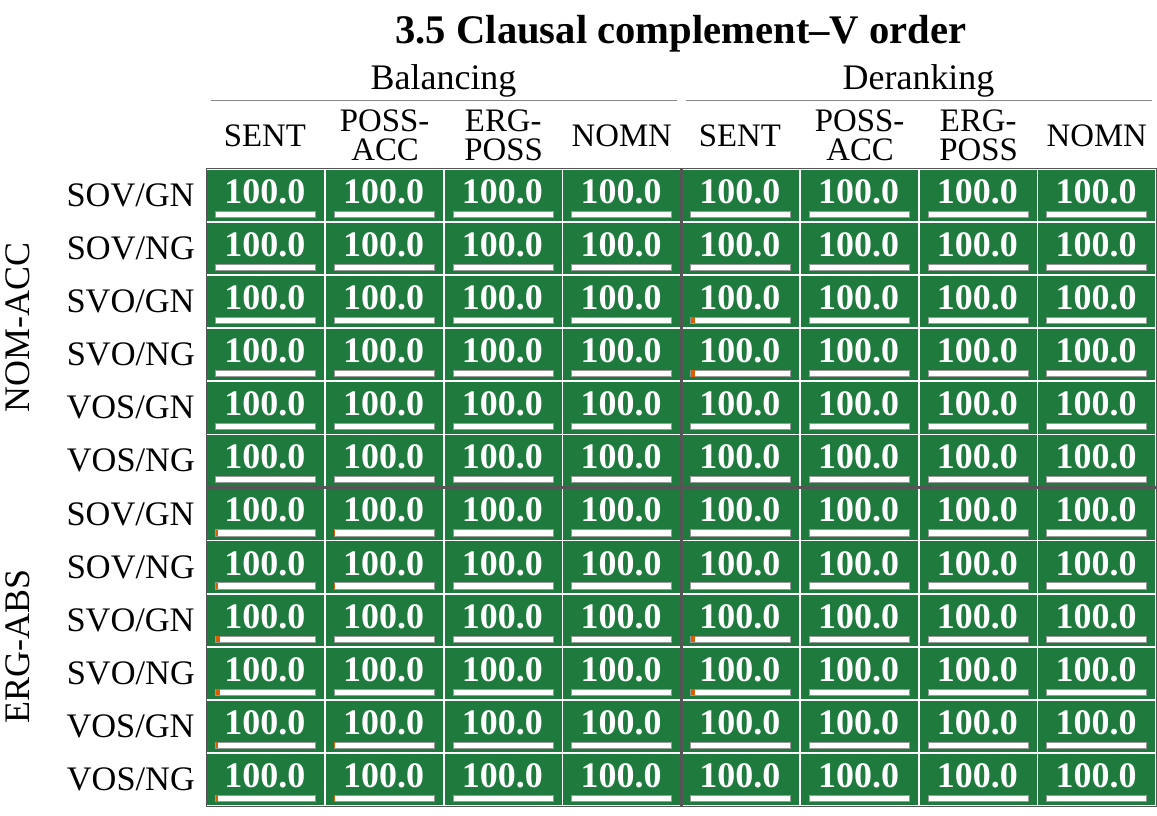}
\caption{Accuracy and BAD parse rate for 3.5 Clausal complement--V order.}
\label{fig:e1-18-clausal-comp-v-order}
\end{figure}

Figure~\ref{fig:e1-18-clausal-comp-v-order} shows that the models learned this phenomenon almost perfectly. The mean accuracy across the 96 configurations is 100.0\%, with a minimum of 100.0\% and a maximum of 100\%. The BAD parse rate is also low, with a mean of 0.4\% and a maximum of 4.0\%.

\ancevalphenomenon{3.6}{Clausal Complement CV Valency}{app:e1-19-clausal-cv-valency}

This phenomenon tests whether the models learned which matrix verbs can select a clausal complement. The GOOD sentence is a finite matrix clause whose matrix verb is a clausal-complement verb, and whose complement is an embedded intransitive or transitive clause. The source templates cover both embedded IV and embedded TV complements.

This phenomenon uses the MegaAcceptability dataset \citep{white_rawlins_2020} to construct a valency pair table. The GOOD verb is a clausal-complement verb that is highly acceptable in a clausal-complement frame in human judgments. The BAD verb is an ordinary transitive verb that is highly acceptable in a transitive frame, but has low acceptability in a clausal-complement frame. The BAD sentence is constructed by replacing the matrix verb stem. An example is shown below:

\begin{exe}
\ex\label{ex:e1-19-clausal-cv-valency}
\begin{xlist}
\ex GOOD: \textit{Scientist speculates patient follows instructionca}
\ex BAD:  \textit{Scientist names patient follows instructionca}
\end{xlist}
\end{exe}

\begin{figure}[htbp]
\centering
\includegraphics[width=\columnwidth]{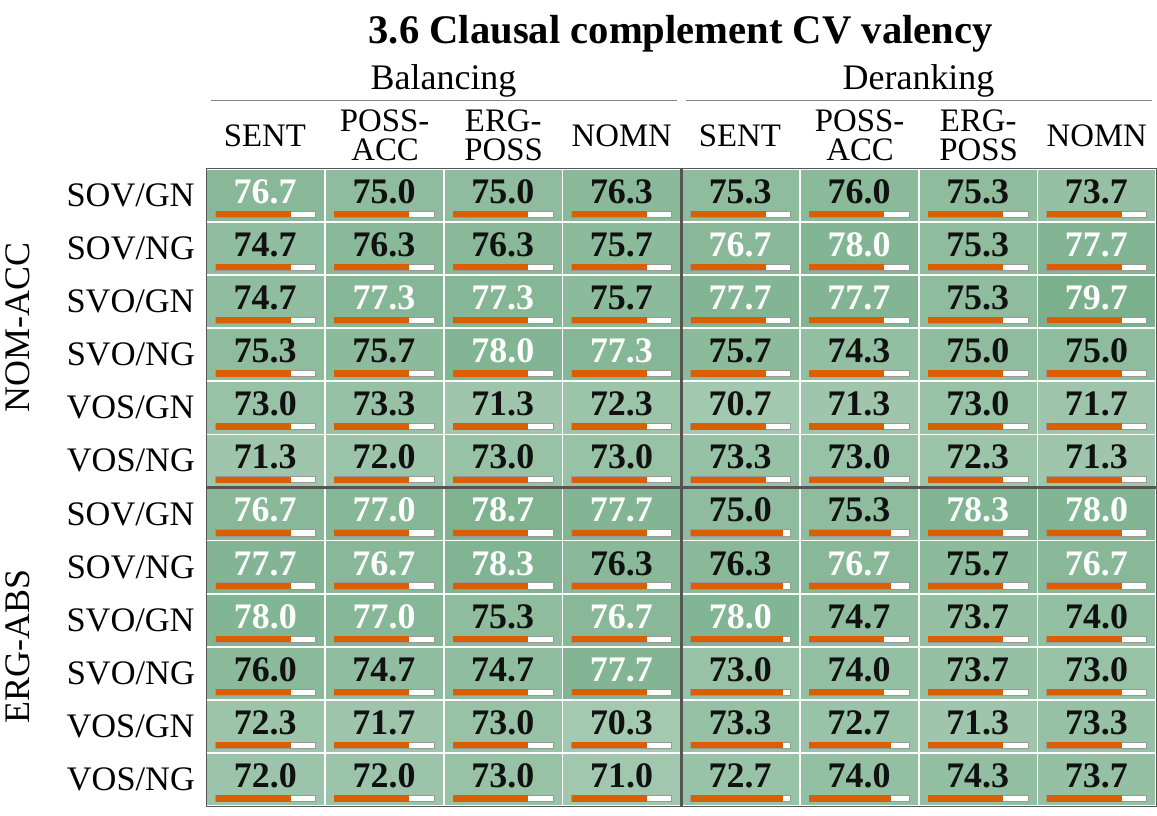}
\caption{Accuracy and BAD parse rate for 3.6 Clausal complement CV valency.}
\label{fig:e1-19-clausal-cv-valency}
\end{figure}

Figure~\ref{fig:e1-19-clausal-cv-valency} shows that the mean accuracy across the 96 configurations is 74.9\%, with a minimum of 70.3\% and a maximum of 79.7\%. The BAD parse rate is high, with a mean of 76.4\% and a range of 75.0\%--92.0\%, showing that many BAD verbs also take clausal complements in the training corpus.

Compared with the other valency phenomena, 3.6 is close to Transitive V Valency in Section~\ref{app:e1-10-tran-v-valency} (76.2\%) and higher than Intransitive V Valency in Section~\ref{app:e1-9-intran-v-valency} (70.4\%). This suggests that the models learned the distinction between CVs and TVs reasonably well. Overall, the models learned part of the CV/TV distributional distinction, but lexical valency remains a difficult phenomenon.

\subsubsection{ANC Internal}
\label{app:anc-internal}

\ancevalphenomenon{4.1}{ANC Verb Form}{app:e1-20-anc-v-form}

This phenomenon tests whether the models learned the verb form inside ANCs. The source templates are finite transitive clauses in which the ANC functions as either matrix A or matrix P. Since ANCs allow argument omission, the ANC itself covers bare IV ANCs, IV+S ANCs, bare TV ANCs, TV+A ANCs, TV+P ANCs, and TV+A+P ANCs.

In all target grammars, the ANC verb form is the nonfinite form \textit{V-ing}. The BAD sentence is constructed by changing only the target ANC verb from \textit{V-ing} to the finite form \textit{V-s}. An example is shown below:

\begin{exe}
\ex\label{ex:e1-20-anc-v-form}
\begin{xlist}
\ex GOOD: \textit{Engineer checks techniciange measureingca}
\ex BAD:  \textit{Engineer checks techniciange measuresca}
\end{xlist}
\end{exe}

\begin{figure}[htbp]
\centering
\includegraphics[width=\columnwidth]{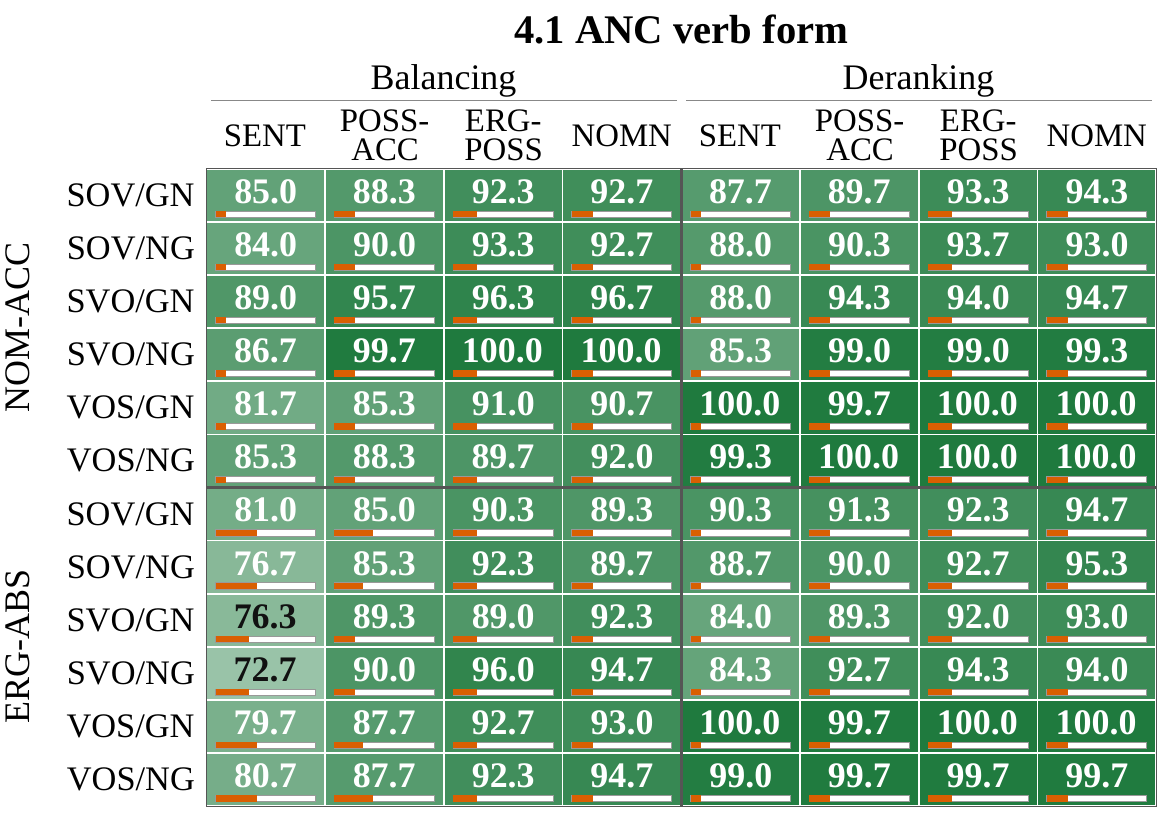}
\caption{Accuracy and BAD parse rate for 4.1 ANC verb form.}
\label{fig:e1-20-anc-v-form}
\end{figure}

Figure~\ref{fig:e1-20-anc-v-form} shows that the models generally learned ANC verb form. The mean accuracy across the 96 configurations is 91.9\%, with a minimum of 72.7\% and a maximum of 100\%. The BAD parse rate ranges from 10.0\% to 41.0\%, with a mean of 21.3\%.

Across parameters, Deranking grammars perform better, with a mean accuracy of 94.6\%, while Balancing grammars are weaker, at 89.3\%. ANC strategy also has an effect. SENT is the weakest strategy, with a mean accuracy of 86.4\%; POSS-ACC, ERG-POSS, and NOMN reach 92.0\%, 94.4\%, and 94.8\%, respectively. The weakest group is Balancing + SENT, with a mean accuracy of 81.6\%.

Overall, the models learned that verbs inside ANCs should take the nonfinite form. The main difficulty appears in the Balancing + SENT configuration. In these grammars, clausal complements use finite verb forms, and the SENT strategy makes ANCs more similar to clausal structures, so finite \textit{V-s} can make the BAD sentence look more like a clausal complement.

\ancevalphenomenon{4.2}{ANC S Marker}{app:e1-21-anc-s-marker}

This phenomenon tests whether the models learned the marker on S inside intransitive ANCs. The GOOD sentence is a finite transitive clause in which matrix P is an intransitive ANC with an overt S.

In the target grammars, the ANC-internal S marker is determined by ANC strategy. Under SENT, ANC-internal S has zero marking. Under POSS-ACC, ERG-POSS, and NOMN, ANC-internal S receives the genitive-like marker \textit{-ge}. The BAD sentence is constructed by changing only the marker on the ANC-internal S head, adding \textit{-ge} under SENT and removing \textit{-ge} under the other three strategies. An example is shown below:

\begin{exe}
\ex\label{ex:e1-21-anc-s-marker}
\begin{xlist}
\ex GOOD: \textit{Farmer notices soilge contractingca}
\ex BAD:  \textit{Farmer notices soil contractingca}
\end{xlist}
\end{exe}

\begin{figure}[htbp]
\centering
\includegraphics[width=\columnwidth]{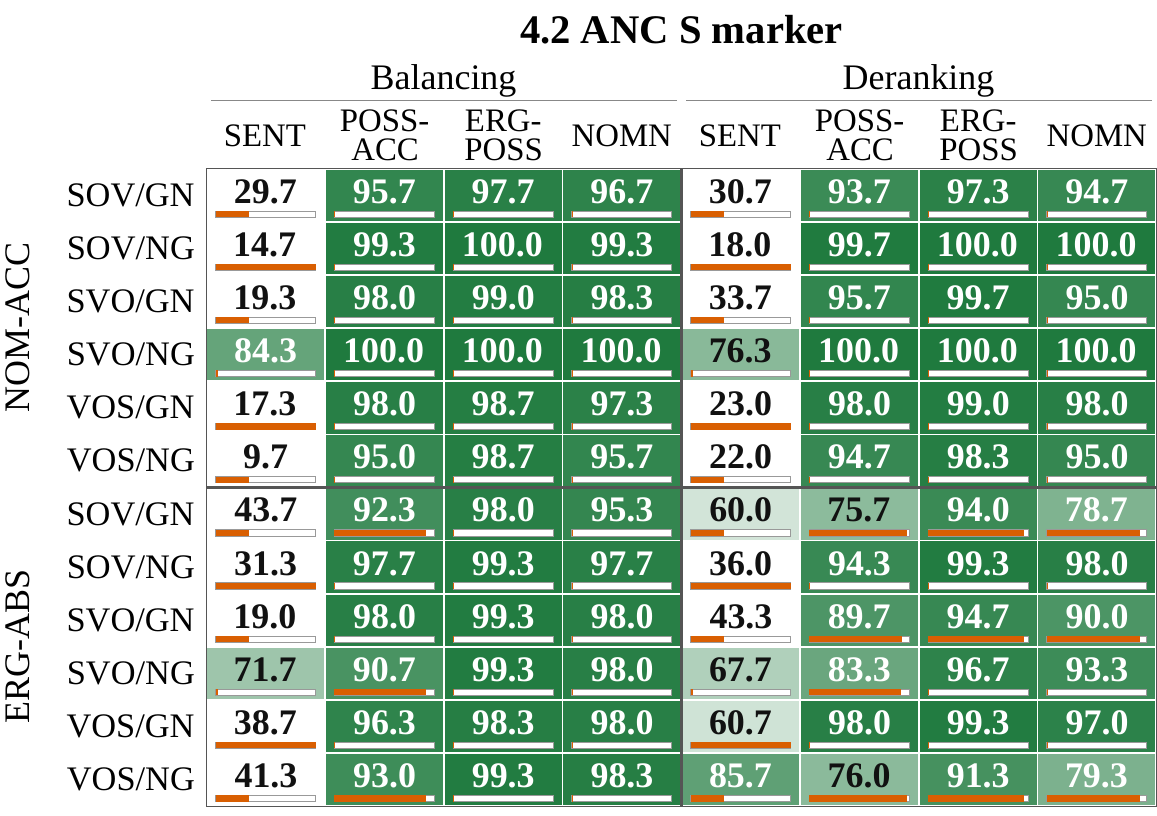}
\caption{Accuracy and BAD parse rate for 4.2 ANC S marker.}
\label{fig:e1-21-anc-s-marker}
\end{figure}

Figure~\ref{fig:e1-21-anc-s-marker} shows the results for this phenomenon. The mean accuracy across the 96 configurations is 82.1\%, with a minimum of 9.7\% and a maximum of 100\%. The BAD parse rate ranges from 1.0\% to 100\%, with a mean of 25.9\%.

The main contrast comes from ANC strategy. SENT has a mean accuracy of only 40.7\%, with a BAD parse rate of 50.2\%. By contrast, POSS-ACC, ERG-POSS, and NOMN reach mean accuracies of 93.9\%, 98.2\%, and 95.5\%, with BAD parse rates of 28.1\%, 12.9\%, and 12.5\%, respectively. The other parameters have smaller overall effects.

The low accuracy under SENT is mainly because ANC-internal S and finite-clause S have the same marking. In both environments, S has zero marking. Therefore, when the BAD sentence adds \textit{-ge} to ANC-internal S, the surface form can instead resemble an ordinary genitive NP. For example, under SVO/GN + NOM-ACC + Balancing + SENT, the GOOD sentence \textit{Host welcomes guest arriveingca} is perturbed into \textit{Host welcomes guestge arriveingca}. The BAD sentence is roughly like ``the host welcomes [the guest's arriving]'', where \textit{guest} is interpreted as the possessor of \textit{arriveing}. This surface form is natural, but under the SENT strategy, ANC-internal S cannot take \textit{-ge}, so the BAD sentence is ungrammatical in this grammar.

\ancevalphenomenon{4.3}{ANC A Marker}{app:e1-22-anc-a-marker}

This phenomenon tests whether the models learned the marker on A inside transitive ANCs. The GOOD sentence is a finite transitive clause in which matrix P is a transitive ANC with overt A and overt P.

In the target grammars, the ANC-internal A marker is determined by ANC strategy and alignment. Under SENT, ANC-internal A takes the corresponding finite A marker. Under POSS-ACC and NOMN, ANC-internal A receives the genitive-like marker \textit{-ge}. Under ERG-POSS, ANC-internal A receives \textit{-ob}. The BAD sentence is constructed by changing only the marker on the ANC-internal A head. Under SENT, it is changed to \textit{-ge}; under the other three strategies, it is changed back to the corresponding finite A marker. An example is shown below:

\begin{exe}
\ex\label{ex:e1-22-anc-a-marker}
\begin{xlist}
\ex GOOD: \textit{Editor accepts translatorge renderingca titleob}
\ex BAD:  \textit{Editor accepts translator renderingca titleob}
\end{xlist}
\end{exe}

\begin{figure}[htbp]
\centering
\includegraphics[width=\columnwidth]{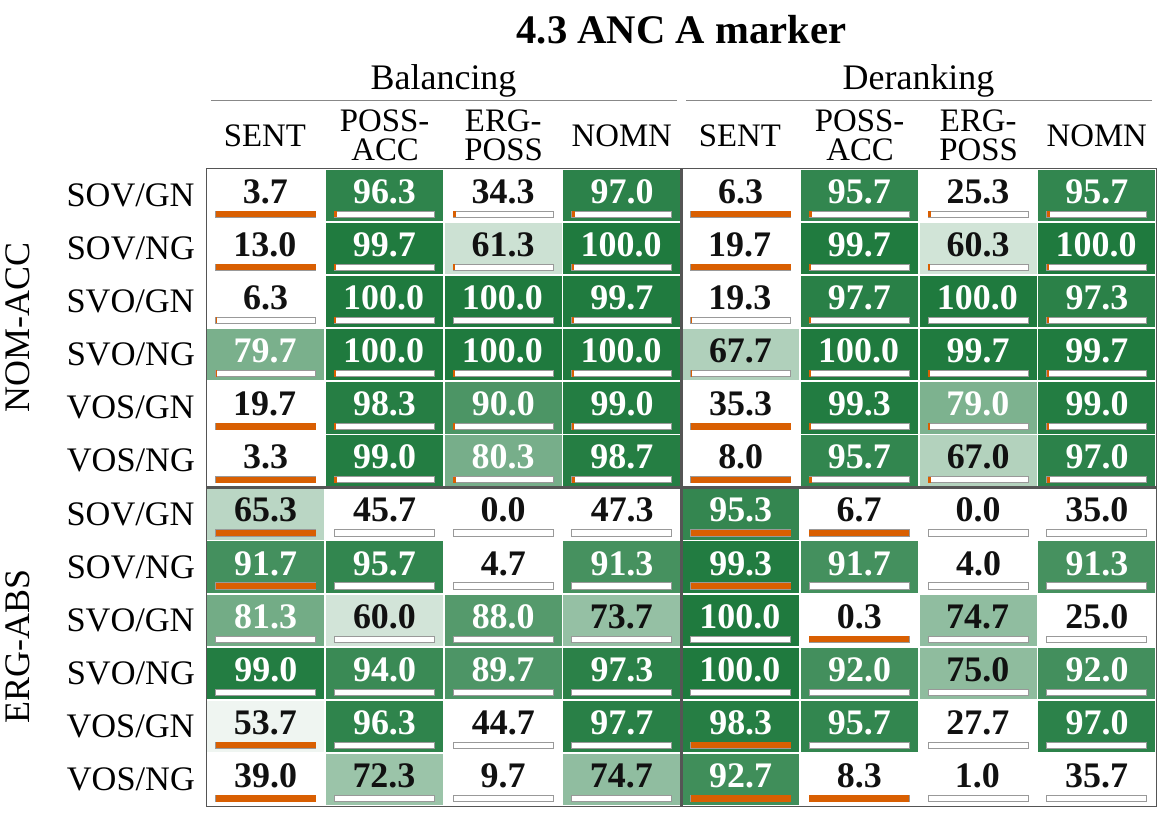}
\caption{Accuracy and BAD parse rate for 4.3 ANC A marker.}
\label{fig:e1-22-anc-a-marker}
\end{figure}

Figure~\ref{fig:e1-22-anc-a-marker} shows the results for this phenomenon. The mean accuracy across the 96 configurations is 68.7\%, with a minimum of 0.0\% and a maximum of 100\%. The BAD parse rate ranges from 0.0\% to 100\%, with a mean of 20.7\%.

This phenomenon is harder than ANC S Marker. Across ANC strategies, POSS-ACC and NOMN perform relatively well, with mean accuracies of 80.8\% and 85.0\%, respectively. SENT and ERG-POSS are much weaker, at 54.1\% and 54.8\%. Alignment also has an effect, with NOM-ACC at 73.8\% and ERG-ABS at 63.6\%.

The errors fall into two main types. The first type appears under SENT + NOM-ACC. In these configurations, ANC-internal A has zero marking in the GOOD sentence, while the BAD sentence adds \textit{-ge}. This makes the ANC portion of the BAD sentence resemble an ordinary genitive + \textit{V-ing} nominal pattern, similar to ANC S Marker in Section~\ref{app:e1-21-anc-s-marker}.

The second type appears under non-SENT strategies with ERG-ABS, especially ERG-POSS. ERG-POSS + ERG-ABS has a mean accuracy of only 34.9\%, but its BAD parse rate is 0.0\%. For example, under SOV/GN + ERG-ABS + Balancing + ERG-POSS, the GOOD sentence \textit{Supervisorca traineeob tablege cleaning observes} is perturbed into \textit{Supervisorca traineeca tablege cleaning observes}. On the surface, \textit{cleaning} can only be an ANC head in a Balancing grammar, and under ERG-POSS, ANC-internal A should take \textit{-ob}, not \textit{-ca}. Therefore, this type of BAD sentence cannot be parsed by the grammar and does not have an obvious alternative structural analysis. It remains unclear why the model prefers this form.

\ancevalphenomenon{4.4}{ANC P Marker}{app:e1-23-anc-p-marker}

This phenomenon tests whether the models learned the marker on P inside transitive ANCs. The GOOD sentence is a finite transitive clause in which matrix P is a transitive ANC with overt A and overt P.

In the target grammars, the ANC-internal P marker is determined by ANC strategy and alignment. Under SENT and POSS-ACC, ANC-internal P takes the corresponding finite P marker. Under ERG-POSS, ANC-internal P receives the genitive-like marker \textit{-ge}. Under NOMN, ANC-internal P receives \textit{-ob}. The BAD sentence is constructed by changing only the marker on the ANC-internal P head. Under SENT and POSS-ACC, it is changed to \textit{-ge}; under ERG-POSS and NOMN, it is changed back to the corresponding finite P marker. An example is shown below:

\begin{exe}
\ex\label{ex:e1-23-anc-p-marker}
\begin{xlist}
\ex GOOD: \textit{Reporter notes scientistge measureingca sampleob}
\ex BAD:  \textit{Reporter notes scientistge measureingca sampleca}
\end{xlist}
\end{exe}

\begin{figure}[htbp]
\centering
\includegraphics[width=\columnwidth]{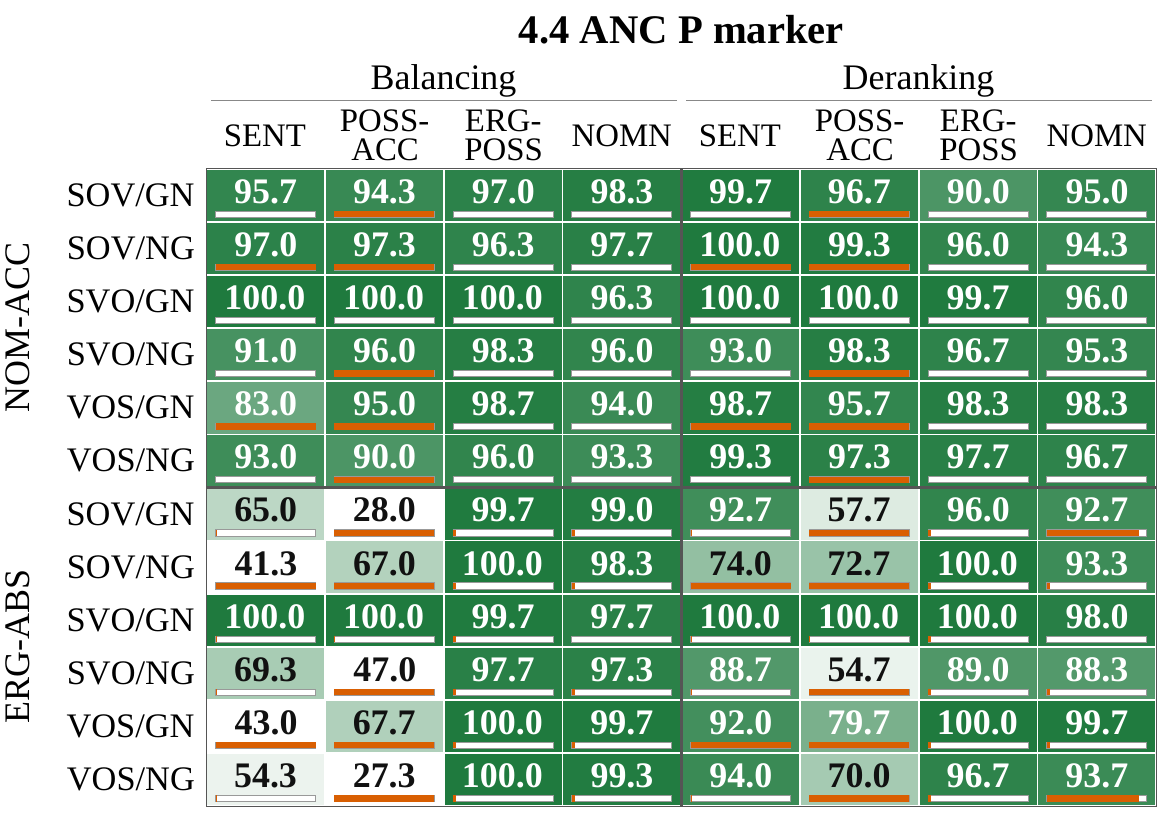}
\caption{Accuracy and BAD parse rate for 4.4 ANC P marker.}
\label{fig:e1-23-anc-p-marker}
\end{figure}

Figure~\ref{fig:e1-23-anc-p-marker} shows the results for this phenomenon. The mean accuracy across the 96 configurations is 90.1\%, with a minimum of 27.3\% and a maximum of 100\%. The BAD parse rate ranges from 0.0\% to 100\%, with a mean of 31.8\%.

This phenomenon is clearly easier than ANC A Marker. ERG-POSS and NOMN perform best, with mean accuracies of 97.6\% and 96.2\%, respectively. SENT is also relatively high, at 86.0\%. POSS-ACC is the weakest strategy, with a mean accuracy of 80.5\%. Across alignment, NOM-ACC reaches 96.4\%, compared with 83.8\% for ERG-ABS.

The main errors are concentrated in POSS-ACC + ERG-ABS. This combination has a mean accuracy of only 64.3\%, with a BAD parse rate of 83.5\%. In these configurations, ANC-internal P and finite P are both zero-marked, and the BAD sentence adds \textit{-ge} to P. For example, under SOV/GN + ERG-ABS + Balancing + POSS-ACC, the GOOD sentence \textit{Reporterca scientistge sample measureing notes} is perturbed into \textit{Reporterca scientistge samplege measureing notes}. The GOOD sentence roughly corresponds to ``the reporter notes [the scientist's measuring of the sample].'' In the BAD sentence, \textit{scientistge samplege} forms a genitive NP, roughly ``the scientist's sample'', a semantically natural possessive relation between the original A and P. This NP can function as ANC-internal S, or as A with P omitted, making the BAD sentence parsable while changing the original participant roles.

\ancevalphenomenon{4.5}{ANC S--V Order}{app:e1-24-anc-sv-order}

This phenomenon tests whether the models learned the relative order between overt S and the ANC verb inside intransitive ANCs. The GOOD sentence is a finite transitive clause in which matrix P is an intransitive ANC with an overt S.

In the target grammars, ANC-internal S--V order is determined by the effective ANC intransitive order. As shown in Table~\ref{tab:anc-word-order}, this order is determined jointly by clause word order, NP word order, and ANC strategy. The BAD sentence is constructed by swapping the ANC-internal S token and the ANC verb token. An example is shown below:

\begin{exe}
\ex\label{ex:e1-24-anc-sv-order}
\begin{xlist}
\ex GOOD: \textit{Farmer notices soilge contractingca}
\ex BAD:  \textit{Farmer notices contractingca soilge}
\end{xlist}
\end{exe}

\begin{figure}[htbp]
\centering
\includegraphics[width=\columnwidth]{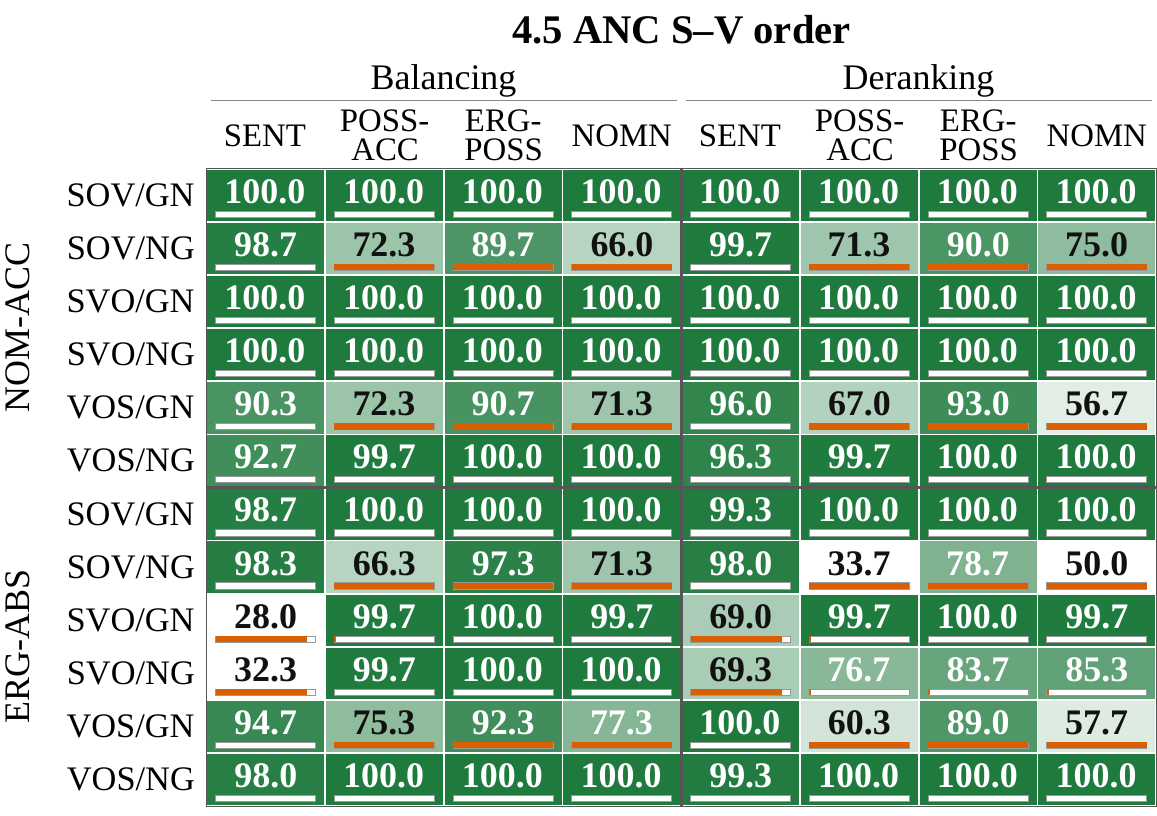}
\caption{Accuracy and BAD parse rate for 4.5 ANC S--V order.}
\label{fig:e1-24-anc-sv-order}
\end{figure}

Figure~\ref{fig:e1-24-anc-sv-order} shows that the models generally learned this phenomenon well. The mean accuracy across the 96 configurations is 90.3\%, with a minimum of 28.0\% and a maximum of 100\%. The BAD parse rate ranges from 0.0\% to 100\%, with a mean of 28.9\%.

The main problems fall into two types. The first is the combination of SOV/NG or VOS/GN word order and non-SENT strategies, especially POSS-ACC and NOMN. In SOV/NG, the non-SENT strategies have a mean accuracy of 71.8\%, with a BAD parse rate of 100\%. In VOS/GN, the non-SENT strategies have a mean accuracy of 75.2\%, also with a BAD parse rate of 100\%. By contrast, the SENT strategy performs much better in these two profiles. SOV/NG + SENT has a mean accuracy of 98.7\%, and VOS/GN + SENT has a mean accuracy of 95.2\%; both have a BAD parse rate of 0\%.

This error pattern follows from how ANC-internal order is defined. For non-SENT strategies, ANCs have both clausal and nominal properties, so their internal order is determined jointly by clause word order and NP word order. In SOV/NG and VOS/GN, the two orders point in different directions. In these cases, the experimental grammars align non-SENT ANCs primarily with NP word order. As a result, ANC-internal order can depart from the more familiar linear patterns of finite clauses.

The second difficulty appears under SVO + SENT, especially with ERG-ABS alignment. In the SVO + SENT + ERG-ABS configurations, the mean accuracy is only 49.7\%, while the BAD parse rate is 91.0\%.

This problem comes from ANC argument omission. Since arguments can be omitted inside ANCs, if an SVO-like ANC contains only one overt argument, that argument can sometimes appear on either side of the verb. For verbs with transitive/intransitive alternations, if the overt argument can naturally be interpreted either as S or as P, both the GOOD and BAD sentences may receive grammatical and semantically natural analyses. For example, the GOOD sentence \textit{Scientistca observes cloud forming} is perturbed into \textit{Scientistca observes forming cloud}. The GOOD sentence roughly means ``the scientist observes [the cloud forming]'', while the BAD sentence can be reanalyzed as ``the scientist observes [the forming of the cloud].'' Thus, although a high BAD parse rate does not necessarily mean that the model failed to learn the grammar, it shows that ANC-internal order is less stable in this configuration.

\ancevalphenomenon{4.6}{ANC A--V Order}{app:e1-25-anc-av-order}

This phenomenon tests whether the models learned the relative order between A and the ANC verb inside transitive ANCs. The GOOD sentence is a finite transitive clause in which matrix P is a transitive ANC with overt A and overt P.

In the target grammars, ANC-internal A--V order is determined by the effective ANC transitive order. As shown in Table~\ref{tab:anc-word-order}, this order is determined jointly by clause word order, NP word order, and ANC strategy. The BAD sentence is constructed by swapping the ANC-internal A span with the chunk containing the ANC verb and P. An example is shown below:

\begin{exe}
\ex\label{ex:e1-25-anc-av-order}
\begin{xlist}
\ex GOOD: \textit{Nurse reports patientge takingca medicineob}
\ex BAD:  \textit{Nurse reports takingca medicineob patientge}
\end{xlist}
\end{exe}

\begin{figure}[htbp]
\centering
\includegraphics[width=\columnwidth]{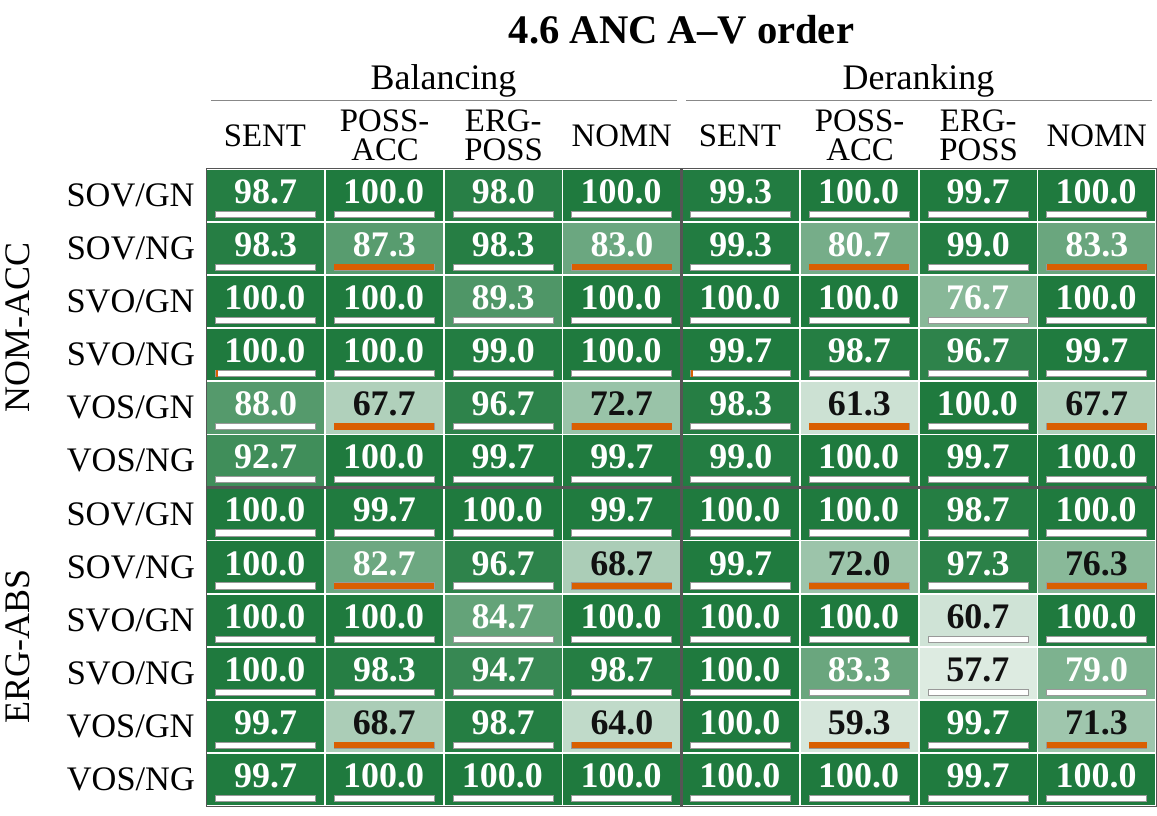}
\caption{Accuracy and BAD parse rate for 4.6 ANC A--V order.}
\label{fig:e1-25-anc-av-order}
\end{figure}

Figure~\ref{fig:e1-25-anc-av-order} shows that the models generally learned this phenomenon well. The mean accuracy across the 96 configurations is 93.1\%, with a minimum of 57.7\% and a maximum of 100\%. The BAD parse rate ranges from 0.0\% to 100\%, with a mean of 16.7\%.

The main source of errors is similar to ANC S--V Order in Section~\ref{app:e1-24-anc-sv-order}: ANC-internal A--V order can conflict with finite-clause A--V order. The affected combinations are SVO/GN + ERG-POSS, VOS/GN + POSS-ACC/NOMN, and SOV/NG + POSS-ACC/NOMN, with mean accuracies of 77.8\%, 64.2\%/68.9\%, and 80.7\%/77.8\%, respectively. By contrast, the remaining combinations have a mean accuracy of 98.1\%.

\ancevalphenomenon{4.7}{ANC V--P Order}{app:e1-26-anc-vp-order}

This phenomenon tests whether the models learned the relative order between the ANC verb and P inside transitive ANCs. The GOOD sentence is a finite transitive clause in which matrix P is a transitive ANC with overt A and overt P.

In the target grammars, ANC-internal V--P order is determined by the effective ANC transitive order, as shown in Table~\ref{tab:anc-word-order}. The BAD sentence is constructed by swapping the ANC verb token and the ANC-internal P head. An example is shown below:

\begin{exe}
\ex\label{ex:e1-26-anc-vp-order}
\begin{xlist}
\ex GOOD: \textit{Supervisor observes traineege cleaningca tableob}
\ex BAD:  \textit{Supervisor observes traineege tableob cleaningca}
\end{xlist}
\end{exe}

\begin{figure}[htbp]
\centering
\includegraphics[width=\columnwidth]{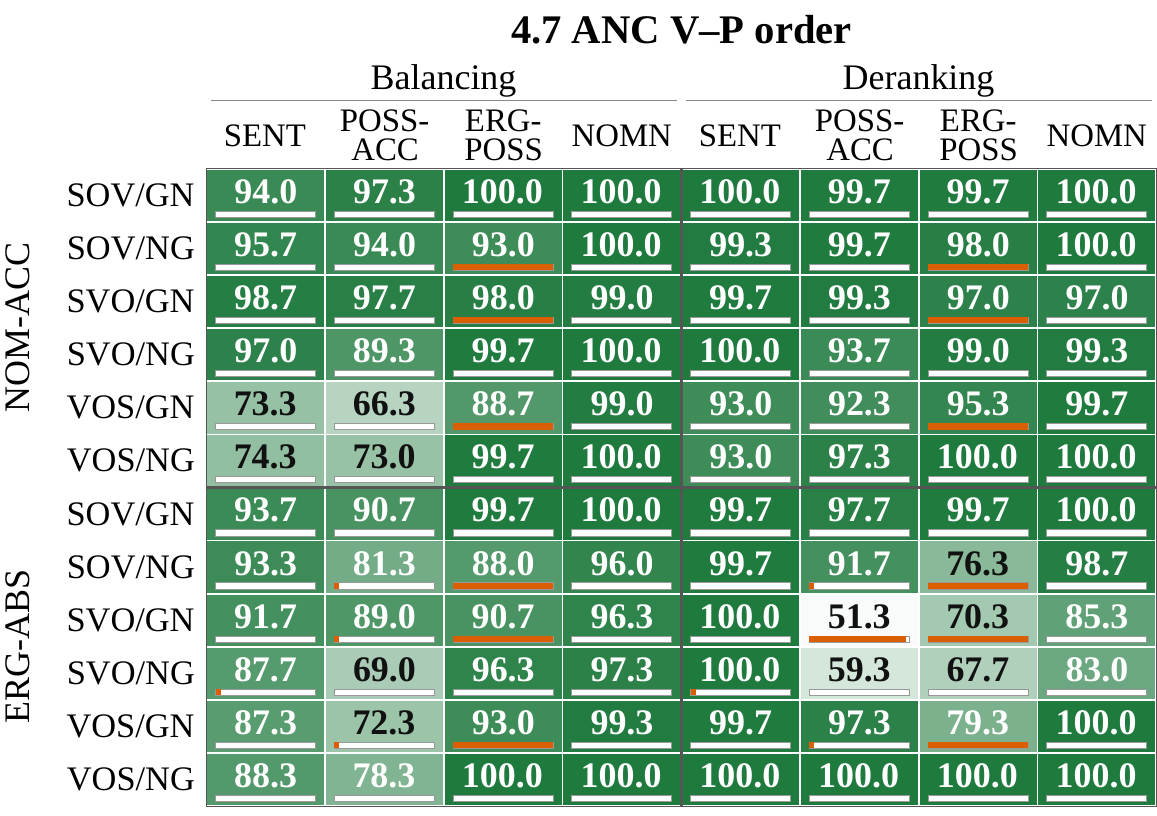}
\caption{Accuracy and BAD parse rate for 4.7 ANC V--P order.}
\label{fig:e1-26-anc-vp-order}
\end{figure}

Figure~\ref{fig:e1-26-anc-vp-order} shows that the models generally learned this phenomenon well. The mean accuracy across the 96 configurations is 92.9\%, with a minimum of 51.3\% and a maximum of 100\%. The BAD parse rate ranges from 0.0\% to 100\%, with a mean of 13.9\%.

The main weakness of this phenomenon appears in SVO profiles with Deranking, ERG-ABS, and non-SENT strategies. Under these configurations, POSS-ACC, ERG-POSS, and NOMN have mean accuracies of 55.3\%, 69.0\%, and 84.2\%, respectively.

These errors may involve alternative parses. For example, under SVO/GN + ERG-ABS + Deranking + POSS-ACC, the GOOD sentence \textit{Supervisorca observes traineege cleaning table} is perturbed into \textit{Supervisorca observes traineege table cleaning}. The GOOD sentence roughly corresponds to ``the supervisor observes [the trainee's cleaning of the table].'' In the BAD sentence, \textit{traineege table} is reanalyzed as a genitive NP functioning as S in a clausal complement, while \textit{cleaning} is its nonfinite predicate. The resulting analysis roughly corresponds to ``the supervisor observes that the trainee's table cleans.''

\subsubsection{ANC External}
\label{app:anc-external}

\ancevalphenomenon{5.1}{ANC External S Marker}{app:e1-27-anc-ext-s-marker}

This phenomenon tests whether the models learned the external marker on the ANC head when the whole ANC functions as matrix intransitive S. The GOOD sentence is a finite intransitive clause in which matrix S is an ANC. The source templates cover several ANC-internal structures, including bare IV ANCs, IV+S ANCs, bare TV ANCs, TV+A ANCs, TV+P ANCs, and TV+A+P ANCs.

In all target grammars, the finite S marker is zero. Therefore, when the whole ANC functions as matrix S, the ANC head in the GOOD sentence should not receive an external overt marker. The BAD sentence is constructed by adding \textit{-ca} or \textit{-ge} to the ANC head. An example is shown below:

\begin{exe}
\ex\label{ex:e1-27-anc-ext-s-marker}
\begin{xlist}
\ex GOOD: \textit{Adjusting seatob succeeds}
\ex BAD: \textit{Adjustingge seatob succeeds}
\end{xlist}
\end{exe}

\begin{figure}[htbp]
\centering
\includegraphics[width=\columnwidth]{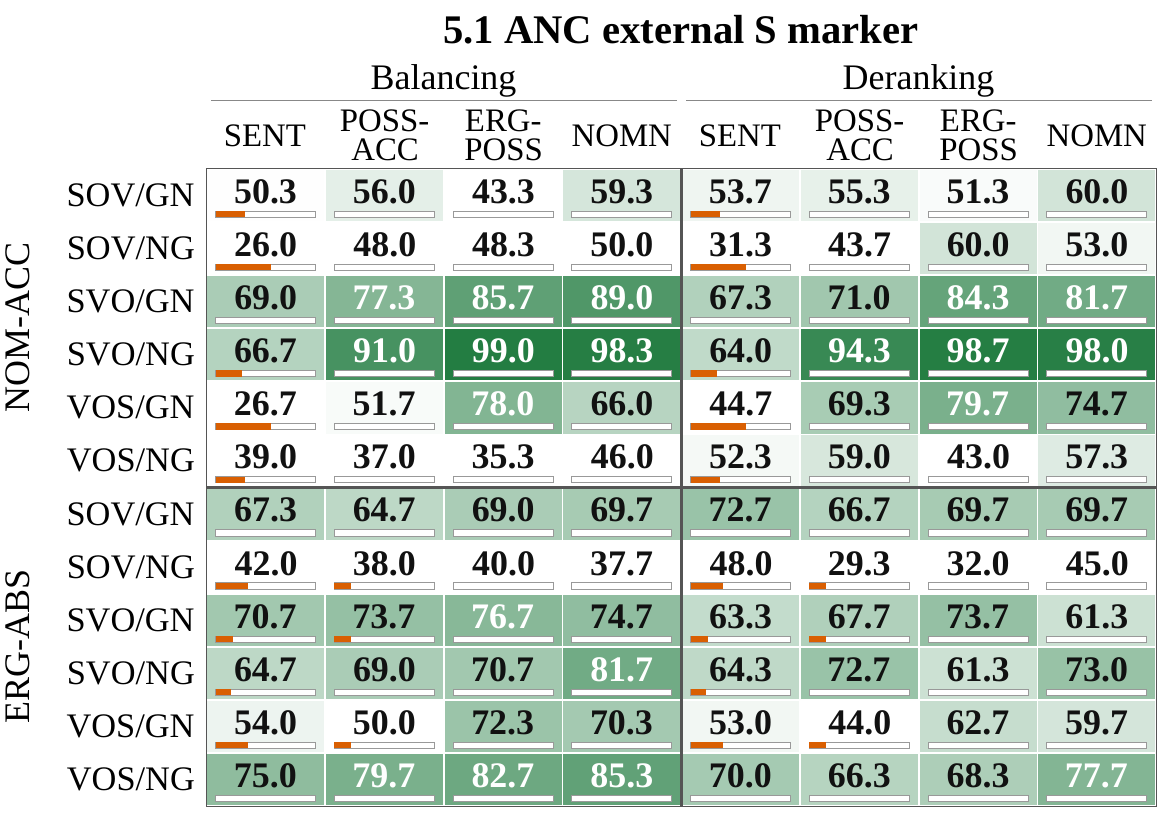}
\caption{Accuracy and BAD parse rate for 5.1 ANC external S marker.}
\label{fig:e1-27-anc-ext-s-marker}
\end{figure}

Figure~\ref{fig:e1-27-anc-ext-s-marker} shows that this phenomenon is difficult for the models. The mean accuracy across the 96 configurations is 62.9\%, with a minimum of 26.0\% and a maximum of 99.0\%. The BAD parse rate ranges from 0.0\% to 55.0\%, with a mean of 7.1\%.

The configurations with relatively strong performance are mainly SVO/NOM-ACC/non-SENT, with a mean accuracy of 89.0\%. By contrast, the remaining configurations have a mean accuracy of only 59.1\%. This shows that, in most configurations, the models did not robustly learn that the ANC head should not take an external marker when the ANC functions as matrix S.

The reason for these errors is not yet fully clear. For example, under VOS/NG + NOM-ACC + Balancing + ERG-POSS, the GOOD sentence \textit{Succeeds adjusting seatge} is perturbed into \textit{Succeeds adjustingge seatge}. This BAD sentence contains two \textit{-ge} markers and does not seem to have any possible alternative parse, but the model still prefers the BAD sentence.

\ancevalphenomenon{5.2}{ANC External A Marker}{app:e1-28-anc-ext-a-marker}

This phenomenon tests whether the models learned the external marker on the ANC head when the whole ANC functions as matrix transitive A. The GOOD sentence is a finite transitive clause in which matrix A is an ANC and matrix P is an ordinary NP. The source templates cover several ANC-internal structures, including bare IV ANCs, IV+S ANCs, bare TV ANCs, TV+A ANCs, TV+P ANCs, and TV+A+P ANCs.

In the target grammars, when an ANC functions as matrix A, its external A marker is the same as the finite A marker. Under NOM-ACC, A is zero-marked; under ERG-ABS, A receives \textit{-ca}. The BAD sentence changes only the external A marker on the ANC head: under NOM-ACC, it adds \textit{-ca} or \textit{-ge}; under ERG-ABS, it deletes \textit{-ca} or replaces it with \textit{-ge}. An example is shown below:

\begin{exe}
\ex\label{ex:e1-28-anc-ext-a-marker}
\begin{xlist}
\ex GOOD: \textit{Chefge cooking delights guestca}
\ex BAD:  \textit{Chefge cookingca delights guestca}
\end{xlist}
\end{exe}

\begin{figure}[htbp]
\centering
\includegraphics[width=\columnwidth]{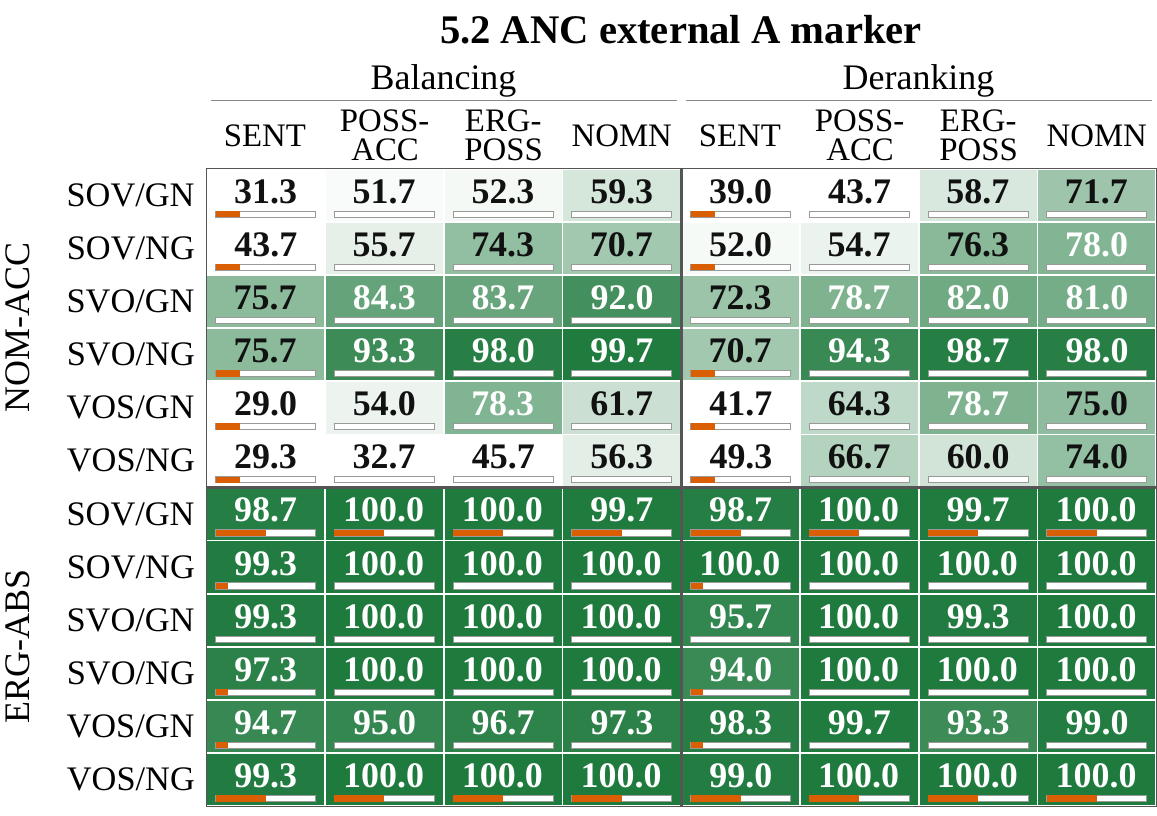}
\caption{Accuracy and BAD parse rate for 5.2 ANC external A marker.}
\label{fig:e1-28-anc-ext-a-marker}
\end{figure}

The results are shown in Figure~\ref{fig:e1-28-anc-ext-a-marker}. The mean accuracy across the 96 configurations is 82.7\%, with a minimum of 29.0\% and a maximum of 100\%. The BAD parse rate ranges from 0.0\% to 50.0\%, with a mean of 11.6\%.

This phenomenon shows the same pattern as Transitive A Marker in Section~\ref{app:e1-5-tran-a-marker}. ERG-ABS configurations perform well, with a mean accuracy of 99.0\%. By contrast, NOM-ACC configurations are much harder, with a mean accuracy of only 66.4\%. This shows that the models easily learned the overt external A marker \textit{-ca}, but did not robustly learn the zero marking of ANC external A under NOM-ACC.

\ancevalphenomenon{5.3}{ANC External P Marker}{app:e1-29-anc-ext-p-marker}

This phenomenon tests whether the models learned the external marker on the ANC head when the whole ANC functions as matrix transitive P. The GOOD sentence is a finite transitive clause in which matrix A is an ordinary NP and matrix P is an ANC. The source templates cover several ANC-internal structures, including bare IV ANCs, IV+S ANCs, bare TV ANCs, TV+A ANCs, TV+P ANCs, and TV+A+P ANCs.

In the target grammars, when an ANC functions as matrix P, its external P marker is the same as the finite P marker. Under NOM-ACC, P receives \textit{-ca}; under ERG-ABS, P is zero-marked. The BAD sentence changes only the external P marker on the ANC head: under NOM-ACC, it deletes \textit{-ca} or replaces it with \textit{-ge}; under ERG-ABS, it adds \textit{-ca} or \textit{-ge}. An example is shown below:

\begin{exe}
\ex\label{ex:e1-29-anc-ext-p-marker}
\begin{xlist}
\ex GOOD: \textit{Team praises playerge improveingca}
\ex BAD:  \textit{Team praises playerge improveing}
\end{xlist}
\end{exe}

\begin{figure}[htbp]
\centering
\includegraphics[width=\columnwidth]{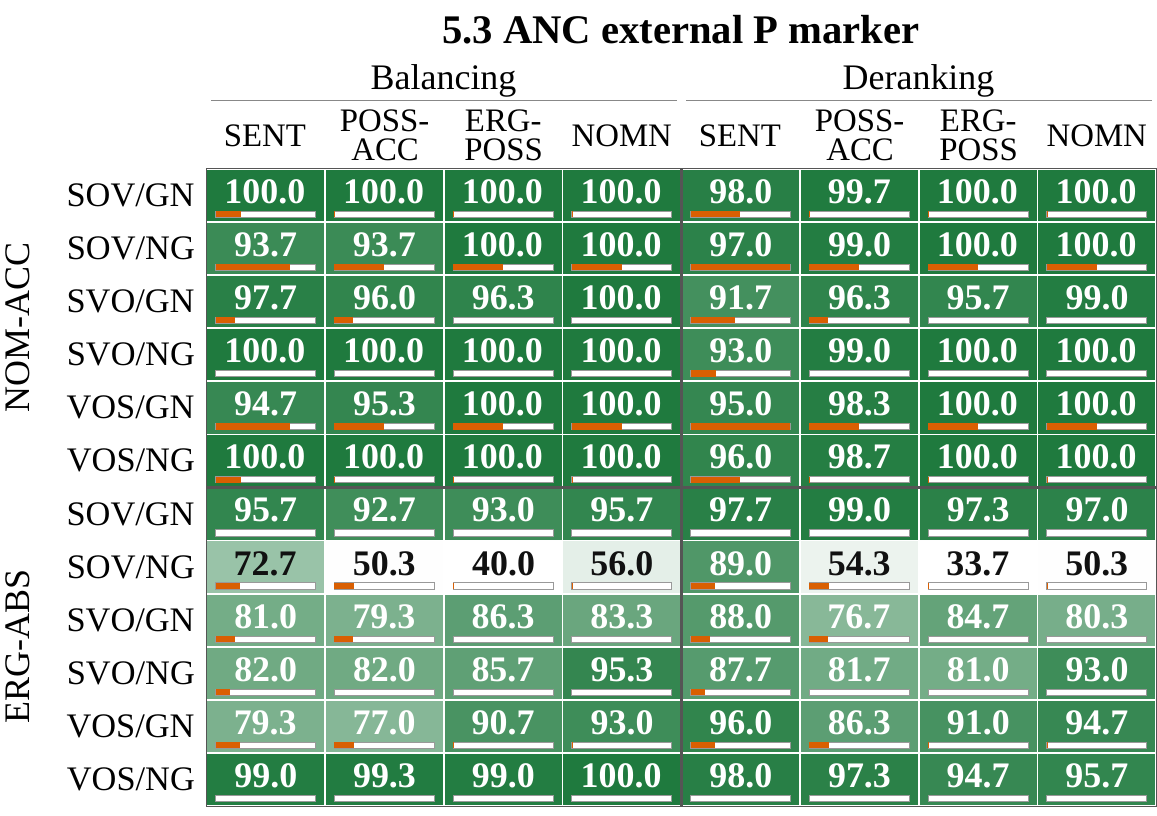}
\caption{Accuracy and BAD parse rate for 5.3 ANC external P marker.}
\label{fig:e1-29-anc-ext-p-marker}
\end{figure}

The results are shown in Figure~\ref{fig:e1-29-anc-ext-p-marker}. The mean accuracy across the 96 configurations is 91.4\%, with a minimum of 33.7\% and a maximum of 100\%. The BAD parse rate ranges from 0.0\% to 99.0\%, with a mean of 15.8\%.

This phenomenon shows the same pattern as Transitive P Marker in Section~\ref{app:e1-6-tran-p-marker}. NOM-ACC configurations perform well, with a mean accuracy of 98.4\%. By contrast, ERG-ABS configurations are harder, with a mean accuracy of 84.4\%. As in ANC External A Marker in Section~\ref{app:e1-28-anc-ext-a-marker}, the models learn overt marking better than zero marking here as well. Within the ERG-ABS configurations, the especially difficult case is SOV/NG + ERG-ABS, with a mean accuracy of only 55.8\%. In this configuration, the ANC head may be separated from the matrix verb by ANC-internal arguments, so the erroneous external P marker is farther from the main-clause verb and becomes harder for the model to identify as a matrix-level marking error.

\ancevalphenomenon{5.4}{ANC External Genitive Marker}{app:e1-30-anc-ext-gen-marker}

This phenomenon tests whether the models learned the external genitive marker on the ANC head when the whole ANC functions as a genitive possessor. The GOOD sentence is a finite clause in which the possessor inside a genitive NP is an ANC. The source templates cover four ANC-internal structures: bare IV ANCs, IV+S ANCs, bare TV ANCs, and TV+A ANCs. TV+P and TV+A+P ANCs are not included in this test because the corresponding English source sentences cannot be parsed reliably. For example, in a structure such as \textit{the reading of the note's effect}, the possessive tends to be analyzed as modifying the internal P of the ANC rather than the whole ANC. Such expressions are also relatively unnatural in English.

In all target grammars, the genitive marker is \textit{-ge}. Therefore, when an ANC functions as a genitive possessor, the ANC head must carry the external genitive marker \textit{-ge}. The BAD sentence is constructed by changing the external genitive marker on the ANC head, either deleting \textit{-ge} or replacing it with \textit{-ca}. An example is shown below:

\begin{exe}
\ex\label{ex:e1-30-anc-ext-gen-marker}
\begin{xlist}
\ex GOOD: \textit{Selectingge size hurts visitorca}
\ex BAD:  \textit{Selectingca size hurts visitorca}
\end{xlist}
\end{exe}

\begin{figure}[htbp]
\centering
\includegraphics[width=\columnwidth]{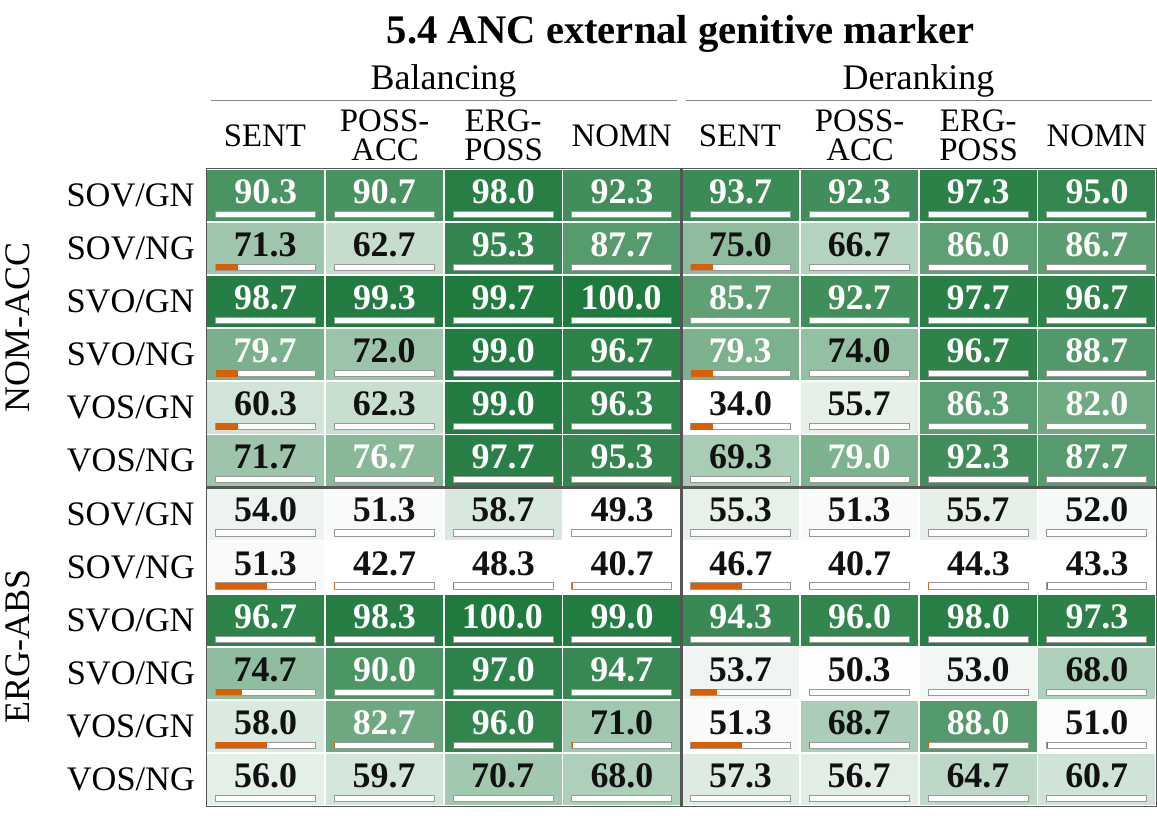}
\caption{Accuracy and BAD parse rate for 5.4 ANC external genitive marker.}
\label{fig:e1-30-anc-ext-gen-marker}
\end{figure}

The results are shown in Figure~\ref{fig:e1-30-anc-ext-gen-marker}. The mean accuracy across the 96 configurations is 75.9\%, with a minimum of 34.0\% and a maximum of 100\%. The BAD parse rate is low, with a mean of 4.2\%.

Across parameters, Balancing and Deranking have relatively small effects, with mean accuracies of 79.2\% and 72.7\%, respectively. By contrast, strategy, word order, and alignment have clearer effects. ERG-POSS and NOMN perform relatively well, with mean accuracies of 84.1\% and 79.2\%, while SENT and POSS-ACC are weaker, at 69.1\% and 71.3\%. In addition, SOV + ERG-ABS configurations are relatively weak, with a mean accuracy of 49.1\%.

The reason for this complex pattern is not yet clear. One possibility is distributional imbalance in the training corpus. P arguments and copular complements, both realized as P in the generated corpora, together account for 76.6\% of ANC occurrences, while only 0.24\% function as genitive possessors.

\subsubsection{ANC Argument Omission}
\label{app:anc-argument-omission}

\ancevalphenomenon{6.1}{ANC Omitted S}{app:e1-31-anc-omit-s}

This phenomenon tests whether the models distinguish bare intransitive ANCs from finite intransitive clauses. Bare intransitive ANCs allow the S argument to be omitted, whereas finite clauses require an overt S. The GOOD sentence is a finite transitive clause in which matrix P is an intransitive ANC without an overt S.

In all target grammars, the ANC verb uses the nonfinite form \textit{V-ing}. The BAD sentence is constructed by replacing the bare IV ANC head with the corresponding finite intransitive predicate. An example is shown below:

\begin{exe}
\ex\label{ex:e1-31-anc-omit-s}
\begin{xlist}
\ex GOOD: \textit{Company questions disappearingca}
\ex BAD:  \textit{Company questions disappears}
\end{xlist}
\end{exe}

\begin{figure}[htbp]
\centering
\includegraphics[width=\columnwidth]{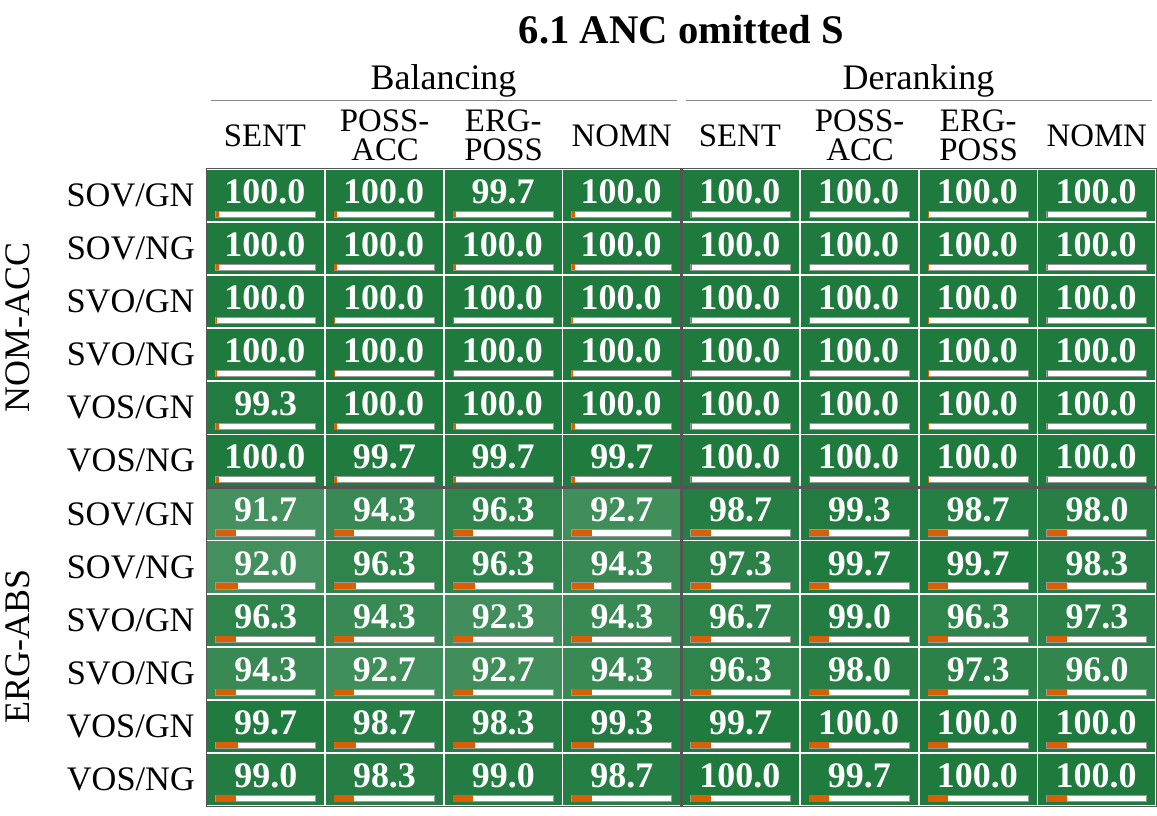}
\caption{Accuracy and BAD parse rate for 6.1 ANC omitted S.}
\label{fig:e1-31-anc-omit-s}
\end{figure}

The results are shown in Figure~\ref{fig:e1-31-anc-omit-s}. The mean accuracy across the 96 configurations is 98.5\%, with a minimum of 91.7\% and a maximum of 100\%. The mean BAD parse rate is 11.0\%. Overall, the models robustly learned that bare intransitive ANCs can function as matrix arguments and that the ANC head should take the nonfinite form.

\ancevalphenomenon{6.2}{ANC Omitted A}{app:e1-32-anc-omit-a}

This phenomenon tests whether the models distinguish transitive ANCs with omitted A from ordinary finite transitive clauses. A transitive ANC with omitted A can retain only the P argument, whereas a finite transitive clause requires both A and P. The GOOD sentence is a finite transitive clause in which matrix P is a transitive ANC with overt P and omitted A.

In the target grammars, the ANC verb uses the nonfinite form \textit{V-ing}, and ANC-internal P receives the P marker determined by the ANC strategy. The BAD sentence is constructed by replacing the ANC head with the corresponding finite transitive predicate and changing the ANC-internal P marker to the finite P marker. An example is shown below:

\begin{exe}
\ex\label{ex:e1-32-anc-omit-a}
\begin{xlist}
\ex GOOD: \textit{They questions countingca ballotob}
\ex BAD:  \textit{They questions counts ballotca}
\end{xlist}
\end{exe}

\begin{figure}[htbp]
\centering
\includegraphics[width=\columnwidth]{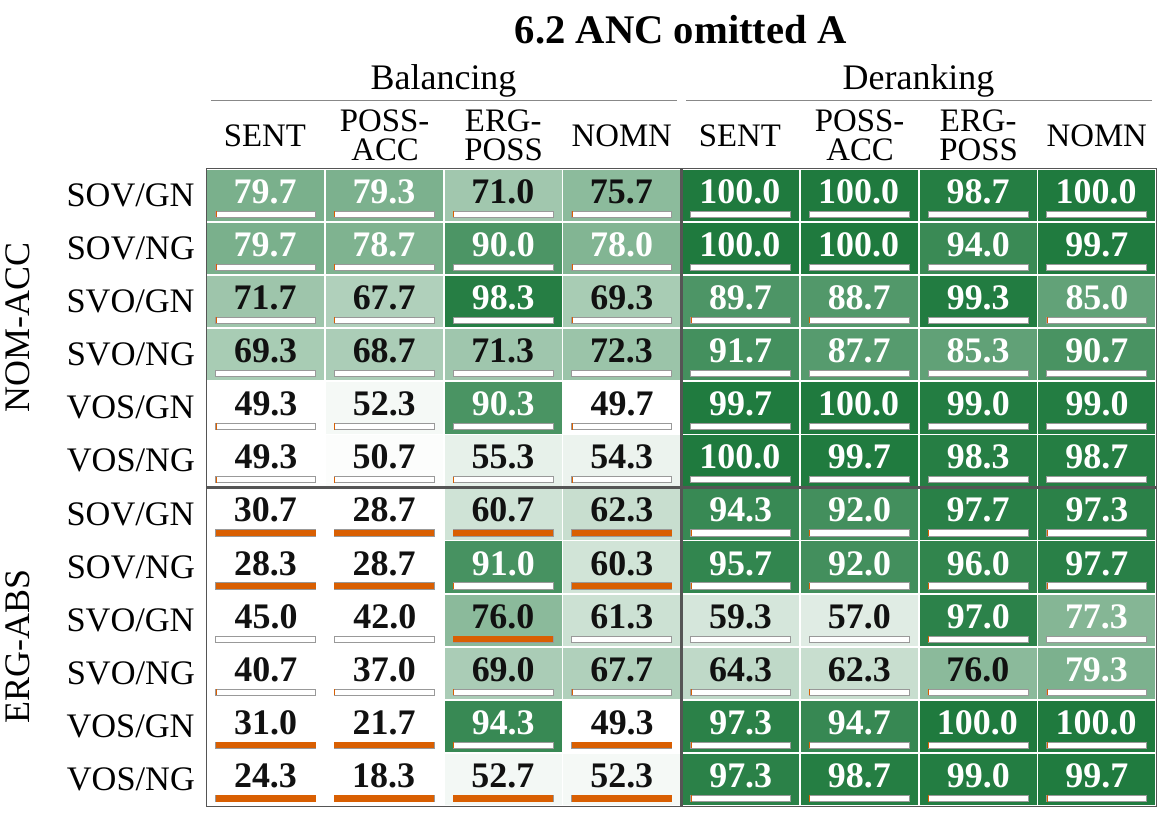}
\caption{Accuracy and BAD parse rate for 6.2 ANC omitted A.}
\label{fig:e1-32-anc-omit-a}
\end{figure}

The results are shown in Figure~\ref{fig:e1-32-anc-omit-a}. The mean accuracy across the 96 configurations is 75.8\%, with a minimum of 18.3\% and a maximum of 100\%. The BAD parse rate ranges from 0.0\% to 100\%, with a mean of 16.1\%.

Across parameters, Deranking is much easier than Balancing. Deranking grammars have a mean accuracy of 92.2\%, while Balancing grammars have a mean accuracy of only 59.3\%. Within Balancing grammars, ERG-ABS is especially difficult, with a mean accuracy of 48.9\% and a BAD parse rate of 62.8\%. By contrast, NOM-ACC + Balancing reaches 69.7\% accuracy, with a BAD parse rate of only 0.7\%.

The main errors come from BAD parses involving transitive/intransitive alternation. A BAD sentence missing one transitive argument can be reanalyzed as an intransitive structure with exactly one overt argument.

There are also errors that are not due to BAD parses. For example, under SVO/NG + ERG-ABS + Balancing + POSS-ACC, the accuracy is only 37.0\%, but the BAD parse rate is only 1.0\%. The GOOD sentence \textit{Theyca questions counting ballot} is perturbed into \textit{Theyca questions counts ballot}. This BAD sentence usually cannot be parsed, because the finite transitive predicate \textit{counts ballot} lacks an A argument. Still, its surface form resembles a clausal-complement structure with a matrix clausal-complement verb followed by an embedded finite predicate.

\ancevalphenomenon{6.3}{ANC Omitted P}{app:e1-33-anc-omit-p}

This phenomenon tests whether the models distinguish transitive ANCs with omitted P from ordinary finite transitive clauses. A transitive ANC with omitted P can retain only the A argument, whereas a finite transitive clause requires both A and P. The GOOD sentence is a finite transitive clause in which matrix P is a transitive ANC with overt A and omitted P.

In the target grammars, the ANC verb uses the nonfinite form \textit{V-ing}, and ANC-internal A receives the A marker determined by the ANC strategy. The BAD sentence is constructed by replacing the ANC head with the corresponding finite transitive predicate and changing the ANC-internal A marker to the finite A marker. An example is shown below:

\begin{exe}
\ex\label{ex:e1-33-anc-omit-p}
\begin{xlist}
\ex GOOD: \textit{Officer notices driverge checkingca}
\ex BAD:  \textit{Officer notices driver checks}
\end{xlist}
\end{exe}

\begin{figure}[htbp]
\centering
\includegraphics[width=\columnwidth]{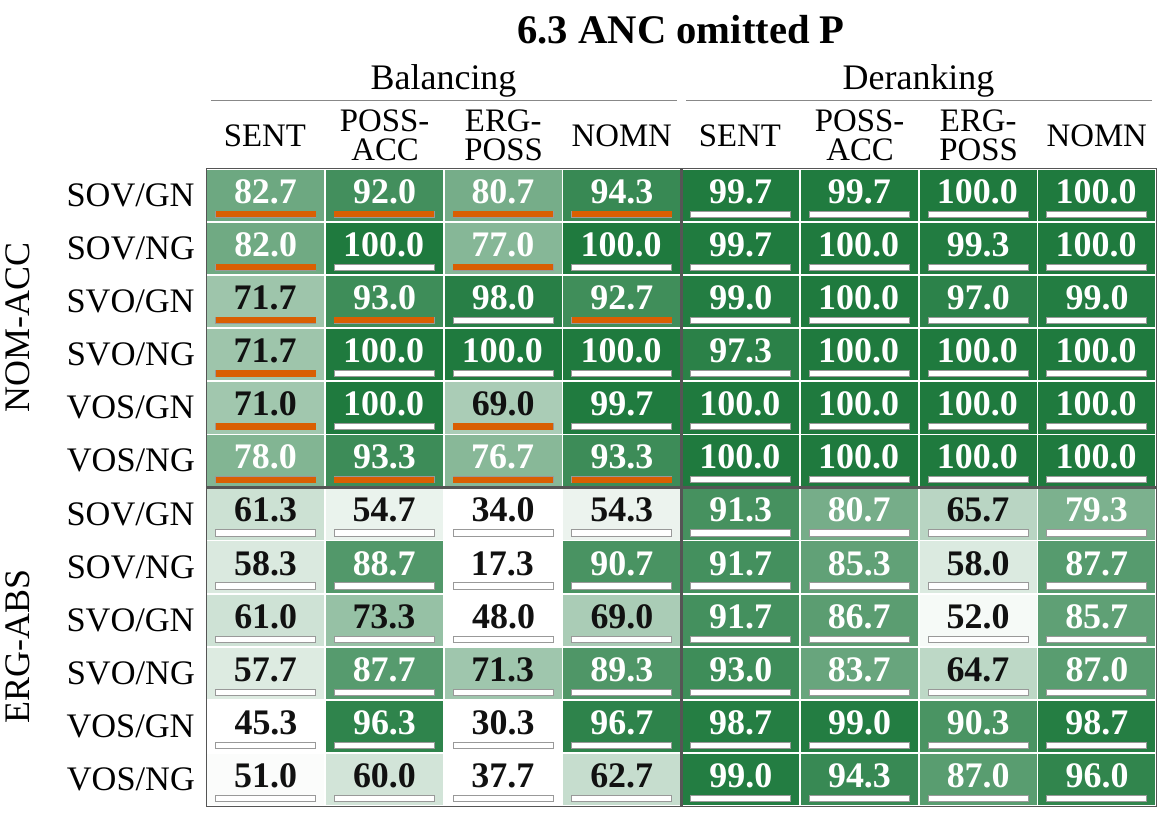}
\caption{Accuracy and BAD parse rate for 6.3 ANC omitted P.}
\label{fig:e1-33-anc-omit-p}
\end{figure}

The results are shown in Figure~\ref{fig:e1-33-anc-omit-p}. The mean accuracy across the 96 configurations is 83.9\%, with a minimum of 17.3\% and a maximum of 100\%. The BAD parse rate ranges from 0.0\% to 100\%, with a mean of 16.7\%.

This phenomenon shows a pattern similar to ANC Omitted A in Section~\ref{app:e1-32-anc-omit-a}. Deranking grammars still perform much better than Balancing grammars, with mean accuracies of 92.5\% and 75.3\%, respectively. Alignment also matters: NOM-ACC reaches 93.9\%, compared with 73.8\% for ERG-ABS.

\ancevalphenomenon{6.4}{ANC Omitted A+P}{app:e1-34-anc-omit-a-p}

This phenomenon tests whether the models distinguish bare transitive ANCs from ordinary finite transitive clauses. Transitive ANCs can omit internal arguments, whereas finite transitive clauses require both A and P. The GOOD sentence is a finite transitive clause in which matrix P is a bare transitive ANC.

In all target grammars, the ANC verb uses the nonfinite form \textit{V-ing}. The BAD sentence is constructed by replacing the bare TV ANC head with the corresponding finite transitive predicate. An example is shown below:

\begin{exe}
\ex\label{ex:e1-34-anc-omit-a-p}
\begin{xlist}
\ex GOOD: \textit{Manager watchs cleaningca}
\ex BAD:  \textit{Manager watchs cleans}
\end{xlist}
\end{exe}

\begin{figure}[htbp]
\centering
\includegraphics[width=\columnwidth]{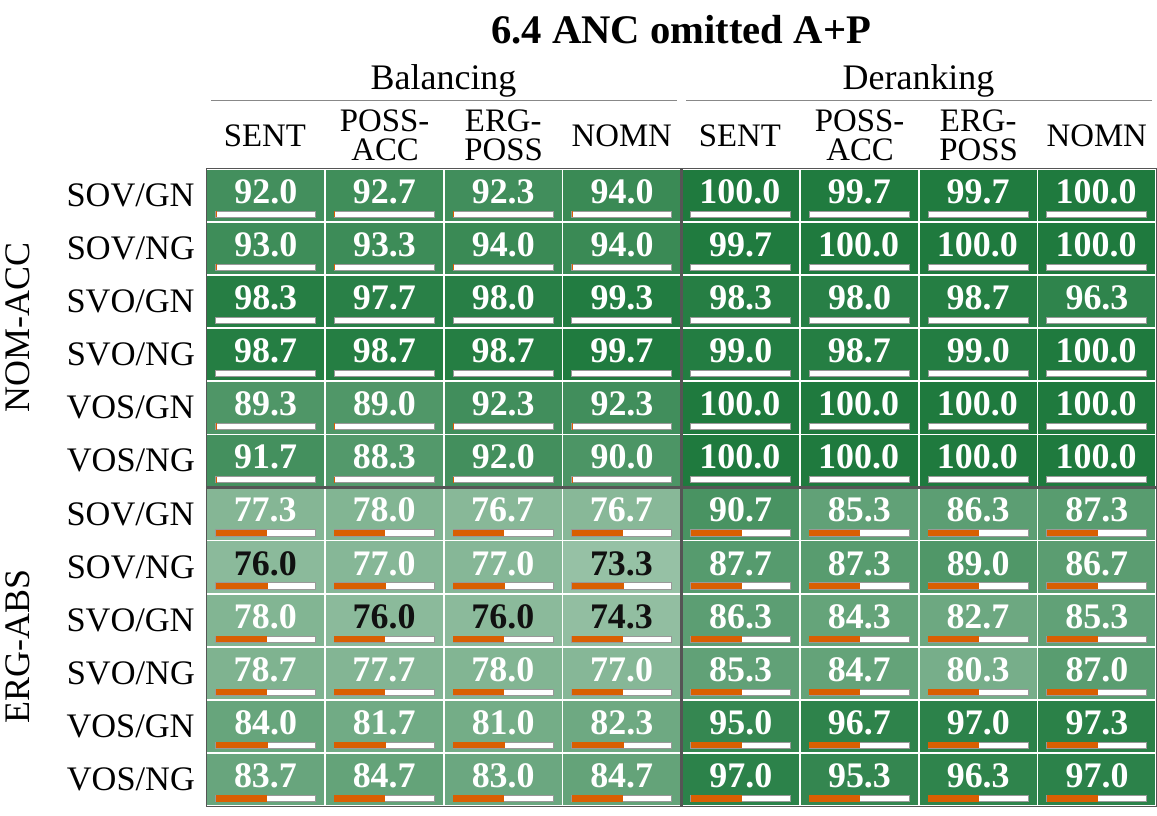}
\caption{Accuracy and BAD parse rate for 6.4 ANC omitted A+P.}
\label{fig:e1-34-anc-omit-a-p}
\end{figure}

The results are shown in Figure~\ref{fig:e1-34-anc-omit-a-p}. The mean accuracy across the 96 configurations is 90.5\%, with a minimum of 73.3\% and a maximum of 100\%. The BAD parse rate ranges from 0.0\% to 52.0\%, with a mean of 25.8\%.

Overall, the models learned this phenomenon well. As in the previous two omitted-argument phenomena, Deranking grammars perform better than Balancing grammars, with mean accuracies of 94.5\% and 86.5\%, respectively. The weakest configuration is Balancing + ERG-ABS, with a mean accuracy of 78.9\%. This phenomenon is easier than ANC Omitted A in Section~\ref{app:e1-32-anc-omit-a} and ANC Omitted P in Section~\ref{app:e1-33-anc-omit-p}, because the BAD sentence lacks both arguments of the finite transitive predicate, making the violation more salient.

\subsection{Variation across Seeds}
\label{app:multi-seed-validation}

To check whether the evaluation results are stable across random seeds, we compare the accuracies of the three models within each phenomenon--configuration condition.

Since the evaluation measures whether a model assigns a higher probability to the grammatical member of each minimal pair, chance-level accuracy is 50\%. We therefore use a softplus transformation to map accuracy to a 0--1 learning score \(L\), so that differences near or below chance are compressed, while differences above chance are more salient. For each phenomenon--configuration condition, we take the difference between the highest and lowest learning scores across the three seeds and define it as seed disagreement \(D\). \(L_s\) and \(D\) are defined as follows:

\[
L_s =
\frac{
\operatorname{softplus}\left(\frac{a_s - 50}{\tau}\right)
-
\operatorname{softplus}\left(\frac{-50}{\tau}\right)
}{
\operatorname{softplus}\left(\frac{50}{\tau}\right)
-
\operatorname{softplus}\left(\frac{-50}{\tau}\right)
}
\]

\[
D = \max_s L_s - \min_s L_s
\]

where \(a_s\) is the accuracy percentage for seed \(s\), \(\tau=5\), and \(\operatorname{softplus}(x)=\log(1+e^x)\).

Overall, the three-seed results are stable. Across the \(34 \times 96\) phenomenon--configuration conditions, the mean seed disagreement is 0.061, the median is 0.020, and the 90th percentile is 0.160. This indicates that most phenomenon--configuration conditions do not depend strongly on a single random seed. The few phenomena with higher seed disagreement also tend to have relatively low overall accuracy. For example, Clausal Complement S Marker in Section~\ref{app:e1-15-clausal-s-marker} has a mean accuracy of 75.2\% and a mean seed disagreement of 0.177, while ANC External Genitive Marker in Section~\ref{app:e1-30-anc-ext-gen-marker} has a mean accuracy of 75.9\% and a mean seed disagreement of 0.141.

\end{document}